\documentclass[12pt]{article}
\pdfoutput=1

\usepackage[margin=1in]{geometry}
\usepackage{amsmath,amsthm,amssymb,amsfonts,bm,physics}
\usepackage{tgtermes}
\usepackage{newtxtext}
\usepackage{newtxmath}
\usepackage{graphicx}
\usepackage{float}
\usepackage{enumitem}
\usepackage{array}
\usepackage{booktabs}
\usepackage{multirow}
\usepackage{afterpage}
\usepackage{listings}
\usepackage{subcaption}
\usepackage{algorithm}
\usepackage{algorithmic}
\usepackage{changepage}
\usepackage[round]{natbib}
\usepackage{authblk}
\usepackage{setspace}
\usepackage{xcolor}
\usepackage{xurl}
\usepackage[hypertexnames=false]{hyperref}
\usepackage[capitalize,noabbrev]{cleveref}

\bibpunct[, ]{(}{)}{,}{a}{}{,}
\definecolor{Bleu}{RGB}{0,0,204}
\hypersetup{
  colorlinks,
  citecolor=Bleu,
  linkcolor=Bleu,
  urlcolor=Bleu
}

\newcommand{\ind}{\mathbf{1}}

\DeclareMathOperator*{\argmin}{argmin}
\DeclareMathOperator*{\argmax}{argmax}

\newcommand{\slightspacing}{\setstretch{1.175}}

\theoremstyle{plain}
\newtheorem{theorem}{Theorem}
\newtheorem{lemma}{Lemma}
\newtheorem{corollary}{Corollary}
\newtheorem{definition}{Definition}

\theoremstyle{remark}
\newtheorem{example}{Example}

\providecommand{\keywords}[1]{%
  \small\textbf{\textit{Keywords---}} #1
}

\title{Semiparametric Double Reinforcement Learning\\
with Applications to Long-Term Causal Inference}
\author[1]{Lars van der Laan\thanks{Corresponding author: \href{mailto:lvdlaan@uw.edu}{lvdlaan@uw.edu}}}
\author[2]{David Hubbard}
\author[2]{Allen Tran}
\author[2,3]{Nathan Kallus}
\author[2]{Aur\'{e}lien Bibaut}
\affil[1]{Department of Statistics, University of Washington, USA}
\affil[2]{Netflix Research, USA}
\affil[3]{Cornell Tech, Cornell University, USA}
\date{}

\begin{document}
\allowdisplaybreaks
\maketitle

\begin{abstract}
\singlespacing
Double reinforcement learning (DRL) provides efficient off-policy inference for policy values in nonparametric Markov decision processes (MDPs), but fully nonparametric estimators can be unstable when intertemporal overlap is weak and occupancy ratios are high-dimensional. This limitation is especially relevant for long-term causal inference from randomized experiments: randomization ensures overlap in treatment assignment, but not over future state trajectories induced by continued intervention use. We develop semiparametric DRL for continuous linear functionals of the infinite-horizon $Q$-function. Rather than impose linear MDP structure on the reward and transition laws, we place working semiparametric restrictions on the $Q$-function itself, the solution of the discounted Bellman equation. When correct, these restrictions can improve efficiency relative to unrestricted DRL while allowing rich, possibly infinite-dimensional models. To avoid relying on correct specification, we define the estimand through weighted Bellman-residual minimization. The resulting projection target remains meaningful under misspecification and recovers the original functional under correct specification. For this class of parameters, we derive efficient influence functions and efficiency bounds, construct model-robust automatically debiased estimators, and develop minimax criteria for estimating the $Q$- and Riesz functions. Under correct specification, optimally weighted versions attain the semiparametric efficiency bound in the restricted model.
\end{abstract}

\keywords{off-policy evaluation; double reinforcement learning; Markov decision processes; infinite horizon; semiparametric efficiency; causal inference; optimal weighting}

\slightspacing

\section{Introduction}
\subsection{Motivation}

Randomized experiments such as A/B tests and clinical trials are widely used to assess the effects of interventions. In industry, econometrics, and clinical applications, these experiments are often short-term, even though the decisions they inform involve long-term outcomes such as survival, customer retention, or revenue 
\citep{chetty2011does, budish2015firms, hohnhold2015focusing, yang2020targeting}. This gap has motivated a central question in recent research: \textit{How can we estimate the long-term causal effect of an intervention when only short-term experiments are available?} \citep{athey2025surrogate, van2023estimating, zhang2023evaluating, nam2024predicting, imbens2025long}

A common strategy is to use \emph{surrogate} outcomes---intermediate variables observed during the experiment that are assumed to mediate the long-term effect \citep{prentice1989surrogate, frangakis2002principal, zhang2023evaluating, athey2025surrogate}. These methods combine short-term experimental data with long-term observational data by treating the surrogate as a proxy for the long-term outcome. For example, a streaming platform might use one-month watch time in an A/B test as a surrogate for annual subscription retention in historical data. A key limitation is that the \emph{short-term} surrogate must fully mediate the \emph{long-term} treatment effect \citep{vanderweele2013surrogate, weintraub2015perils, stijven2025evaluation}. This assumption can fail when treatment effects accumulate over time \citep{tran2023inferring}, as in daily dosing or a continuously deployed recommendation algorithm. For instance, annual retention under a new recommendation algorithm may depend on evolving user behavior not captured by a single short-term surrogate.

\begin{figure}[!htb]
\centering
\includegraphics[width=0.7\linewidth]{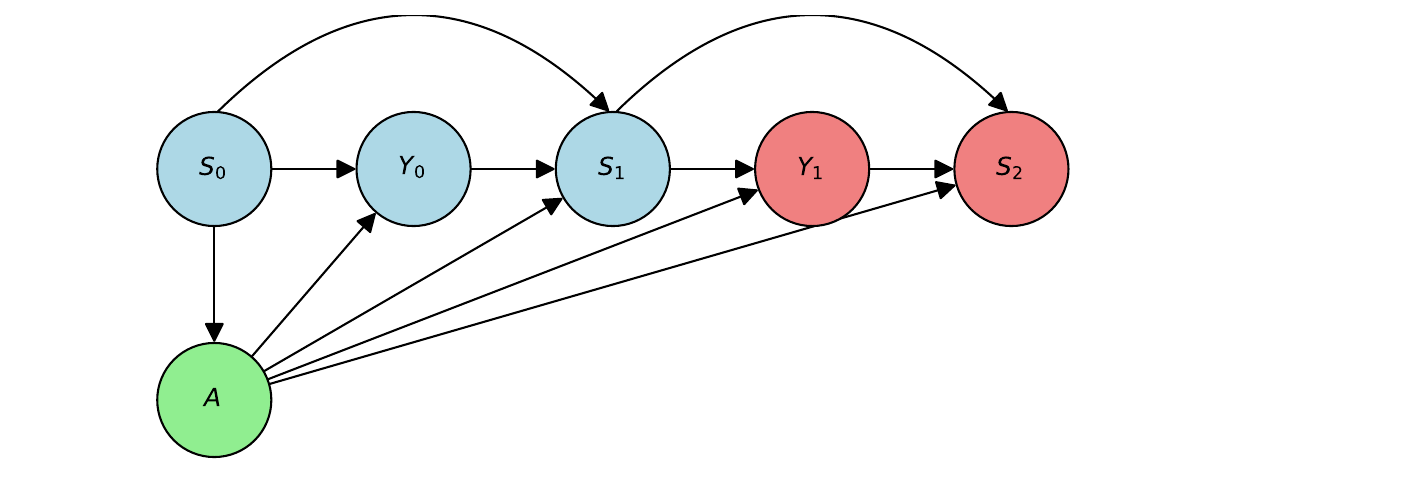}
\caption{\small Dynamic causal graph for the Markov decision process formulation. The variable $A$ denotes a sustained treatment or action, $S_t$ denotes the evolving state, and $Y_t$ denotes an intermediate outcome or reward. In a short-term experiment, later quantities such as $Y_1$ and $S_2$ need not be observed; they are projected through the Markov transition model.}
\label{fig:dag}
\end{figure}

These limitations motivate dynamic approaches that model how treatment effects
accumulate through evolving states and outcomes. To this end,
\citet{tran2023inferring} proposed estimating long-term effects from
short-term experiments using a time-homogeneous Markov decision process (MDP)
(Figure~\ref{fig:dag}). Unlike surrogate-based methods, this approach does not
require the long-term effect to be mediated by a single short-term variable;
instead, it treats short-term experimental data as observations of the
state--action dynamics and uses these dynamics to project long-run outcomes.
In this formulation, the state can include baseline covariates, short-term
metrics, and surrogate variables that inform future behavior, while rewards may
include monthly measures of retention, engagement, or revenue. The parameter of
interest is the long-run value of the intervention: the expected sum of these
monthly rewards along the resulting trajectory, with future rewards optionally
discounted. Related ideas appear in \citet{battocchi2021estimating},
\citet{liao2021off}, and \citet{bibaut2021sequential}. This line of work
connects long-term causal inference with offline reinforcement learning and
enables efficient nonparametric inference via double reinforcement learning
(DRL) \citep{jiang2016doubly,van2018online,kallus2020double,kallus2022efficiently}.

DRL enables statistically efficient inference on policy values in nonparametric MDPs using off-policy data, but fully nonparametric DRL can be highly sensitive to \emph{intertemporal overlap} \citep{xie2023semiparametrically}. This condition requires the observed state--action distribution to provide sufficient coverage of the long-run state--action distribution induced by the target intervention. When this overlap is weak, occupancy ratios can become large or unstable, making DRL estimators noisy and requiring prohibitively large samples, especially in high-dimensional or unbounded state spaces \citep{mehrabi2024off}. This creates a fundamental challenge even in randomized experiments. Randomization removes confounding in treatment assignment and ensures overlap in the actions observed during the experiment, but it does not ensure coverage of the future states generated by sustained intervention use, because those states evolve through the transition law. A closely related difficulty is well known in offline RL: short-horizon behavior data may fail to cover the long-run state distribution induced by the target policy, leading to extrapolation error under temporal distribution shift \citep{gottesman2019guidelines,levine2020offline,jin2021pessimism,uehara2021pessimistic}.

One response to this difficulty is to impose structure on the MDP. For example,
linear-MDP models assume that both the expected reward and transition kernel are
linear in a known finite-dimensional state--action feature map
\citep{wang2020statistical,jin2020provably,xie2023semiparametrically}. Such
restrictions can reduce the dependence of inference on the full occupancy
ratio, replacing overlap in the original state--action space with coverage in a
lower-dimensional feature space. This can substantially improve efficiency when
the linear-MDP model is correct \citep{xie2023semiparametrically}, but it
requires correctly specified structure on both rewards and transitions. We take
a different route. Policy values and many other long-run causal summaries are
linear functionals of a structural function: the $Q$-function, which maps a
current state--action pair to the expected discounted long-run outcome. This
motivates placing semiparametric structure directly on the $Q$-function,
paralleling classical semiparametric regression models in static settings
\citep{robinson1988root}. It also motivates weighted projection targets that
can reduce sensitivity to poorly covered regions of the state--action space,
echoing the role of overlap-weighted estimands in static causal inference
\citep{imbensOverlapEstimand2006,li2019overlapWeights}.

\subsection{Contributions}

We develop a model-robust framework for inference on continuous linear
functionals of the infinite-horizon \(Q\)-function under working
semiparametric restrictions. By placing structure directly on the
\(Q\)-function rather than on the reward and transition laws, the framework
extends DRL beyond policy values and fully nonparametric \(Q\)-function models.
Under correct specification, these restrictions can reduce sensitivity to
intertemporal overlap and improve efficiency relative to unrestricted DRL,
while accommodating rich, possibly infinite-dimensional function classes.

We do not require the \(Q\)-function restriction to be correct. Instead, we use
it to define a weighted Bellman projection and target linear functionals of
this projection. These targets remain well defined under misspecification and
recover the original functionals under correct specification. Our main
contributions are as follows.

\begin{enumerate}
\item \textbf{Projection targets under working restrictions.}
We define weighted Bellman projection parameters for continuous linear
functionals of the infinite-horizon \(Q\)-function and characterize their
approximation bias. This bias has a doubly robust, second-order form: the
projection target remains close to the original functional when either the
\(Q\)-function or its Bellman--Riesz representer is well approximated by the
working classes.

\item \textbf{Automatic, model-robust estimation.}
We construct automatically debiased DRL (autoDRL) estimators of the projection
targets under doubly robust rate conditions. The required nuisance functions
are characterized by minimax \(Q\)-learning and Riesz-learning problems.
For finite-dimensional linear and RKHS classes, profiling out the inner
maximizations reduces these problems to standard supervised-learning
objectives.

\item \textbf{Efficiency and optimal weighting.}
We establish pathwise differentiability, derive efficient influence functions
and von Mises expansions, and prove asymptotic linearity of autoDRL. For each
fixed projection weight, autoDRL attains the nonparametric efficiency bound for
the corresponding projection target. Under correct specification, all
admissible weights target the original functional, and the dynamic analogue of
inverse conditional-variance weighting attains its restricted-model efficiency
bound. Under correct specification, inference remains valid when the estimated
weight converges to an admissible limit, whether or not that limit is optimal.
\end{enumerate}

\medskip
\noindent This paper is organized as follows. Section~\ref{section::prelim} introduces
Markov decision processes, \(Q\)-functions, and the target estimands.
Section~\ref{section::overlap} explains the intertemporal-overlap challenge
that motivates semiparametric restrictions on the \(Q\)-function.
Section~\ref{section::estmeth} defines weighted Bellman projection parameters
and constructs the corresponding model-robust autoDRL estimators, including
minimax nuisance-estimation criteria. Section~\ref{section::theory} develops
the efficiency theory, asymptotic linearity, and optimal weighting results.

\subsection{Related work}
\noindent\textbf{Off-policy evaluation and double reinforcement learning.}
Our work builds on semiparametric off-policy evaluation in Markov decision
processes, especially doubly robust and efficient estimators based on
\(Q\)-functions and marginalized occupancy ratios
\citep{jiang2016doubly,thomas2016data,liu2018breaking,tang2019doubly,
kallus2020double,kallus2022efficiently,shi2021deeply,shi2022statistical,
wang2023projected,mehrabi2024off}. This literature primarily targets policy
values, either nonparametrically or under correctly specified finite-dimensional
linear models or sieves. Several recent works study off-policy inference under
linear function approximation or linear-MDP restrictions
\citep{wang2020statistical,xie2023semiparametrically,wu2024statistical}, with
linear-MDP models imposing parametric structure on both the reward function and transition
kernel. We instead study general linear functionals of the \(Q\)-function under
a working semiparametric restriction that may be misspecified.

Bellman-residual minimization is a classical technique in approximate dynamic
programming and reinforcement learning with function approximation
\citep{sutton1998reinforcement,baird1999reinforcement,munos2008finite,
maillard2010finite,yu2012weighted,patterson2022generalized}. In that
literature, Bellman projection is primarily used as an algorithmic device for
approximating value functions. We use the same projection idea for a different
purpose: to define a statistical target for inference under working-model
misspecification. This leads to an inferential problem distinct from
nonparametric DRL. Even for policy-value functionals, the efficient influence
function for the Bellman projection target contains a projection-residual
correction that is absent from the nonparametric DRL influence function of
\citet{kallus2020double}. Our results derive the semiparametric efficiency
bound for this projection target and show that, under correct specification,
variance-aware weighting attains the restricted-model efficiency bound for the
original policy-value target.

\medskip
\noindent\textbf{Inverse problems, conditional moments, and optimal weighting.}
The Bellman equation can be viewed as a conditional moment restriction and
formulated as a linear inverse problem. This connects our work to
nonparametric instrumental variables, minimax methods for conditional moments,
and efficient GMM
\citep{hansen1982large,chamberlain1987asymptotic,newey1993efficient,
ai2003efficient,ai2012semiparametric,carrasco2007linear,
dikkala2020minimax,bennett2022inference,bennett2023source,
bennett2023minimax,olivas2025source}. Classical GMM and optimal-instrument
results characterize efficiency for correctly specified finite-dimensional
moment models \citep{hansen1982large,chamberlain1987asymptotic,
newey1993efficient}. Semiparametric extensions use sieve methods to estimate
unknown functions entering the moment restrictions \citep{ai2003efficient},
while optimal weighting delivers semiparametric efficiency
\citep{ai2012semiparametric}. Sieve minimum-distance methods provide inference
for possibly irregular functionals of semiparametric and nonparametric
conditional moment models \citep{chen2015sieve}.  
More recent minimax and debiased inverse-problem methods accommodate flexible
nuisance classes through regularization, orthogonalization, and bias correction
\citep{dikkala2020minimax,bennett2022inference,bennett2023source,
bennett2023minimax,olivas2025source}.

For Bellman conditional moments specifically, \citet{chen2022well} recast
nonparametric \(Q\)-function estimation as an NPIV-style inverse problem and
study \(L^2\)-well-posedness. Building on this perspective,
\citet{chen2025inference} develop optimally weighted sieve-based inference for
expectation functionals, including policy values, under correctly specified
conditional moment restrictions.

Our contribution is complementary. We develop semiparametric inference and
efficiency theory for continuous linear functionals of the \(Q\)-function, while
allowing generic machine-learning estimation of nuisance functions beyond sieve
estimation. We establish pathwise differentiability of the projection parameter,
derive efficient influence functions, and obtain root-\(n\) regular estimators. This differs from a related strand of the conditional-moment literature, which
develops debiased inference procedures for functionals of solutions to possibly
ill-posed inverse problems; in such settings, the target functional may fail to
be pathwise differentiable, precluding root-\(n\) regular estimation and a
standard semiparametric efficiency bound, and complicating uniformly valid
inference
\citep{bennett2022inference,van2025nonparametric,smucler2025asymptotic}. Our working semiparametric restriction may also be misspecified. We therefore
target a Bellman projection estimand indexed by a projection weight, and the
efficient influence function contains an additional projection-residual
correction. The role of the projection weight parallels optimal weighting in
semiparametric conditional moment models \citep{ai2012semiparametric}. Under
correct specification, the efficient Bellman weight is the dynamic analogue of
inverse-variance weighting and yields the restricted-model efficiency bound.

\medskip
\noindent\textbf{Automatic debiasing and Riesz representers.}
Automatic debiased machine learning uses Riesz representers to construct bias
corrections for continuous linear functionals of structural functions
\citep{chernozhukov2018double,chernozhukov2018learning,
chernozhukov2022automatic,van2025automatic}. Our work is closest to autoDML
for linear inverse problems, where bias corrections are based on Riesz
representers that solve dual inverse problems
\citep{bennett2022inference,bennett2023source}. Our construction is related,
but differs in the geometry of the dual problem: the Riesz representer is
defined through the weighted Bellman inner product induced by the Bellman
operator and the projection weight. A direct adaptation of the autoDML
estimators of \citet{bennett2022inference,bennett2023source} would not, by
itself, yield the efficient estimator under correct specification or the correct
influence function for the Bellman projection target under misspecification.
Model-robust debiasing requires an additional projection-residual nuisance and
corresponding correction term, while efficiency under correct specification
requires optimal weighting. This differs from static regression settings, where
autoDML can be robust to misspecification without additional correction terms
\citep{chernozhukov2022automatic}.

\section{Problem setup}

 \label{section::prelim}

\subsection{Data structure and Markov decision model}
\label{section::prelim2}
 
We consider a randomized experiment or observational study in which individuals
sequentially receive treatments, or actions, according to a behavior policy. At
each time \(t\ge 0\), an individual occupies a state \(S_t\in\mathcal S\),
receives an action \(A_t\in\mathcal A\), generates an immediate outcome
\(Y_t\in\mathbb R\), and transitions to a next state \(S_{t+1}\). The observed
data consist of \(n\) independent trajectories of common fixed length \(T\),
\(O_i=(S_{i,0},A_{i,0},Y_{i,0},\ldots,S_{i,T})\), with
\(O_1,\ldots,O_n\overset{\mathrm{iid}}{\sim}\mathbb P_0\). For a generic
trajectory \(O\sim\mathbb P_0\), write
\(Z_t=(S_t,A_t,Y_t,S_{t+1})\) and \(X_t=(S_t,A_t)\), with \(Z_{i,t}\) and
\(X_{i,t}\) denoting their values for individual \(i\). For a trajectory
function \(F\) and a one-transition function \(f\), write
\[
\begin{aligned}
\mathbb P_0F&:=E_{\mathbb P_0}\{F(O)\},
&\mathbb P_nF&:=\frac1n\sum_{i=1}^nF(O_i),\\
P_0f&:=\frac1T\sum_{t=0}^{T-1}E_{\mathbb P_0}\{f(Z_t)\},
&P_nf&:=\frac1{nT}\sum_{i=1}^n\sum_{t=0}^{T-1}f(Z_{i,t}).
\end{aligned}
\]
Thus \(P_0\) is the time-averaged one-transition marginal, hereafter the
\emph{pooled transition law}, and \(P_n\) is its empirical analogue; by
contrast, \(\mathbb P_0\) and \(\mathbb P_n\) act on whole trajectories.
More generally, a trajectory law \(\mathbb P\) induces the pooled transition
law \(P\) through
\(Pf=E_{\mathbb P}\{T^{-1}\sum_{t=0}^{T-1}f(Z_t)\}\). The state--action
marginal of \(P_0\) is denoted by \(P_{0,X}\), while the independent sampling
units are the trajectories \(O_i\).

Our goal is inference for a target stationary policy \(\pi(a\mid s)\),
assumed to admit a density with respect to a finite measure \(\nu\) on
\(\mathcal A\). Our semiparametric efficiency results are
relative to the local nonparametric trajectory model \(\mathfrak P\) defined in
Appendix~\ref{app:models}, which contains trajectory laws locally equivalent to \(\mathbb P_0\)
and satisfying bounded likelihood-ratio, policy-overlap, and
target-transition-domination conditions. For notational
convenience, a subscript zero denotes evaluation at the true law.

The observed trajectories are length-\(T\) segments of an infinite-horizon,
time-homogeneous Markov decision process
\citep{puterman1990markov}. We encode the data-generating mechanism using the
nonparametric structural equation model
\citep{pearl2012causal}
\[
\begin{aligned}
A_t &:= f_A(S_t,U_{A_t}),\\
Y_t &:= f_Y(S_t,A_t,U_{Y_t}),\\
S_{t+1} &:= f_S(S_t,A_t,Y_t,U_{S_{t+1}}),
\end{aligned}
\]
where the structural functions are unknown and the exogenous variables are
unobserved, mutually independent, and stationary over time. This model implies
that actions, outcomes, and state transitions depend on the past only through
the current state and action, and that the corresponding
conditional laws are time homogeneous. Consequently, the pooled transition law
\(P_0\) identifies the reward and transition components of the MDP, while
interventions replace the observed action mechanism by the target policy
\(\pi\). These assumptions become more plausible when the state is sufficiently
rich, for example by augmenting \(S_t\) with a fixed number of lagged values or
with low-dimensional summaries of the history \citep{van2018robust}. The NPSEM
provides a formal basis for the counterfactual MDPs used below.

\begin{example}[Policy value in an MDP]
Given a policy \(\pi\), let
\(\{S_t(\pi),A_t(\pi),Y_t(\pi):t\in\mathbb T\}\sim\mathbb P_0^\pi\)
denote the counterfactual trajectory obtained by intervening on the structural
equation for \(A_t\) so that actions are drawn from
\(\pi(\cdot\mid S_t)\). The value of the policy at time \(t\) is
\(\mathbb E_0^\pi[Y_t(\pi)]\), and its expected discounted cumulative outcome,
for discount factor \(\gamma\in[0,1)\), is
\[
\mathbb E_0^\pi\left[\sum_{t=0}^{\infty}\gamma^tY_t(\pi)\right],
\]
where \(\mathbb E_0^\pi\) denotes expectation under \(\mathbb P_0^\pi\).
\qed
\end{example}

\subsection{Inferential objective}
\label{section::prelim3}

Our goal is to perform inference on \(L^2\)-continuous linear functionals of the
\(Q\)-function, which links long-term causal effects to short-term observed
data. For a discount factor \(\gamma\in[0,1)\), the \(Q\)-function associated
with a policy \(\pi\) is
\[
q_0^\pi(s,a)
:=
\mathbb E_0^\pi\!\left[
\sum_{t=0}^\infty \gamma^t Y_t(\pi)
\,\middle|\,
S_0=s,A_0=a
\right],
\]
where \(\mathbb E_0^\pi\) denotes expectation under the counterfactual MDP
induced by policy \(\pi\) \citep{kaelbling1996reinforcement}. Throughout, since
the policy is fixed, we omit the superscript and write \(q_0\). Under the
convention \(0^0=1\), setting \(\gamma=0\) reduces \(q_0(s,a)\) to the reward
function \(r_0(s,a):=E_{P_0}(Y\mid S=s,A=a)\).

Our estimand is \(\psi_0:=E_{\mathbb P_0}\{m(S_0,q_0)\}\), where
\(q\mapsto m(S_0,q)\) is linear. A key example is the long-term policy
value \(\mathbb E_0^\pi[\sum_{t=0}^\infty \gamma^t Y_t(\pi)]\), which
corresponds to \(m(s,q)=\int q(s,a')\pi(a'\mid s)\nu(da')\). The
\(Q\)-function is identified from the pooled transition law \(P_0\) as the
solution to the Bellman equation
\citep{bellman1966dynamic,sutton1998reinforcement}
\begin{equation}
q_0(X)
=
E_{P_0}\!\left[
Y+\gamma V^\pi(q_0)(S')
\,\middle|\,
X
\right],
\label{eqn::bellman1}
\end{equation}
where \(V^\pi(q)(s'):=\int q(s',a')\pi(a'\mid s')\nu(da')\). Thus,
\(q_0(X)\) is the conditional mean of the immediate reward plus the
discounted future value under policy \(\pi\).

To develop our semiparametric theory, we view \(\psi_0\) as a linear
functional of the solution to a linear inverse problem
\citep{carrasco2007linear}. Let \(P_X\) denote the
marginal law of \(X\), and define \(\|h\|_P^2 := E_P\{h(X)^2\}\). For
each \(\mathbb P \in \mathfrak P\), write
\(r_P(X) := E_P(Y\mid X)\). Rearranging
\eqref{eqn::bellman1}, the \(Q\)-function \(q_P\) solves
\begin{equation}
\mathcal T_P(q_P) = r_P,
\qquad
\mathcal T_P(h)(X)
:=
h(X)
-
\gamma E_P\!\left[V^\pi(h)(S')\mid X\right],
\label{eqn::bellman2}
\end{equation}
where \(\mathcal T_P\) is the Bellman operator. Thus \(\psi_0=\Psi(\mathbb P_0)\),
where
\[
\Psi(\mathbb P)
:=
E_{\mathbb P}\{m(S_0,q_P)\}
\]
defines a statistical parameter over the nonparametric trajectory model \(\mathfrak P\). Setting
\(\gamma=0\) recovers the well-studied problem of inference on linear
functionals of an outcome regression
\citep{chernozhukov2018double,chernozhukov2022automatic}. A notable example is
the long-term causal effect in a randomized experiment
\citep{tran2023inferring}.

\begin{example}[Long-term causal effect in an A/B test]
\label{example::abtest}
Consider an A/B test in which individuals are randomly assigned at baseline to
an arm \(D\in\{0,1\}\), which is maintained over time. In the notation above,
let \(\widetilde S_t=(S_t,D)\) denote the augmented state, set \(A_t:=D\), and
take the fixed policy to be
\(\pi(a\mid \widetilde s=(s,d))=\ind\{a=d\}\). Writing
\(q_0(s,d)\) as shorthand for \(q_0((s,d),d)\), the Bellman equation is
\(q_0(s,d)=E_{P_0}\{Y+\gamma q_0(S',d)\mid S=s,D=d\}\). Following
\citet{tran2023inferring}, the long-term causal effect is
\(\psi_0=E_{\mathbb P_0}\{q_0(S_0,1)-q_0(S_0,0)\}\). This is a linear
functional of the augmented-state \(Q\)-function, with
\(m(s,q)=q(s,1)-q(s,0)\). \qed
\end{example}

\section{Inferential challenges from intertemporal overlap}
\label{section::overlap}

Long-term causal inference in randomized experiments and A/B tests presents an
overlap problem that randomization alone does not solve. Randomization removes
confounding in one-step treatment assignment and, by design, gives positive
probability to the experimental actions at the states observed during the
experiment. It does not ensure that the future states visited under sustained
deployment of a target policy are well represented in the data. If the
intervention shifts the state distribution over time, the target policy may
place mass on state--action regions that are rare under the experimental
data-generating process, even when each one-step treatment assignment is
randomized.

This difficulty appears directly in the nonparametric efficiency theory for
offline policy evaluation. The policy value
$E_{\mathbb P_0}\{V^\pi(q_0)(S_0)\}$, a special case of a linear functional of the
$Q$-function, can be estimated efficiently using DRL
\citep{van2018online,kallus2020double,kallus2022efficiently,
tran2023inferring}. In the fully nonparametric model, however, root-$n$
regular inference requires the unnormalized discounted state--action occupancy
ratio
\begin{equation}
d_0(a,s)
:=
\frac{
d\sum_{j=0}^\infty \gamma^j
\mathbb{P}_0^\pi\{(A_j,S_j)\in\cdot\}
}{dP_{0,X}}(a,s)
\label{eqn::stateratio}
\end{equation}
to exist and have finite variance \citep{mehrabi2024off}. When the
initial-state law is unknown, the trajectory-level efficiency bound achieved by
DRL estimators is
\[
\operatorname{Var}_{\mathbb P_0}\{V^\pi(q_0)(S_0)\}
+T^{-1}E_{P_0}\{d_0(X)^2\sigma_0^2(X)\}.
\]
The first term is absent when the
initial-state law is known or fixed. Here
$\sigma_0^2(X):=E_{P_0}[\{Y+\gamma V^\pi(q_0)(S')-q_0(X)\}^2\mid X]$ is the
conditional variance of the Bellman residual. This finite-variance requirement
is the formal expression of intertemporal overlap: the discounted
state--action visitation distribution induced by the target policy must be
sufficiently covered by the data-generating distribution. Near violations
inflate the efficiency bound through the squared occupancy ratio $d_0(X)^2$,
especially in high-dimensional settings or when interventions induce rare or
novel states
\citep{jin2021pessimism,d2021overlap,mehrabi2024off,kallus2022efficiently}.

This dependence is not removed by restricting only one primitive of the MDP. To
see this, suppose the initial distribution of $S$ is known. If $r_0$ is known
but the transition law is unrestricted, then the transition component of the
efficiency bound for the transition-law functional
$P\mapsto E_{\mathbb P_0}\{V^\pi(q_P)(S_0)\}$ is
\[
T^{-1}\operatorname{Var}_{P_0}\left[
d_0(X)\{r_0(X)+\gamma V^\pi(q_0)(S')-q_0(X)\}
\right].
\]
Conversely, if $P_0(S'\mid X)$ is known but the reward law is
unrestricted, then the reward component is
\[
T^{-1}\operatorname{Var}_{P_0}\left[d_0(X)\{Y-r_0(X)\}\right].
\]
Both identities
follow by projecting the full nonparametric influence function onto the
corresponding restricted tangent spaces, in which only the transition law or
only the reward law varies, respectively
\citep{bickel1993efficient,laan2003unified}. In either case, the bound still
depends on the full occupancy ratio $d_0$. Reducing sensitivity to
intertemporal overlap therefore requires structure on the reward--transition
combination that determines $q_0$, either explicitly, as in linear MDPs, or
implicitly, through restrictions on the $Q$-function itself. The following
example illustrates the latter approach.

\begin{example}[Semiparametric contrast restriction]
\label{example::contrast}
Continuing with the augmented-state representation in
Example~\ref{example::abtest}, let \(\bar S_t\) denote the state variables
other than the persistent environment label \(D\), and write
\(S_t=(\bar S_t,D)\). Here \(D=0\) may denote a large historical or production
environment, while \(D=1\) denotes a new intervention environment with limited
observations. A natural semiparametric restriction leaves the historical \(Q\)-function
unrestricted while restricting the between-environment contrast. For a chosen
feature map \(\phi\), one may take
\[
H
:=
\left\{
h\in L^\infty(P_{0,X}):
h((\bar s,1),a)-h((\bar s,0),a)
=
\beta^\top \phi(\bar s,a)
\text{ for some } \beta\in\mathbb R^p
\right\},
\]
where the equality is understood on the common support of \((\bar S,A)\)
across environments. This is a partially linear restriction on the
\(Q\)-function: the historical baseline \(h((\bar s,0),a)\) is unrestricted,
but the contrast across environments is finite dimensional. We return to this
data-fusion application in Appendix~\ref{appendix::datafusion}. Similar
semiparametric restrictions have been used in static settings by
\citet{kallus2018removing} and \citet{van2024adaptive}.  
\end{example}

\section{Semiparametric double reinforcement learning}
\label{section::estmeth}

This section develops the proposed semiparametric DRL methodology. We first
define weighted Bellman projection parameters, which remain well defined when
the working \(Q\)-function model is misspecified. We then construct a
model-robust autoDRL estimator for these targets, including the correction
needed under misspecification. Finally, we describe minimax \(Q\)- and
Riesz-learning criteria for estimating the required nuisance functions.

\subsection{Weighted Bellman projection parameters}
\label{section::estmeth::projection}

We first define a family of weighted Bellman projection parameters indexed by a
fixed weight function. The weight specifies the norm in which the Bellman
equation is projected onto the working semiparametric restriction. Thus, under
misspecification, different weights generally define different projection
targets. This is analogous to misspecified GMM, where the weighting matrix can
affect both the pseudo-true parameter and the estimator's large-sample
distribution \citep{hall2003large}.  
The unweighted choice \(w\equiv1\) is a natural default and corresponds to
projection in the norm induced by the behavior distribution.
Section~\ref{section::theory::asymptotic} characterizes the resulting
approximation bias under misspecification. Under correct specification, the
target is invariant to \(w\), but efficiency is not;
Section~\ref{section::estmeth::optimal} characterizes the optimal weight for
the restricted model.

Let \(H\subset L^\infty(P_{0,X})\) be a possibly infinite-dimensional linear
working semiparametric restriction for the \(Q\)-function, with \(L^2(P_{0,X})\)-closure
\(\overline H\). The model is correctly specified if \(q_0\in\overline H\).
For \(\mathbb P\in\mathfrak P\) with pooled transition law \(P\), write
\(\mathcal T_P(\overline H):=\{\mathcal T_P(h):h\in\overline H\}\) and define
\(\|f\|_{w,P}:=\{E_P[w(X)f(X)^2]\}^{1/2}\). The \emph{weighted Bellman
projection} of \(q_P\) onto \(H\) is defined by weighted Bellman-residual
minimization:
\begin{equation}
    q_{P,H,w}
    :=
    \operatorname*{arg\,min}_{q \in \overline H}
    \|r_P - \mathcal T_P(q)\|_{w,P}^2
    =
    \operatorname*{arg\,min}_{q \in \overline H}
    E_P\!\left[
        w(X)
        \{Y - \mathcal T_P(q)(X)\}^2
    \right].
\label{eqn::projection}
\end{equation}
Equivalently, \(\mathcal T_P(q_{P,H,w})
=\Pi_{\mathcal T_P(\overline H),w}r_P\), where
\(\Pi_{\mathcal T_P(\overline H),w}r_P
:=
\operatorname*{arg\,min}_{g\in\mathcal T_P(\overline H)}
\|r_P-g\|_{w,P}\) is the \(L^2_w(P)\)-projection of \(r_P\) onto the Bellman
image space \(\mathcal T_P(\overline H)\). The corresponding projection
parameter is
\[
    \Psi_{H,w}(\mathbb P)
    :=
    E_{\mathbb P}\{m(S_0,q_{P,H,w})\}.
\]
At \(\mathbb P_0\), this parameter equals the original target \(\Psi(\mathbb P_0)\) under
correct specification. Under misspecification, it remains well defined: the
same functional is evaluated at the element of \(\overline H\) whose Bellman
image best approximates \(r_P\) in the \(\|\cdot\|_{w,P}\) norm.

The projection is well defined at \(P_0\) under the following conditions. Let $\Lambda_0(B\mid x)
:=
\int \ind\{(s',a')\in B\}\pi(da'\mid s')P_0(ds'\mid X=x),$ for measurable \(B\subseteq\mathcal X\).

\begin{enumerate}[label=\textbf{(C\arabic*)}, ref=C\arabic*, resume=cond]
\item \label{cond::bounded} There exist constants \(0<c_w\le C_w<\infty\),
\(C<\infty\), \(C_\Lambda<\infty\), and \(\kappa_0<\infty\) such that:
    \begin{enumerate}[label=(\roman*), ref=C\arabic{enumi}(\roman*)]
        \item \label{cond::boundedWeightsPolicyRewards}
        \textit{(Bounded weights, target policy, and baseline rewards)}
        \(c_w\le w(X)\le C_w\) and
        \(\{d\pi(\cdot\mid S)/d\nu\}(A)+|Y|\le C\)
        \(P_0\)-almost surely.

        \item \label{cond::targetTransitionDomination}
        \textit{(One-step target-transition domination)}
         $\Lambda_0(B\mid x)\le C_\Lambda P_{0,X}(B)$
        for every measurable \(B\subseteq\mathcal X\) and
        \(P_{0,X}\)-almost every \(x\).

        \item \label{cond::targetMapContinuity}
        \textit{(Target-map continuity)}
        \(\|m(S_0,h)\|_{\mathbb P_0}\le C\|h\|_{P_0}\) for all
        \(h\in\overline H\).

        \item \label{cond::restrictedBellmanInverse}
        \textit{(Restricted Bellman inverse)}
        \(\|h\|_{P_0}\le \kappa_0\|\mathcal T_0h\|_{P_0}\) for all
        \(h\in\overline H\).
    \end{enumerate}
\end{enumerate}

Condition~\ref{cond::targetTransitionDomination} ensures that \(\mathcal T_0\)
is bounded, while Condition~\ref{cond::targetMapContinuity} ensures that
\(h\mapsto m(S_0,h)\) is bounded from \(\overline H\) into \(L^2(\mathbb P_0)\), and
therefore that \(h\mapsto E_{\mathbb P_0}\{m(S_0,h)\}\) is bounded on \(\overline H\).  For our asymptotic linearity results,
Condition~\ref{cond::targetTransitionDomination} can be relaxed to require only
that \(\mathcal T_0\) be bounded on \(\overline H\); however, we use the
stronger form for the efficiency results.
Condition~\ref{cond::restrictedBellmanInverse} ensures that \(\mathcal T_0\)
is invertible on \(\overline H\) and that the inverse problem is well posed.
Sufficient conditions are given in Appendix~\ref{appendix::condsinvertible};
see also \citet{chen2022well}. In particular, if the discounted state--action
occupancy ratio exists and is uniformly bounded, then \(\mathcal T_0\) is
invertible on \(L^2(P_{0,X})\), which implies
Condition~\ref{cond::restrictedBellmanInverse}. Moreover, when the
conditional-expectation operator entering \(\mathcal T_0\) is compact,
invertibility follows from injectivity, since \(\mathcal T_0\) is a Fredholm
operator of index zero \citep{conway1994course}.

\subsection{Model-robust automatic DRL}
\label{section::estmeth::riesz}

We construct a model-robust automatic DRL estimator for the weighted Bellman
projection parameter \(\Psi_{H,w}(\mathbb P_0)\). The key object is the Riesz
representer associated with the weighted Bellman inner product. By
Condition~\ref{cond::bounded}, the functional
\(q \mapsto E_{\mathbb P_0}\{m(S_0,q)\}\) is continuous with respect to
\(\|\mathcal T_0(q)\|_{w,P_0}\). Hence there exists a unique
\(\alpha_{0,H,w}\in \overline H\) such that, for every
\(q\in \overline H\),
\begin{equation}
\label{eqn:rieszrep}
E_{\mathbb P_0}\{m(S_0,q)\}
=
E_{P_0}\!\left[
w(X)\mathcal T_0(\alpha_{0,H,w})(X)\mathcal T_0(q)(X)
\right].
\end{equation}
The autoDRL correction uses two Bellman-image nuisances:
\[
\tau_{0,H,w}(X):=\mathcal T_0(\alpha_{0,H,w})(X),
\qquad
e_{0,H,w}(X):=\mathcal T_0(q_{0,H,w})(X)-r_0(X).
\]

The first nuisance, \(\tau_{0,H,w}\), gives a reward-weighting representation
of the projection target. Taking \(q=q_{0,H,w}\) in
\eqref{eqn:rieszrep} and using the projection normal equations gives
\[
\Psi_{H,w}(\mathbb P_0)
=
E_{\mathbb P_0}\{m(S_0,q_{0,H,w})\}
=
E_{P_0}\!\left[
w(X)\tau_{0,H,w}(X)Y
\right].
\]
In the unrestricted policy-value problem,
\(\widetilde d_{0,H,w}:=w\tau_{0,H,w}\) reduces to the discounted
state--action occupancy ratio \(d_0\). Under restrictions on \(H\), it instead
balances the Bellman moments in \eqref{eqn:rieszrep} only along directions in
\(\mathcal T_0(\overline H)\). In this sense, it can be viewed as a projected
occupancy-balancing weight \citep{wang2023projected}. The second nuisance,
\(e_{0,H,w}\), is the projection residual and enters the model-robust correction
below.

Let \(q_{n,H,w} \in H\), \(\alpha_{n,H,w} \in H\), \(\tau_{n,H,w}\), and
\(e_{n,H,w}\) estimate \(q_{0,H,w}\), \(\alpha_{0,H,w}\),
\(\tau_{0,H,w}\), and \(e_{0,H,w}\), respectively. Define the autoDRL estimator
\[
\resizebox{\linewidth}{!}{$\displaystyle
\begin{aligned}
\psi_{n,H,w}^*
&=n^{-1}\sum_{i=1}^n\Bigg[m(S_{i,0},q_{n,H,w})+T^{-1}\sum_{t=0}^{T-1}\Bigg\{
w(X_{i,t})\tau_{n,H,w}(X_{i,t})\{Y_{i,t}+\gamma V^\pi(q_{n,H,w})(S_{i,t+1})-q_{n,H,w}(X_{i,t})\}\\
&\hspace{29mm}-w(X_{i,t})e_{n,H,w}(X_{i,t})
\{\alpha_{n,H,w}(X_{i,t})-\gamma V^\pi(\alpha_{n,H,w})(S_{i,t+1})
-\tau_{n,H,w}(X_{i,t})\}\Bigg\}\Bigg].
\end{aligned}
$}
\]
The estimator augments the plug-in term with two debiasing corrections. The first is the standard DRL Bellman-residual correction, with reward weight \(w\tau_{n,H,w}\). The second is specific to weighted projection targets and uses the projection residual \(e_{n,H,w}\) to cancel the additional first-order bias that arises when \(H\) is misspecified. Under correct specification, \(q_0\in\overline H\), the residual vanishes, \(e_{0,H,w}=0\), so this correction can be omitted by setting \(e_{n,H,w}\equiv0\). A feasible optimally weighted version of the autoDRL estimator, which is efficient in the restricted model under correct specification, is introduced and analyzed in Section~\ref{section::estmeth::optimal}.

In practice, we recommend cross-fitting by trajectory to reduce reliance on Donsker-type
empirical-process conditions \citep{van2011cross,chernozhukov2018double}:
the nuisance functions are trained on auxiliary folds and evaluated on held-out
trajectories. For notational simplicity, our asymptotic theory analyzes a
single-split version, in which the nuisance estimators are trained once on an
auxiliary sample and then evaluated on an independent analysis sample.

\subsection{Minimax estimation of nuisance functions}
\label{subsection::nuisance}

The preceding estimator treats the nuisance estimates as given. We now describe
an implementation for estimating the weighted Bellman projection
\(q_{0,H,w}\), the weighted Bellman-Riesz representer \(\alpha_{0,H,w}\), and
the Bellman-image nuisances
\(\tau_{0,H,w}:=\mathcal T_0(\alpha_{0,H,w})\) and
\(e_{0,H,w}:=\mathcal T_0(q_{0,H,w})-r_0\).

A direct two-step implementation would first estimate the two \(H\)-valued
nuisances. At the population level, \(q_{0,H,w}\) minimizes, over \(q\in H\),
the Bellman risk \(q\mapsto
\frac{1}{2}\|\mathcal T_0(q)\|_{w,P_0}^2
-
E_{P_0}\{w(X)Y\mathcal T_0(q)(X)\}\), while \(\alpha_{0,H,w}\) minimizes, over
\(\alpha\in H\), the Riesz risk \(\alpha\mapsto
\frac{1}{2}\|\mathcal T_0(\alpha)\|_{w,P_0}^2
-E_{\mathbb P_0}\{m(S_0,\alpha)\}\). One could then estimate
\(\tau_{0,H,w}=\mathcal T_0(\alpha_{0,H,w})\) and
\(e_{0,H,w}=\mathcal T_0(q_{0,H,w})-r_0\) in a second step. This direct route,
however, requires estimating the conditional Bellman operator
\(\mathcal T_0(h)(X)=h(X)-\gamma E_{P_0}\{V^\pi(h)(S')\mid X\}\), as well as the
reward regression \(r_0(X)=E_{P_0}(Y\mid X)\). This direct plug-in route would therefore require nested regression and
operator-estimation steps, which can be computationally burdensome and
statistically delicate under flexible function approximation.

 We instead use a minimax formulation that identifies the required nuisances
through variational moment equations based on the pooled transitions and
baseline states. This avoids direct estimation of the conditional Bellman operator
and gives a unified implementation for both the Bellman projection and the
Bellman--Riesz representer. Related minimax criteria have been used for
\(Q\)-function estimation, discounted occupancy-ratio estimation, and nuisance
estimation in conditional moment and inverse-problem models
\citep{nachum2019dualdice,jiang2020minimax,uehara2020minimax,
imaizumi2021minimax,uehara2023offline,wang2023projected,
dikkala2020minimax,bennett2022inference,bennett2023minimax,
bennett2023source,olivas2025source}. In these settings, variational
formulations can yield finite-sample and asymptotic guarantees, while providing
an alternative to fitted value-iteration or fixed-point methods whose analysis
often relies on Bellman completeness or related approximation assumptions
\citep{munos2008finite,chen2019information,xie2020q,van2025fitted}.

The criteria used here follow from the variational representation of the
squared weighted Bellman norm \(\|\mathcal T_0(f)\|_{w,P_0}^2\) in
Appendix~\ref{appendix::variational}. Under
Condition~\ref{cond::bounded}, \((e_{0,H,w},q_{0,H,w})\) is a saddle-point
solution of
\[
\min_{e\in L^2_w(P_{0,X})}
\sup_{q\in H}
E_{P_0}\!\left[
\frac{1}{2}w(X)e(X)^2
+
w(X)e(X)\{Y-q(X)+\gamma V^\pi(q)(S')\}
\right],
\]
and \((\tau_{0,H,w},\alpha_{0,H,w})\) is a saddle-point solution of
\[
\min_{\tau\in L^2_w(P_{0,X})}
\sup_{\alpha\in H}
\left\{
E_{P_0}\!\left[
\frac{1}{2}w(X)\tau(X)^2
-
w(X)\tau(X)\{\alpha(X)-\gamma V^\pi(\alpha)(S')\}
\right]
+E_{\mathbb P_0}\{m(S_0,\alpha)\}
\right\}.
\]
These criteria depend only on the observed Bellman residuals
\(Y-q(X)+\gamma V^\pi(q)(S')\) and
\(\alpha(X)-\gamma V^\pi(\alpha)(S')\). Consequently, the two saddle-point
problems recover all four nuisance functions needed by the model-robust autoDRL
estimator without nested estimation of \(\mathcal T_0\). Unlike many uses of
related minimax criteria, where only one component of the saddle pair is
retained, both components enter the final autoDRL estimator.

The criteria above are written in dual form, with the Bellman-image nuisance in
the outer minimization and the \(H\)-valued function in the inner maximization.
This form is computationally convenient: for structured classes \(H\), the inner
maximization can often be profiled out analytically, reducing the saddle-point
problem to a single-level regularized criterion for the Bellman-image nuisance.
Thus implementation avoids alternating minimax optimization, nested estimation
of \(\mathcal T_0\), and fitted Bellman iteration. Appendix~\ref{appendix::variational} gives the equivalent primal
formulations, including the more common minimax \(Q\)-learning orientation
\citep{uehara2020minimax}. 

 When \(H\) is finite-dimensional, the unpenalized
empirical criteria reduce to constrained quadratic programs with empirical
Bellman moment constraints. For high-dimensional linear classes or RKHSs, a
quadratic penalty on the inner \(H\)-valued variable relaxes exact moment
matching into a regularized dual-norm penalty; after profiling, this yields
standard ridge-type or kernel least-squares systems
\citep{feng2019kernel,wang2023projected,olivas2025source}. Appendix~\ref{appendix::profiledminimax}
details the corresponding profiled population and empirical criteria.

\section{Efficiency theory and inference}
\label{section::theory}

This section develops the efficiency theory for weighted Bellman projection
targets and the corresponding autoDRL estimators. We first fix the projection
weight and allow the working \(Q\)-model to be misspecified. In this setting, we
derive the efficient influence function, a von Mises expansion, and asymptotic
linearity of the estimator for the projection target. We then specialize to the
correctly specified case, where the projection target coincides with the
original functional, and characterize the optimal projection weight that attains
the restricted-model efficiency bound.

\subsection{Pathwise differentiability and efficiency bounds}
\label{section::estmeth::robusttheory}

We now derive the first-order statistical behavior of the fixed-weight Bellman
projection target. Throughout this subsection, the weight \(w\) is treated as
fixed, and we work in the locally nonparametric trajectory model \(\mathfrak P\) defined in
Appendix~\ref{app:models}. The following theorem shows that
\(\Psi_{H,w}\) is pathwise differentiable, identifies its efficient influence
function, and therefore gives the semiparametric efficiency bound
\citep{bickel1993efficient,van2000asymptotic}. It also provides a von Mises expansion that will
be used to establish asymptotic linearity of the proposed estimator. For
background on semiparametric efficiency in double reinforcement learning, see
\citet{kallus2022efficiently}.

To state the efficient influence function and the associated von Mises expansion
\citep{mises1947asymptotic}, fix \(\mathbb P\in\mathfrak P\), let \(P\) be its
pooled transition law, and let \(\alpha_{\mathbb P,H,w}\in\overline H\) solve
\[
E_{\mathbb P}\{m(S_0,h)\}
=
\left\langle
\mathcal T_P(\alpha_{\mathbb P,H,w}),\mathcal T_P(h)
\right\rangle_{w,P},
\qquad h\in\overline H.
\]
Set
\(\tau_{\mathbb P,H,w}:=\mathcal T_P(\alpha_{\mathbb P,H,w})\) and
\(e_{P,H,w}:=\mathcal T_P(q_{P,H,w})-r_P\). For a trajectory
\(o=(s_0,a_0,y_0,\ldots,s_T)\), write \(x_t=(s_t,a_t)\), and define
\[
\begin{aligned}
\Phi^*_{\mathbb P,H,w}(o)
&:=
m(s_0,q_{P,H,w})-\Psi_{H,w}(\mathbb P) \\
&\quad+\frac1T\sum_{t=0}^{T-1}\Big[
w(x_t)\tau_{\mathbb P,H,w}(x_t)
\{y_t+\gamma V^\pi(q_{P,H,w})(s_{t+1})-q_{P,H,w}(x_t)\} \\
&\hspace{31mm}-w(x_t)e_{P,H,w}(x_t)
\{\alpha_{\mathbb P,H,w}(x_t)-\gamma V^\pi(\alpha_{\mathbb P,H,w})(s_{t+1})
-\tau_{\mathbb P,H,w}(x_t)\}\Big].
\end{aligned}
\]

\begin{samepage}
\begin{enumerate}[label=\textbf{(C\arabic*)}, ref=C\arabic*, resume=cond]
    \item \label{cond::boundednuis} \textit{(Uniformly bounded nuisances)} $ \sup_{\mathbb P\in\mathfrak P}
        \left\{
        \|q_{P,H,w}\|_\infty
        +\|\alpha_{\mathbb P,H,w}\|_\infty
        \right\}<\infty .$
\end{enumerate}
Condition~\ref{cond::boundednuis} requires uniform boundedness of the projected
\(Q\)-function and projected Bellman-Riesz representer over the local model. For the
policy-value target,
Appendix~\ref{appendix::policyValueBoundedNuisanceVerification} verifies this
condition for finite-dimensional semiparametric restrictions and for the unrestricted space
\(\overline H=L^2(P_{0,X})\) under standard conditions.
\end{samepage}

For \(\overline{\mathbb P}\in\mathfrak P\), let \(\overline P\) denote its
pooled transition law and define
\[
\begin{aligned}
R^*_{H,w}(\overline{\mathbb P},\mathbb P_0)
&:=
\left\langle
\tau_{\overline{\mathbb P},H,w}
-\tau_{0,H,w},
\mathcal T_0(q_{0,H,w}-q_{\overline P,H,w})
\right\rangle_{w,P_0} \\
&\quad+
\left\langle
\tau_{\overline{\mathbb P},H,w}
-\mathcal T_0(\alpha_{\overline{\mathbb P},H,w}),
e_{\overline P,H,w}-e_{0,H,w}
\right\rangle_{w,P_0}.
\end{aligned}
\]

\begin{theorem}[Pathwise differentiability]
\label{theorem::projectionPD}
Suppose Conditions~\ref{cond::bounded} and~\ref{cond::boundednuis} hold. Then the map
\(\Psi_{H,w}:\mathfrak P\to\mathbb R\) is pathwise differentiable at \(\mathbb P_0\) with efficient influence function
\(\Phi^*_{0,H,w}:=\Phi^*_{\mathbb P_0,H,w}\). Moreover, for any
\(\overline{\mathbb P}\in\mathfrak P\),
\[
\Psi_{H,w}(\overline{\mathbb P})-\Psi_{H,w}(\mathbb P_0)
=
- \mathbb P_0\Phi^*_{\overline{\mathbb P},H,w}
+
R^*_{H,w}(\overline{\mathbb P},\mathbb P_0).
\]
\end{theorem}

\noindent The theorem identifies both the efficient influence function and the
semiparametric efficiency bound for the projection target. In particular, over
the model \(\mathfrak P\), the asymptotic variance of any regular estimator of
\(\Psi_{H,w}(\mathbb P_0)\) is bounded below by
\(E_{\mathbb P_0}\{(\Phi^*_{0,H,w})^2\}\). The influence function has the same general
structure as the nonparametric DRL influence function, but includes an
additional correction for misspecification through the projection residual
\(e_{0,H,w}=\mathcal T_0(q_{0,H,w})-r_0\). When the \(Q\)-model is correctly
specified, \(q_{0,H,w}=q_0\), so the projection residual vanishes and no such
correction is needed.

For the fixed-weight projection parameter corresponding to policy value, at a
correctly specified \(P_0\), this bound is
\[
\operatorname{Var}_{\mathbb P_0}\{V^\pi(q_0)(S_0)\}
+T^{-1}E_{P_0}[\{w(X)\tau_{0,H,w}(X)\}^2\sigma_0^2(X)].
\]
Here
\(\sigma_0^2(X)\) is the conditional variance of the Bellman residual; the
first term is absent when the initial-state law is known. Relative to the
nonparametric DRL bound, the transition term replaces the squared occupancy
ratio \(d_0^2(X)\) with the squared projected occupancy weight
\(\{w(X)\tau_{0,H,w}(X)\}^2\). Thus, when this projected occupancy weight is
smaller or less variable in the variance-relevant directions, semiparametric
restrictions can mitigate the intertemporal-overlap challenges discussed in
Section~\ref{section::overlap}. This observation motivates the optimal-weighting
problem studied in Section~\ref{section::estmeth::optimal}.

\subsection{Asymptotic theory}
\label{section::theory::asymptotic}

We next use the von Mises expansion above to establish asymptotic linearity and
efficiency of the autoDRL estimator for the fixed-weight Bellman projection
parameter \(\Psi_{H,w}(\mathbb P_0)\). Let \(\Phi^*_{n,H,w}\) denote the
nuisance plug-in, oracle-centered version of \(\Phi^*_{0,H,w}\): it is
obtained by replacing
\((q_{0,H,w},\alpha_{0,H,w},\tau_{0,H,w},e_{0,H,w})\) with
\((q_{n,H,w},\alpha_{n,H,w},\tau_{n,H,w},e_{n,H,w})\), while keeping the
centering constant \(\Psi_{H,w}(\mathbb P_0)\) fixed.

\begin{enumerate}[label=\textbf{(C\arabic*)}, ref=C\arabic*, resume=cond]
\item \label{cond::projectionSampleSplit}
\textit{Sample-splitting.} The estimators $q_{n,H,w}$, $ \alpha_{n,H,w}$, $\tau_{n,H,w}$, and $e_{n,H,w}$ are trained on trajectories independent of $\{O_i\}_{i=1}^n$.

\item \label{cond::projectionIFConsistency}
\textit{Influence-function consistency.} $\|\Phi^*_{n,H,w}-\Phi^*_{0,H,w}\|_{\mathbb P_0}=o_p(1).$

\item \label{cond::projectionRates}
\textit{Nuisance product rates.} Both of the following hold:
\begin{enumerate}
    \item[(i)] $\left\|
\tau_{n,H,w}
-\tau_{0,H,w}
\right\|_{w,P_0}
\left\|
\mathcal T_0(q_{n,H,w})
-\mathcal T_0(q_{0,H,w})
\right\|_{w,P_0}
=o_p(n^{-1/2}),$
\item[(ii)] $\left\|
\mathcal T_0(\alpha_{n,H,w})
-\tau_{n,H,w}
\right\|_{w,P_0}
\left\|
e_{n,H,w}
-e_{0,H,w}
\right\|_{w,P_0}
=o_p(n^{-1/2}).$
\end{enumerate}
\end{enumerate}

\begin{theorem}[Asymptotic linearity]
\label{theorem::projectionEfficiency}
Assume Condition~\ref{cond::bounded}, \ref{cond::projectionSampleSplit},
\ref{cond::projectionIFConsistency}, and~\ref{cond::projectionRates}. Then
\[
\psi^*_{n,H,w}-\Psi_{H,w}(\mathbb P_0)
=
(\mathbb P_n-\mathbb P_0)\Phi^*_{0,H,w}
+
o_p(n^{-1/2}).
\]
Consequently, if
\(\sigma^2_{0,H,w}:=E_{\mathbb P_0}\{(\Phi^*_{0,H,w})^2\}\in(0,\infty)\), then $\sqrt n\{\psi^*_{n,H,w}-\Psi_{H,w}(\mathbb P_0)\}
\rightsquigarrow
N(0,\sigma^2_{0,H,w}).$
Moreover, under Condition~\ref{cond::boundednuis}, its asymptotic variance equals the nonparametric efficiency bound for
\(\Psi_{H,w}\) at \(\mathbb P_0\) with respect to \(\mathfrak P\).
\end{theorem}

With cross-fitting, \(\sigma^2_{0,H,w}\) is estimated by the sample variance of
the \(n\) estimated trajectory influence-function values.

Theorem~\ref{theorem::projectionEfficiency} shows that the DRL estimator has a
doubly robust structure. The first product in
Condition~\ref{cond::projectionRates} couples the errors in the \(Q\)-function
and the Bellman-Riesz image \(\tau_{0,H,w}\), generalizing the double robustness of
nonparametric DRL for policy-value estimation. The second product is specific
to projection targets and appears when the semiparametric restriction may be misspecified.
Under correct specification, \(q_0\in\overline H\), the projection residual
\(e_{0,H,w}\) is zero. In this case, if the estimator is implemented with its
empirical analogue \(e_{n,H,w}\) set to zero, the second product vanishes,
leaving only the first product-rate requirement in \ref{cond::projectionRates}.

Theorem~\ref{theorem::projectionEfficiency} gives inference for the projection
target \(\Psi_{H,w}(\mathbb P_0)\). We now quantify its approximation error relative
to the original target \(\Psi(\mathbb P_0)\), thereby identifying when inference for
the projection target also yields inference for \(\Psi(\mathbb P_0)\). Let
\(\Pi_{0,H,w}\) denote the \(L^2_w(P_{0,X})\)-projection onto
\(\mathcal T_0(\overline H)\). When it exists, let
\(\alpha_{0,w}^{\rm np}\) denote the unrestricted Bellman--Riesz representer,
defined by
\[
E_{\mathbb P_0}\{m(S_0,h)\}
=
\left\langle \mathcal T_0(\alpha_{0,w}^{\rm np}),
\mathcal T_0(h) \right\rangle_{w,P_0},
\qquad h\in L^2_w(P_{0,X}).
\]

\begin{theorem}[Projection approximation error]
\label{theorem::projectionApprox}
Suppose Condition~\ref{cond::bounded} holds and
\(\alpha_{0,w}^{\rm np}\) exists. Then
\[
\begin{aligned}
\Psi(\mathbb P_0)-\Psi_{H,w}(\mathbb P_0)
&=
\left\langle
\mathcal T_0(\alpha_{0,w}^{\rm np})
-\Pi_{0,H,w}\mathcal T_0(\alpha_{0,w}^{\rm np}),
\mathcal T_0(q_0)-\Pi_{0,H,w}\mathcal T_0(q_0)
\right\rangle_{w,P_0}.
\end{aligned}
\]
\end{theorem}

The theorem gives a double-robust characterization of the projection bias. The
projection target coincides with the original target if either the
\(Q\)-function is correctly represented, \(q_0\in\overline H\), or the
unrestricted Bellman-Riesz representer is correctly represented,
\(\alpha_{0,w}^{\rm np}\in\overline H\). Thus
\(\Psi_{H,w}(\mathbb P_0)=\Psi(\mathbb P_0)\) on the union of these two specification
conditions, and the approximation error is second order when both residuals are
small. For policy-value targets, \(w\,\mathcal T_0(\alpha_{0,w}^{\rm np})=d_0\),
where \(d_0\) is the state--action occupancy ratio. Equivalently, the second
route can be expressed as the occupancy-ratio condition
\(d_0/w\in\mathcal T_0(\overline H)\). Consequently, the model-robust autoDRL
estimator yields inference for the original target \(\Psi(\mathbb P_0)\) whenever
either route holds, and remains valid under approximate specification when the
resulting second-order projection bias is negligible on the \(n^{-1/2}\) scale.

\subsection{Correct specification and optimal weighting}
\label{section::estmeth::optimal}

The preceding results treat \(w\) as fixed. Under misspecification, the choice of
\(w\) determines the projection norm and therefore the projection target
\(\Psi_{H,w}(\mathbb P_0)\), along with its efficiency bound and approximation error. A
natural default is \(w \equiv 1\), which projects in the behavior-distribution
norm and avoids estimating additional nuisance functions.

Under correct specification, weighting plays a different role. If
\(q_0 \in \overline H\), then \(q_{0,H,w}=q_0\) and
\(\Psi_{H,w}(\mathbb P_0)=\Psi(\mathbb P_0)\) for every weight satisfying
Condition~\ref{cond::boundedWeightsPolicyRewards}. Thus, \(w\) no longer changes the estimand; it
only affects the influence function and the estimator's asymptotic variance.
This suggests choosing \(w\) to minimize the variance of the influence function.
The optimal weight is the inverse conditional variance of the Bellman residual:
\[
    w_{\rm opt}(X)
    :=
    \{\sigma_0^2(X)\}^{-1},
    \qquad
    \sigma_0^2(X)
    :=
    E_{P_0}[\{Y+\gamma V^\pi(q_0)(S')-q_0(X)\}^2 \mid X].
\]
In the static case \(\gamma=0\), this recovers the optimal
inverse-conditional-variance weight for heteroskedastic least squares
regression.

Under correct specification, let
\(\Phi^{\rm eff}_{0,H}:=\Phi^*_{0,H,w_{\rm opt}}\). The following theorem
gives efficiency under the time-homogeneous restricted model
\(\mathfrak P_H^{\rm M}\) defined in Appendix~\ref{app:models}.

\begin{enumerate}[label=\textbf{(C\arabic*)}, ref=C\arabic*, resume=cond]
\item \label{cond::restrictedModelRegularity}
\textit{(Correct specification and non-degeneracy)}
\(q_0\in \overline H\), and
\(0<c\le \sigma_0^2(X)\le C<\infty\) \(P_0\)-almost surely.
\end{enumerate}

\begin{theorem}[Restricted-model efficiency and optimal weighting]
\label{theorem::optimalWeighting}
Suppose Conditions~\ref{cond::bounded} and~\ref{cond::boundednuis} hold with
\(w=w_{\rm opt}\), and suppose
Condition~\ref{cond::restrictedModelRegularity} holds. Then
\(\Psi_{H,w_{\rm opt}}(\mathbb P_0)=\Psi(\mathbb P_0)\). Moreover,
\(\Phi^{\rm eff}_{0,H}\) is the efficient influence function for \(\Psi\) at
\(\mathbb P_0\) in the restricted model \(\mathfrak P_H^{\rm M}\).
\end{theorem}

Thus, under the restricted model \(\mathfrak P_H^{\rm M}\), the semiparametric
efficiency bound for regular estimation of the original target \(\Psi(\mathbb P_0)\) is
\(E_{\mathbb P_0}\{(\Phi^{\rm eff}_{0,H})^2\}\). The optimal weight therefore does more
than minimize the variance of a fixed-weight projection estimator: under correct
specification, the corresponding estimator attains the efficiency bound for the
original target in the restricted model. Since
\(\mathfrak P_H^{\rm M}\subseteq\mathfrak P\), this restricted-model bound is no larger
than the nonparametric efficiency bound.

This result motivates a feasible optimally weighted estimator under correct
specification. Let \(\widehat{\sigma}_n^2\) be an estimator of \(\sigma_0^2\)
and define
\(\widehat{w}_n(X):=\{\widehat{\sigma}_n^2(X)\}^{-1}\). Let \(q_{n,H,\widehat{w}_n} \in H\) and
\(\tau_{n,H,\widehat{w}_n}\) denote the corresponding weighted estimates of
\(q_0\) and \(\tau_{0,H,\widehat w_n}\). The optimally weighted autoDRL
estimator is
\[
\begin{aligned}
    \psi_{n,H,\widehat{w}_n}^{\rm opt}
    &=n^{-1}\sum_{i=1}^n\Bigg[m(S_{i,0},q_{n,H,\widehat{w}_n})\\
&\qquad +T^{-1}\sum_{t=0}^{T-1}\widehat{w}_n(X_{i,t})\tau_{n,H,\widehat{w}_n}(X_{i,t})
    \{Y_{i,t}+\gamma V^\pi(q_{n,H,\widehat{w}_n})(S_{i,t+1})
    -q_{n,H,\widehat{w}_n}(X_{i,t})\}
    \Bigg].
\end{aligned}
\]

We state the asymptotic result in a slightly more general form, allowing
\(\widehat w_n\) to converge to a limit \(w_\infty\), which need not be
\(w_{\rm opt}\). Define
\[
\begin{aligned}
\Phi^{w_\infty}_{0,H}(O)
:=
&m(S_0,q_0)-\Psi(\mathbb P_0)  \\
&+\frac1T\sum_{t=0}^{T-1}
w_\infty(X_t)\tau_{0,H,w_\infty}(X_t)
\{Y_t+\gamma V^\pi(q_0)(S_{t+1})-q_0(X_t)\},
\end{aligned}
\]
the restricted-model influence function for the limiting weight. Let \(\Phi^{\widehat w_n}_{n,H}\) denote the plug-in version obtained by replacing \((w_\infty, q_0, \tau_{0,H,w_\infty})\) with \((\widehat w_n, q_{n,H,\widehat w_n}, \tau_{n,H,\widehat w_n})\), while leaving the centering constant \(\Psi(\mathbb P_0)\) unchanged.

\begin{enumerate}[label=\textbf{(C\arabic*)}, ref=C\arabic*, resume=cond]
\item \label{cond::optweightSampleSplit}
\textit{Sample-splitting.} The estimators
\(\widehat w_n\), \(q_{n,H,\widehat w_n}\), and \(\tau_{n,H,\widehat w_n}\) are trained on trajectories independent of $\{O_i\}_{i=1}^n$.

\item \label{cond::optweightIFConsistency}
\textit{Estimated-weight influence-function consistency.} $\|\Phi^{\widehat{w}_n}_{n,H}
-\Phi^{w_\infty}_{0,H}\|_{\mathbb P_0}=o_p(1).$

\item \label{cond::optweightRates}
\textit{Estimated-weight nuisance product rate.}
\[
\left\|
\tau_{n,H,\widehat w_n}
-\tau_{0,H,\widehat w_n}
\right\|_{\widehat w_n,P_0}
\left\|
\mathcal T_0(q_{n,H,\widehat w_n})
-\mathcal T_0(q_{0,H,\widehat w_n})
\right\|_{\widehat w_n,P_0}
=o_p(n^{-1/2}).
\]
\end{enumerate}
Conditions~\ref{cond::optweightSampleSplit} and~\ref{cond::optweightIFConsistency}
are the estimated-weight analogues of
Conditions~\ref{cond::projectionSampleSplit} and~\ref{cond::projectionIFConsistency}.
Condition~\ref{cond::optweightRates} is the estimated-weight analogue of
Condition~\ref{cond::projectionRates}(i).

\begin{theorem}[Asymptotic linearity of the optimally weighted estimator]
\label{theorem::optWeightedAL}
Assume Conditions~\ref{cond::restrictedModelRegularity},
\ref{cond::optweightSampleSplit}, \ref{cond::optweightIFConsistency},
and~\ref{cond::optweightRates}. Suppose also that
Condition~\ref{cond::bounded} holds with \(w=w_\infty\), and that
Condition~\ref{cond::boundedWeightsPolicyRewards} holds with \(w=\widehat w_n\) with probability
tending to one. Then
\[
    \psi_{n,H,\widehat w_n}^{\rm opt}-\Psi(\mathbb P_0)
    =
    (\mathbb P_n-\mathbb P_0)\Phi^{w_\infty}_{0,H}
    +
    o_p(n^{-1/2}).
\]
If Theorem~\ref{theorem::optimalWeighting} holds, the estimator attains the
restricted-model efficiency bound whenever \(w_\infty\) is proportional to
\(w_{\rm opt}\) almost surely.
\end{theorem}

The theorem separates robustness from optimality. Under correct specification,
the estimand is invariant to the choice of weight: any limiting weight
\(w_\infty\) satisfying Condition~\ref{cond::boundedWeightsPolicyRewards} yields asymptotically
valid inference for \(\Psi(\mathbb P_0)\), provided the nuisance-rate and
split-sample influence-function consistency conditions hold. Thus, the variance model used to construct
\(\widehat w_n\) need not be correctly specified. Simple working variance models
can therefore be used for stabilization and may improve precision relative to
the unweighted choice \(w\equiv 1\). By
Theorem~\ref{theorem::optimalWeighting}, efficiency requires the limiting
weight to be proportional to the inverse conditional variance
\(\{\sigma_0^2(X)\}^{-1}\).

\begin{example}[Partially linear A/B contrast with an auxiliary baseline]
\label{example::abtestParametricContrast}
For this illustration, take \(T=1\). Continue with
Example~\ref{example::contrast}, and write
\(q_d(\bar s):=q_0((\bar s,d),d)\). Suppose the target initial law of
\(\bar S_0\) is known and the production baseline
\(q_b(\bar s):=q_0((\bar s,0),0)\) is treated as known to first order, as when
estimated from a much larger auxiliary production sample. Under the correctly
specified restriction
\(q_1(\bar s)-q_b(\bar s)=\beta^\top\phi(\bar s)\), the contrast is
\(\psi_\Delta=c^\top\beta\), where \(c:=E_{\mathbb P_0}\{\phi(\bar S_0)\}\). Let
\(p_1:=P_0(D=1)\), \(E_1(\cdot):=E_{P_0}(\cdot\mid D=1)\), and
\[
\xi_\beta
:=
Y+\gamma q_b(\bar S')-q_b(\bar S)
-\beta^\top\{\phi(\bar S)-\gamma\phi(\bar S')\},
\qquad
g_\phi(\bar s)
:=
E_1\{\phi(\bar S)-\gamma\phi(\bar S')\mid \bar S=\bar s\}.
\]
Write \(\sigma_1^2(\bar s):=E_1(\xi_\beta^2\mid\bar S=\bar s)\) and
\(J_1:=E_1\{g_\phi(\bar S)g_\phi(\bar S)^\top/\sigma_1^2(\bar S)\}\).
In this restricted model, the efficient influence function and efficiency
bound are
\[
\varphi_\Delta(Z)
=
\frac{\ind\{D=1\}}{p_1}
c^\top J_1^{-1}
\frac{g_\phi(\bar S)}{\sigma_1^2(\bar S)}
\xi_\beta,
\qquad
E_{P_0}\{\varphi_\Delta(Z)^2\}
=
\frac{1}{p_1}c^\top J_1^{-1}c .
\]
Thus the contrast restriction replaces the unrestricted discounted occupancy
ratio with the finite-dimensional weight \(g_\phi/\sigma_1^2\). In the
constant-contrast model \(q_1(\bar s)-q_b(\bar s)=\beta\), we have
\(g_\phi=1-\gamma\), and the bound reduces to
\(\{p_1(1-\gamma)^2E_1[\sigma_1^{-2}(\bar S)]\}^{-1}\). The corresponding
inverse-variance weighted estimator is
\[
\widehat\beta
=
\frac{
\sum_{i:D_i=1}\widehat\sigma_1^{-2}(\bar S_i)
\{Y_i+\gamma \widehat q_b(\bar S_i')-\widehat q_b(\bar S_i)\}
}{
(1-\gamma)\sum_{i:D_i=1}\widehat\sigma_1^{-2}(\bar S_i)
}.
\]
\end{example}

\clearpage
\section{Numerical experiments}
\label{section::experiments}

We study the \(T=1\) specialization of the long-term A/B-test setting in
Example~\ref{example::abtestParametricContrast}. We consider three settings:
correct working-model specification with moderate overlap, weak future-state
overlap, and working-model misspecification. These settings isolate,
respectively, the efficiency gains from optimal weighting, sensitivity to
occupancy overlap, and the change in target induced by misspecification.

The data-generating process is a finite-state Markov chain represented by four
normalized state variables: engagement, churn risk, tenure, and an overlap
coordinate. The parameter \(\beta\) shifts treatment-arm transition mass toward
states that are rare under the baseline distribution, weakening intertemporal
overlap while preserving randomized treatment assignment. We consider a
correctly specified linear treatment--control \(Q\)-contrast and a misspecified
constant contrast. The transition matrices give exact values of both the
original and Bellman-projection targets, and nuisance functions are estimated
on an independent sample. Appendix~\ref{appendix::numericalDiagnostics} gives
further design details and representative numerical summaries.

\begin{figure}[H]
\centering
\captionsetup{font=small,skip=4pt}
\includegraphics[width=0.86\linewidth]{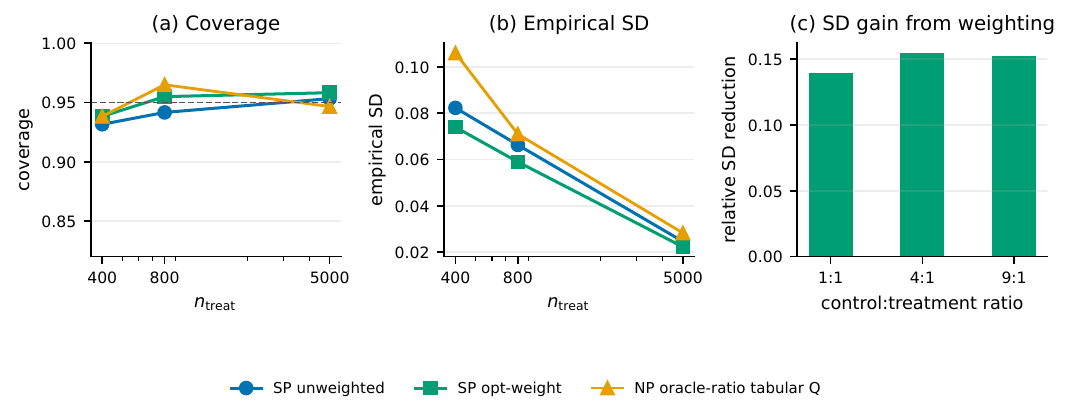}
\caption{Correct specification. Panels (a)--(b) average results at \(\beta=0\)
over control:treatment ratios; panel (c) averages results with
\(n_{\rm treat}\ge800\) and \(\beta\le0.6\) within each ratio. The oracle-ratio
estimator is an infeasible reference.}
\label{fig:correct-spec-optweight}
\end{figure}

Under correct specification and moderate overlap, the simulations recover the
efficiency gain predicted by the theory. Coverage approaches the nominal level
as the treatment-arm sample size increases. For \(n_{\rm treat}\ge800\) and
\(\beta\le0.6\), estimated optimal weighting reduces empirical standard
deviation by about 15\% on average relative to the unweighted estimator, with
similar coverage (Figure~\ref{fig:correct-spec-optweight}).

\begin{figure}[H]
\centering
\captionsetup{font=small,skip=4pt}
\includegraphics[width=0.86\linewidth]{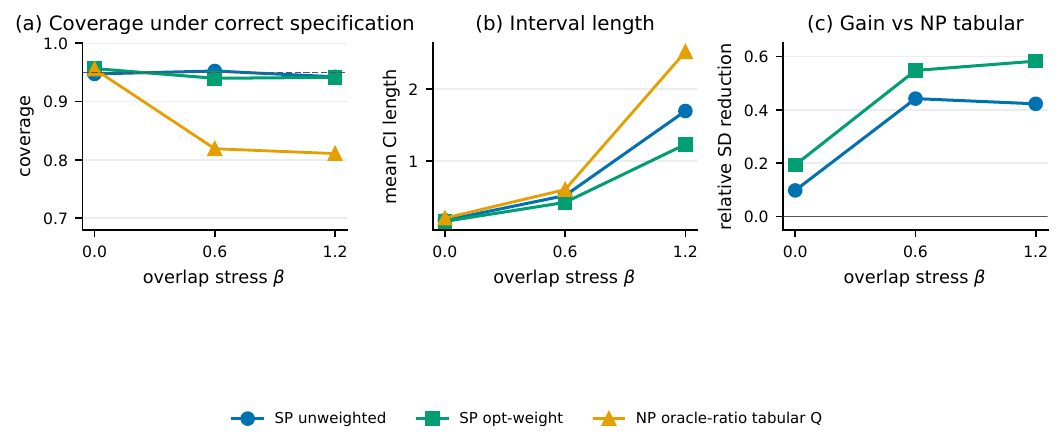}
\caption{Intertemporal overlap. Points average correctly specified results with
\(n_{\rm treat}\ge800\) over sample sizes and control:treatment ratios. In panel
(c), the relative empirical-SD reduction is computed against oracle-ratio
tabular DRL.}
\label{fig:overlap-stress}
\end{figure}

Weakening future-state overlap increases the separation between the
semiparametric and nonparametric estimators. Randomization guarantees overlap at
treatment assignment, but nonparametric DRL additionally requires overlap
between the discounted future state--action occupancy distributions. As
\(\beta\) increases, the oracle-ratio nonparametric estimator becomes more
variable and undercovers, whereas the semiparametric estimators remain
comparatively stable. Their empirical standard-deviation advantage therefore
grows as overlap weakens (Figure~\ref{fig:overlap-stress}). Because the
nonparametric estimator uses the exact occupancy ratio, this deterioration
reflects weak overlap rather than occupancy-ratio estimation error.

\begin{figure}[H]
\centering
\captionsetup{font=small,skip=4pt}
\includegraphics[width=0.86\linewidth]{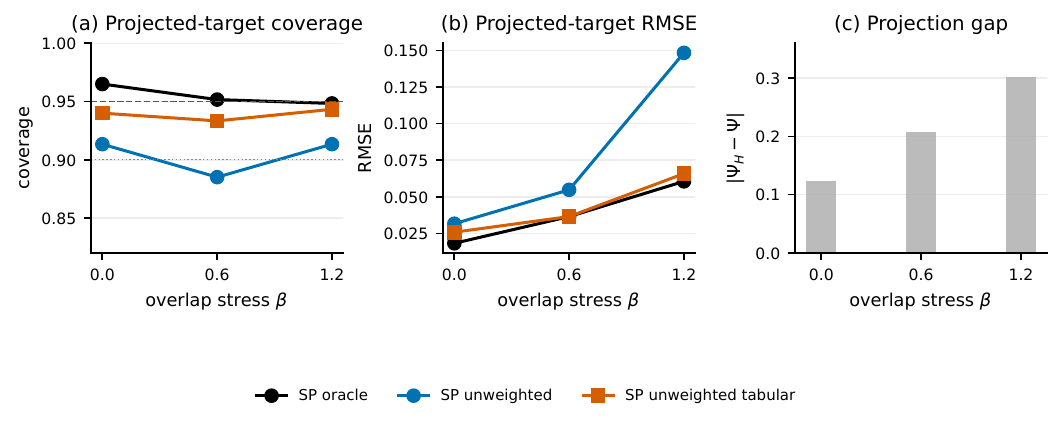}
\caption{Misspecified constant contrast. Panels average results at
\(n_{\rm treat}=5000\) over control:treatment ratios. Coverage is for the
unweighted Bellman-projection target; panel (c) shows its distance from the
original target.}
\label{fig:misspec-projection}
\end{figure}

Under misspecification, the target itself changes. The constant-contrast working
class excludes the true treatment--control \(Q\)-contrast, so the original and
Bellman-projection targets no longer coincide.
Figure~\ref{fig:misspec-projection} reports coverage for the projected target
together with the corresponding projection gap. Replacing the additive
Bellman-image nuisance fit with a tabular fit improves projected-target
coverage: mean coverage is \(0.935\) across the 18 combinations of sample size,
overlap, and allocation ratio, with a minimum of \(0.905\). The two fits use the
same estimator and target, so this comparison isolates nuisance approximation.
We omit optimally weighted estimators because, under misspecification, changing
the weight also changes the projection target.

\clearpage

\section{Empirical application}
\label{section::mybpmylife}

We apply semiparametric autoDRL to evaluate a state-dependent notification policy in the myBPmyLife microrandomized trial. Our primary estimand is the target-minus-behavior difference in infinite-horizon discounted policy value,
with discount factor $\gamma=0.95$ and initial states distributed as participants' first retained active-phase decision points. At four decisions per day, this discount factor has an effective horizon of approximately 20 decisions, or five days. We apply the stationary Markov model to the pooled decision-step distribution. For a working $Q$-class $H$, the estimator targets the corresponding Bellman-projection contrast; under correct specification of the two policy $Q$-functions, this contrast equals the discounted sustained-policy value difference.

Among 298 participants assigned to the intervention arm, 287 entered the published microrandomized-trial analytic cohort \citep{golbus2026notifications}. At four daily decision points, each participant was assigned no notification, an activity notification, or a diet notification with probabilities $0.75$, $0.125$, and $0.125$. We reconstruct 179,165 retained complete-case adjacent transitions and define the reward as the log of $0.5$ plus steps in the subsequent 60 minutes. The target policy retains the trial assignment probabilities when predecision heart-rate data are absent. Otherwise, it shifts probability from no notification to an activity notification when recent activity is low and recent activity-message burden is limited, while leaving the diet-notification probability unchanged.

We use an unpenalized linear $Q$-class with treatment-specific intercepts and coefficients. The working classes are shared across folds. We estimate all nuisance functions using the proposed single-level minimax criteria with ten-fold cross-fitting, assigning each participant's entire trajectory to one fold. We compare with flexible random-feature FQE combined with Google DualDICE. Data construction, tuning, and implementation details are given in Appendix~\ref{appendix::mybpmylife}.

\begin{table}[ht]
\centering
\caption{Discounted policy-value estimates in the myBPmyLife application}
\label{tab:mybpmylife-results}
\scriptsize
\setlength{\tabcolsep}{5pt}
\begin{tabular}{@{}lc@{}}
\toprule
Estimator / policy & Estimate [95\% CI] \\
\midrule
AutoDRL target value, linear class & $90.859\ [87.608,94.110]$ \\
AutoDRL behavior-policy value, linear class & $91.087\ [87.914,94.261]$ \\
AutoDRL target minus behavior, linear class & $-0.228\ [-0.640,0.184]$ \\
FQE + DualDICE target value & $90.529\ [87.782,93.276]$ \\
FQE + DualDICE target minus behavior & $-0.086\ [-1.525,1.353]$ \\
\bottomrule
\end{tabular}
\begin{minipage}{0.94\linewidth}
\footnotesize
\textit{Note:} Values are reported on the unnormalized discounted log-reward scale. The structured estimates are class-specific Bellman-projection functionals and equal policy values under correct specification of the relevant $Q$-functions. Confidence intervals use participant-clustered standard errors; target-minus-behavior intervals are based on paired participant scores.
\end{minipage}
\end{table}

Table~\ref{tab:mybpmylife-results} reports the policy-value estimates. The estimated target-policy value is $90.86$, compared with $91.09$ under the behavior policy. The paired difference is $-0.23$ (95\% CI $-0.64$ to $0.18$). This interval includes zero. Multiplying the contrast and its confidence limits by $1-\gamma$ and exponentiating gives a target-to-behavior ratio of discounted geometric means for steps plus $0.5$ of $0.989$ (95\% CI $0.968$ to $1.009$).

For the target-minus-behavior contrast, the estimated standard errors are $0.210$ for autoDRL and $0.734$ for FQE plus DualDICE, a 71\% reduction. The proximal analysis of \citet{golbus2026notifications} found no overall increase in 60-minute steps from activity notifications. That analysis concerns a single-decision excursion effect, whereas ours concerns the discounted value of a sustained state-dependent policy.

\section{Conclusion}

We developed weighted Bellman projection estimators for semiparametric double
reinforcement learning in time-homogeneous Markov decision processes. The
resulting projection target provides a principled estimand for inference under
working semiparametric restrictions, including when these restrictions are
misspecified. In that case, the target is no longer the original policy
functional but rather its Bellman projection analogue. The approximation error
nevertheless has a favorable second-order structure: it depends on the product
of approximation errors for the Bellman image of the \(Q\)-function and the
corresponding Bellman-Riesz image. This yields a bias--variance--interpretability
tradeoff. Richer restrictions reduce projection bias and recover the original
functional under correct specification, whereas more structured restrictions
can improve precision relative to fully nonparametric DRL.

Choosing an effective restriction in advance can be difficult. One possibility
is to select the restriction from the data, for example through variable
selection or basis selection. Appendix~\ref{sec::ADMLEmain} extends the
adaptive DML framework of \citet{van2023adaptive,van2024adaptive} to this
setting. When the selected restriction converges to a correctly specified
oracle model, the resulting estimator is asymptotically linear and can be
superefficient relative to the unrestricted nonparametric model.
These precision gains come at the cost of irregularity: the inference
guarantees are local to the oracle model and need not hold uniformly
over the full nonparametric model. Nonetheless, some degree of irregularity may
be worthwhile when the resulting gains in precision are substantial, as argued
by \citet{moosavi2023costs}.

There are several directions for future work, including finite-horizon and time-inhomogeneous settings \citep{kallus2022efficiently}, single long-trajectory data \citep{bibaut2021sequential}, undiscounted and average-reward settings
\citep{liao2022batch}, and extensions to other linear inverse problems.

 \medskip

\noindent \textbf{Data Availability Statement}
The empirical application uses public, deidentified myBPmyLife data from
\url{https://github.com/WIRED-L-um/MRT_myBPmyLife} and
\url{https://github.com/WIRED-L-um/myBPmyLife-Primary-Analysis}. The retrieval
code pins the source commits and verifies file hashes. All other numerical
results use synthetic data generated from the simulation design described in
Appendix~\ref{appendix::numericalDiagnostics}.

{
\singlespacing
\bibliographystyle{abbrvnat}
\bibliography{ref}
}

\clearpage

\appendix

\slightspacing

\setcounter{equation}{0}
\renewcommand{\theequation}{S\arabic{equation}}
\setcounter{theorem}{0}
\setcounter{figure}{0}
\setcounter{table}{0}
\setcounter{lemma}{0}
\setcounter{corollary}{0}
\renewcommand{\thetheorem}{S\arabic{theorem}}
\renewcommand{\thecorollary}{S\arabic{corollary}}
\renewcommand{\thelemma}{S\arabic{lemma}}
\renewcommand{\thefigure}{S\arabic{figure}}
\renewcommand{\thetable}{S\arabic{table}}
\renewcommand{\thealgorithm}{S\arabic{algorithm}}

\section{Simulation design and supplementary results}
\label{appendix::numericalDiagnostics}

The numerical experiments use a finite-state A/B-test design. Each state is
represented by four normalized coordinates: engagement, churn risk, tenure, and
an overlap coordinate. The baseline initial distribution has full support.
Treatment assignment controls the sample-size ratio only; it does not change
the estimand. The overlap parameter \(\beta\) changes the treatment-arm
transition matrix by moving future states toward lower-density regions of the
baseline distribution. Rewards are generated from the Bellman equation, so the
arm-specific \(Q\)-functions, arm values, full-state occupancy ratios, and
Bellman projection targets are exactly computable from the transition matrices.

Each coordinate takes values in \(\{0,0.5,1\}\), giving 81 states, and the
discount factor is \(\gamma=0.9\). The treatment-arm sample size is
\(n_{\rm treat}\in\{400,800,5000\}\); the control:treatment ratio is
\(1{:}1\), \(4{:}1\), or \(9{:}1\); and the overlap parameter is
\(\beta\in\{0,0.6,1.2\}\). Rewards include mean-zero heteroskedastic noise.
Each analysis sample has an independent nuisance-training sample with the same
arm sizes.

The correctly specified working model uses an additive spline baseline in the
state variables and an intercept-plus-linear treatment-control contrast. The
misspecified working model restricts the contrast to be constant. In the
misspecified regime, coverage is reported for the unweighted Bellman projection
parameter. True-target summaries are used only to show the projection gap.

The nuisance fits use profiled minimax Bellman criteria with ridge/Tikhonov
regularization. The optimally weighted estimators estimate the conditional Bellman-residual variance
and use inverse-variance weights. Nonparametric oracle-ratio comparisons use
the exact full-state occupancy ratio and are included as reference comparisons
rather than deployable ratio estimators. In the misspecification experiment,
the ``SP unweighted tabular'' row replaces the additive Bellman-image nuisance
basis with state indicators while holding the unweighted projection target and
estimator formula fixed.

Each main simulation setting uses 200 Monte Carlo replications; the binomial Monte Carlo
standard error is approximately 0.015 when coverage is 0.95.

The appendix also reports a deployable estimated-ratio diagnostic based on the
Google DICE-RL DualDICE implementation at \(n_{\rm treat}=800\), using 50
Monte Carlo replications per setting. This comparison illustrates the practical
difficulty of estimating discounted occupancy ratios and is not used in the
main performance claims.

\begin{table}[t]
\centering
\small
\setlength{\tabcolsep}{3.5pt}
\resizebox{\linewidth}{!}{%
\begin{tabular}{llrrrrrrrrrr}
\hline
Claim & Method & $n_{\rm treat}$ & $\beta$ & Ctrl:Trt & Target & Bias & SD & SE & Cov. & CI len. & RMSE \\
\hline
correct & SP unweighted & 5000 & 0.0 & 4.0 & true & -0.001 & 0.023 & 0.024 & 0.960 & 0.094 & 0.023 \\
correct & SP opt-weight & 5000 & 0.0 & 4.0 & true & -0.002 & 0.021 & 0.022 & 0.970 & 0.085 & 0.021 \\
correct & NP oracle-ratio tabular Q & 5000 & 0.0 & 4.0 & true & -0.001 & 0.026 & 0.028 & 0.945 & 0.109 & 0.026 \\
correct & SP unweighted & 5000 & 1.2 & 4.0 & true & -0.034 & 0.208 & 0.215 & 0.945 & 0.843 & 0.211 \\
correct & SP opt-weight & 5000 & 1.2 & 4.0 & true & -0.043 & 0.151 & 0.149 & 0.945 & 0.583 & 0.157 \\
correct & NP oracle-ratio tabular Q & 5000 & 1.2 & 4.0 & true & -0.084 & 0.379 & 0.384 & 0.875 & 1.504 & 0.388 \\
misspec projected & SP oracle & 5000 & 0.0 & 4.0 & projected & 0.002 & 0.022 & 0.023 & 0.985 & 0.089 & 0.022 \\
misspec projected & SP unweighted & 5000 & 0.0 & 4.0 & projected & -0.014 & 0.021 & 0.023 & 0.910 & 0.088 & 0.025 \\
misspec projected & SP unweighted tabular & 5000 & 0.0 & 4.0 & projected & 0.006 & 0.022 & 0.023 & 0.950 & 0.089 & 0.023 \\
misspec projected & SP oracle & 5000 & 1.2 & 4.0 & projected & -0.001 & 0.057 & 0.057 & 0.960 & 0.223 & 0.057 \\
misspec projected & SP unweighted & 5000 & 1.2 & 4.0 & projected & 0.058 & 0.091 & 0.089 & 0.880 & 0.349 & 0.108 \\
misspec projected & SP unweighted tabular & 5000 & 1.2 & 4.0 & projected & 0.011 & 0.058 & 0.056 & 0.950 & 0.219 & 0.059 \\
\hline
\end{tabular}
}
\caption{\small Numerical summary for representative A/B-test simulation settings. Correct-specification entries report true-target metrics; misspecified entries report the unweighted Bellman projection target.}
\label{tab:appendix-numerical-summary}
\end{table}

\begin{figure}[t]
\centering
\includegraphics[width=\linewidth]{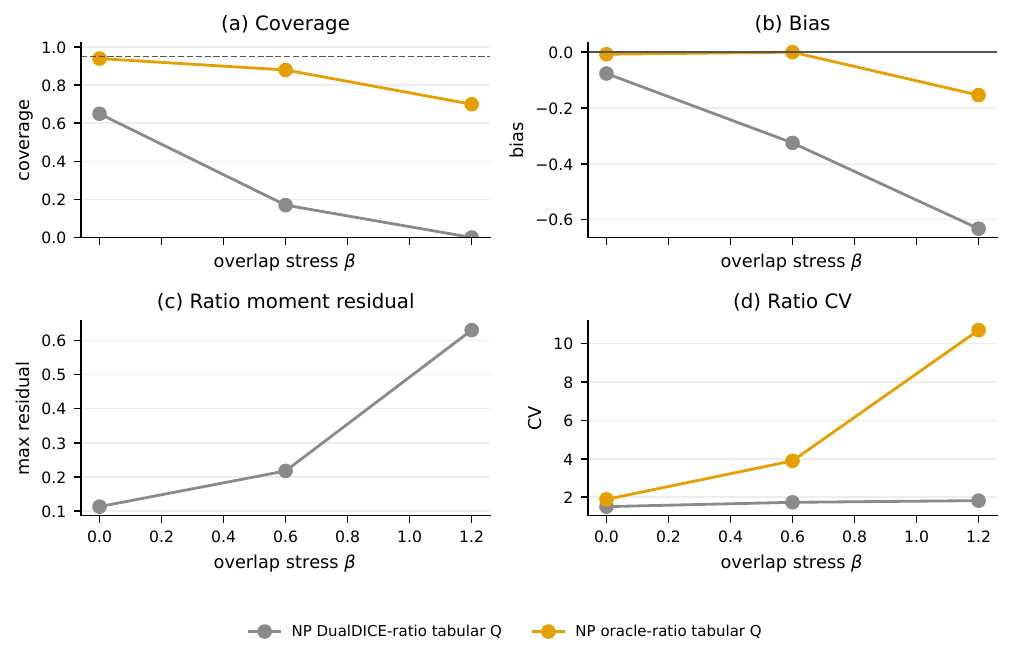}
\caption{\small DualDICE estimated-ratio nonparametric DRL comparison using
tabular \(Q\)-fits. The DualDICE row is a deployable stress baseline for
occupancy-ratio estimation; the oracle-ratio row is a nondeployable reference.
The panels show coverage, bias, empirical occupancy-ratio moment residuals,
and ratio dispersion.}
\label{fig:appendix-dualdice}
\end{figure}

\clearpage

\medskip

\section{myBPmyLife data construction and implementation}
\label{appendix::mybpmylife}

\paragraph{Analytic sample.}
We use the public myBPmyLife release associated with \citet{golbus2026notifications}. The analysis is restricted to the 287-participant microrandomized-trial cohort and the active intervention phase. For each preferred decision time, we retain the primary matched randomization record, join consecutive scheduled decisions for the same participant, and require observed predecision steps at the current and next decision together with the current 60-minute step outcome. This gives 179,165 transitions from 186,785 active-phase randomizations. The action is the randomized assignment.

The state contains cyclic time-of-day and day-of-week terms; predecision steps, heart-rate availability and count, wear time, and app activity; strictly lagged activity- and diet-notification counts over the previous four and sixteen randomized records; and baseline demographic and clinical variables. Participant identifiers and postdecision measurements are excluded. The initial distribution assigns equal weight to each participant's first retained state. The reward is $R_t=\log(0.5+\mathrm{steps}_{t,60})$, where $\mathrm{steps}_{t,60}$ is the step count in the 60 minutes after decision $t$.

\paragraph{Target policy.}
Let $x(s)$ be the predecision step count, $b_4(s)$ the number of activity-message assignments among the preceding four randomized records, and $I_{\rm HR}(s)$ the indicator that predecision heart-rate data are present. Define
\[
\ell(s)=\left\{1+\exp\!\left[\log\{1+x(s)\}-\log(101)\right]\right\}^{-1},
\qquad
u(s)=I_{\rm HR}(s)\,[\ell(s)-0.08b_4(s)]_0^1,
\]
where $[z]_0^1=\min\{1,\max(0,z)\}$. The target fixes the diet-notification probability at $0.125$ and exponentially tilts the activity-notification probability within the remaining mass:
\[
\pi_{\rm diet}(s)=0.125,\qquad
\{\pi_{\rm none}(s),\pi_{\rm activity}(s)\}
=0.875\,\frac{\{0.75,\ 0.125\exp(u(s))\}}{0.75+0.125\exp(u(s))}.
\]
Thus the target equals the trial policy when heart-rate data are absent and increases activity-message assignment smoothly when recent activity is low and recent activity-message burden is limited. Over the retained observed states, its mean assignment probabilities are 0.696, 0.179, and 0.125; one-step target-to-behavior action ratios range from 0.805 to 2.168.

\paragraph{Working classes and estimation.}
The $Q$-class is the unpenalized linear span of treatment-specific intercepts and standardized state variables. The auxiliary moment class augments this basis with piecewise-linear hinge terms $(x-t)_+$ at up to five interior empirical-quantile knots for thirteen prespecified variables and tensor products of three-knot marginal hinge bases for five prespecified interactions: recent steps with the sine and cosine of decision time and with recent activity-message burden, heart-rate availability with recent steps, and baseline steps with recent steps. All terms have treatment-specific coefficients. Imputation, centering, scaling, and knot placement use pooled state covariates and are fixed across folds; nuisance coefficients are fitted on training trajectories only.

We use ten folds, assigning each participant's entire trajectory to one fold. The $Q$-class is fixed. Separate held-out moment criteria select the projection-residual and Bellman--Riesz penalties. For computational convenience, tuning reuses the cross-fitted predictions used to construct the final estimator. The two single-level saddle systems jointly recover $(e,q)$ and $(\tau,\alpha)$ with zero inner penalty, and final estimates use only out-of-fold nuisance predictions. The selected auxiliary penalties $(\rho_e,\rho_\tau)$ are $(10^{3},10^{3})$ for the target policy and $(10^{3},10^{3})$ for the behavior reference.

The flexible comparator uses a 128-feature joint random-Fourier FQE fit, with its penalty selected by pairwise Bellman validation, and the official Google DICE-RL neural DualDICE implementation \citep{nachum2019dualdice}. The selected FQE penalty is $10^{-4}$. For each policy, DualDICE uses two 64-unit hidden layers and a common training duration across folds, selected by pooled cross-fitted Bellman balance (1,000 updates for the target policy and 2,000 for the behavior policy). Participant-level scores are the unit of inference, and the reported Wald intervals use their empirical variance.

\paragraph{Matched-budget policies.}
We additionally compare the dynamic policy with two interpretable alternatives: a uniform activity-message increase and a low-baseline rule motivated by the moderator analysis of \citet{golbus2026notifications}. The latter assigns activity-message probability $0.25$ below $7{,}408.06$ baseline steps per day and $0.110125$ above it. All three policies are calibrated to mean activity-message probability $0.1888$ over observed heart-rate-present states, retain the trial policy when heart-rate data are absent, and keep diet-message probability at $0.125$. The policies have the same mean message probability over observed states, although their long-run message use can differ. This comparison is exploratory because the low-baseline rule was motivated by an outcome analysis of the same trial.

\begin{table}[ht]
\centering
\caption{Observed-state budget-matched policy comparison}
\label{tab:mybpmylife-policy-comparison}
\scriptsize
\setlength{\tabcolsep}{4pt}
\begin{tabular}{@{}lcc@{}}
\toprule
Policy & Value [95\% CI] & Difference vs. behavior [simultaneous 95\% CI] \\
\midrule
Uniform activity increase & $90.773\ [87.503,94.042]$ & $-0.315\ [-0.968,0.338]$ \\
Prior-informed low-baseline activity & $90.881\ [87.597,94.166]$ & $-0.206\ [-0.640,0.228]$ \\
Dynamic low-recent activity & $90.859\ [87.608,94.110]$ & $-0.228\ [-0.732,0.276]$ \\
\bottomrule
\end{tabular}
\begin{minipage}{0.96\linewidth}
\footnotesize
\textit{Note:} The linear-class behavior-policy estimate is $91.087$ (95\% CI $87.914$ to $94.261$). Simultaneous intervals use a three-comparison Bonferroni critical value.
\end{minipage}
\end{table}

All three simultaneous target-minus-behavior intervals include zero. Direct target-versus-target differences range from $-0.022$ to $0.109$, and all three simultaneous intervals include zero; the data therefore do not distinguish among the matched-budget policies.

\section{Definitions of statistical models}
\label{app:models}

We work on the fixed \(P_0\)-Hilbert space \(\overline H\) defined
before~\eqref{eqn::projection}. Let \(\mathcal P\) be the local model of pooled
transition laws used in the Bellman and projection arguments: its elements are
locally equivalent to \(P_0\), have uniformly bounded likelihood ratios and
rewards, and satisfy the policy-overlap and target-transition-domination
conditions below. Thus, for a sufficiently small fixed \(\delta>0\),
\(0<c<C<\infty\), and
\(C_Y,C_g,C_\Lambda<\infty\),
\[
\mathrm H(P,P_0)\le\delta,\qquad P\sim P_0,\qquad
c\le dP/dP_0\le C\quad P_0\text{-a.s.},\qquad
|Y|\le C_Y\quad P\text{-a.s.}
\]
Let \(\mathfrak P\) be the corresponding collection of
trajectory laws \(\mathbb P\) whose pooled transition law \(P\) belongs to
\(\mathcal P\) and satisfies
\[
\mathrm H(\mathbb P,\mathbb P_0)\le\delta,\qquad
\mathbb P\sim\mathbb P_0,\qquad
c\le \frac{d\mathbb P}{d\mathbb P_0}\le C
\quad \mathbb P_0\text{-a.s.},\qquad
\max_{t<T}|Y_t|\le C_Y\quad \mathbb P\text{-a.s.},
\]
where the constants do not depend on \(\mathbb P\). For each \(P\in\mathcal P\),
the behavior policy \(g_P(a\mid s) := P(A = a \mid S = s)\) satisfies
\(d\pi(\cdot\mid s)/dg_P(\cdot\mid s)\le C_g\) uniformly in \(s\). Finally,
with
\[
    \Lambda_P(B\mid x)
    :=
    \int \ind\{(s',a')\in B\}\pi(da'\mid s')P(ds'\mid X=x),
\]
\(\Lambda_P(B\mid x)\le C_\Lambda P_{0,X}(B)\) for every measurable
\(B\subseteq\mathcal X\) and \(P_{0,X}\)-almost every \(x\).
All differentiability and stability statements below are relative to these
local models.

\begin{definition}[Time-homogeneous restricted model]
\label{def::restrictedLocalModel}
For fixed \(H\), the local restricted model is
\[
    \mathfrak P_H^{\rm M}
    :=
    \{\mathbb P\in\mathfrak P:\mathbb P\text{ is time-homogeneous Markov and }
    q_P\in\overline H\},
\]
where \(q_P\) denotes the bounded Bellman solution under \(P\). Under bounded
rewards and \(\gamma<1\), this solution exists and is unique by the usual
sup-norm contraction.
\end{definition}

\section{Verification of Condition~\ref{cond::bounded}}
\label{appendix::condsinvertible}

This section verifies Condition~\ref{cond::bounded} for the policy-value target
and gives two sufficient routes for the restricted inverse condition in
Condition~\ref{cond::restrictedBellmanInverse}. The weight bound in
Condition~\ref{cond::boundedWeightsPolicyRewards} converts unweighted bounds
into the weighted bounds used by the projection arguments:
\[
    \sqrt{c_w}\|f\|_P\le \|f\|_{w,P}\le \sqrt{C_w}\|f\|_P .
\]
Therefore, Condition~\ref{cond::restrictedBellmanInverse} implies
\[
    \|h\|_{w,P_0}
    \le
    \sqrt{\frac{C_w}{c_w}}\,
    \kappa_0\|\mathcal T_0h\|_{w,P_0},
    \qquad h\in\overline H,
\]

\subsection{Verifying Condition~\ref{cond::bounded} for the policy-value functional}
\label{appendix::policyValueConditionVerification}

We verify Condition~\ref{cond::bounded} for the policy-value target
\[
m(s_0,h)=V^\pi(h)(s_0).
\]
Throughout this subsection, Condition~\ref{cond::boundedWeightsPolicyRewards} holds, the
one-step target-transition domination in Condition~\ref{cond::targetTransitionDomination}
holds at \(P_0\), the local model \(\mathcal P\) satisfies the
behavior-policy overlap and target-transition domination requirements in
Appendix~\ref{app:models}, and \(\gamma<1\).

First, the target integrand is continuous in the pooled \(P_0\) norm. By
Jensen's inequality, behavior-policy overlap at baseline, and inclusion of
\(t=0\) in the pooled transition law,
\[
\begin{aligned}
E_{\mathbb P_0}\{V^\pi(h)(S_0)^2\}
&\le
E_{\mathbb P_0}\left[\int h(S_0,a)^2\pi(da\mid S_0)\right]  \\
&\le
CT E_{P_0}\{h(X)^2\}.
\end{aligned}
\]
Thus Condition~\ref{cond::targetMapContinuity} holds for every
\(\overline{H}\subseteq L^2(P_{0,X})\).

\begin{lemma}[Policy-value verification for finite-dimensional \(H\)]
\label{lemma::policyValueFiniteDimConditionC}
Let \(H=\{h_\theta:\theta\in\mathbb R^k\}\), where
\(h_\theta(x)=\theta^\top\phi(x)\). Assume \(\|\phi(X)\|_\infty<\infty\) and
that the matrices
\[
G_{\phi,0}:=E_{P_0}\{\phi(X)\phi(X)^\top\},
\qquad
G_{\gamma,0}:=
E_{P_0}\{\phi_{\gamma,0}(X)\phi_{\gamma,0}(X)^\top\}
\]
have eigenvalues bounded away from zero on \(\mathbb R^k\), where
\[
\phi_{\gamma,0}(X):=
\phi(X)-\gamma E_{P_0}\{V^\pi(\phi)(S')\mid X\}.
\]
Then Condition~\ref{cond::bounded} holds for the policy-value target.
\end{lemma}

\begin{proof}
Condition~\ref{cond::boundedWeightsPolicyRewards} holds by assumption,
Condition~\ref{cond::targetTransitionDomination} holds by assumption, and
Condition~\ref{cond::targetMapContinuity} was verified above. For
\(h_\theta=\theta^\top\phi\),
\[
\|h_\theta\|_{P_0}^2=\theta^\top G_{\phi,0}\theta,
\qquad
\|\mathcal T_0 h_\theta\|_{P_0}^2=\theta^\top G_{\gamma,0}\theta .
\]
The eigenvalue assumptions therefore imply Condition~\ref{cond::restrictedBellmanInverse}.
\end{proof}

\begin{lemma}[Policy-value verification for \(\overline{H}=L^2(P_{0,X})\)]
\label{lemma::policyValueL2ConditionC}
Suppose \(\overline{H}=L^2(P_{0,X})\). If the discounted-occupancy inverse
condition in Lemma~\ref{lemma::discountedBellmanInverse} holds at \(P_0\), then
Condition~\ref{cond::bounded} holds for the policy-value target.
\end{lemma}

\begin{proof}
Condition~\ref{cond::boundedWeightsPolicyRewards} holds by assumption,
Condition~\ref{cond::targetTransitionDomination} holds by assumption, and
Condition~\ref{cond::targetMapContinuity} was verified at the start of this
subsection. Condition~\ref{cond::restrictedBellmanInverse} follows from
Lemma~\ref{lemma::discountedBellmanInverse} and the equivalence of the weighted
and unweighted norms under Condition~\ref{cond::boundedWeightsPolicyRewards}.
\end{proof}

\subsection{Invertibility under bounded occupancy ratios}
\label{appendix::discountedOccupancyInverseRoute}

One route to the restricted inverse condition uses bounded discounted-occupancy
ratios; this is the route taken in
Appendix~\ref{appendix::policyValueConditionVerification} and \cite{chen2022well}. Specifically,
discounted-occupancy domination yields a bounded inverse for \(\mathcal T_P\)
on the relevant \(L^2(P_X)\)-closure. Lemma~\ref{lemma::localNormEquivalence}
identifies this closure with \(\overline H\) under the local model and
Condition~\ref{cond::boundedWeightsPolicyRewards}. Applying this argument at \(P=P_0\) verifies
Condition~\ref{cond::restrictedBellmanInverse}.

Write \(X:=(S,A)\). For \(P\in\mathcal P\), let
\(\mathrm G_{P,X}^{\pi}\) denote the unnormalized discounted state-action
occupancy measure generated by the transition law under \(P\) and target policy
\(\pi\), initialized from \(X_0\sim P_X\):
\[
    \mathrm G_{P,X}^{\pi}(C)
    :=
    \sum_{t=0}^\infty \gamma^t \Pr_P^\pi(X_t\in C).
\]
For measurable \(f\), define
\(\|f\|_{P,\gamma}^2:=\int f(x)^2\,d\mathrm G_{P,X}^{\pi}(x)\). Since the
\(t=0\) term equals \(P_X\), \(\|f\|_P\le \|f\|_{P,\gamma}\).

\begin{lemma}[Discounted-occupancy inverse for the Bellman operator]
\label{lemma::discountedBellmanInverse}
For \(P\in\mathcal P\), let \(\mathcal T_P=I-\gamma\mathcal P_P^\pi\), where
\((\mathcal P_P^\pi h)(X):=E_P[V^\pi(h)(S')\mid X]\). Then
\(\mathcal T_P\) has a bounded inverse on its range in
\(L^2(\mathrm G_{P,X}^{\pi})\), and
\[
\|h\|_{P,\gamma}
\le
\frac{1}{1-\sqrt{\gamma}}\|\mathcal T_P h\|_{P,\gamma}.
\]
If, in addition, \(\mathrm G_{P,X}^{\pi}\ll P_X\) and
\(\|d\mathrm G_{P,X}^{\pi}/dP_X\|_\infty\le B_\gamma\), then
\[
\|h\|_P
\le
\frac{\sqrt{B_\gamma}}{1-\sqrt{\gamma}}\|\mathcal T_P h\|_P .
\]
Thus \(\mathcal T_P\) is injective on \(L^2(P_X)\). On every
\(L^2(P_X)\)-closed linear subspace on which \(\mathcal T_P\) is continuous,
its range is closed and its inverse on that range is bounded.
\end{lemma}

\begin{proof}
If \(\gamma=0\), then \(\mathrm G_{P,X}^{\pi}=P_X\) and
\(\mathcal T_P=I\), so both displayed bounds hold. Assume \(\gamma>0\). Let
\(X_t=(S_t,A_t)\) be the state--action chain generated by \(P\) and \(\pi\).
For every nonnegative measurable \(f\),
\[
\int \mathcal P_P^\pi f\,d\mathrm G_{P,X}^{\pi}
=
\sum_{t=0}^\infty \gamma^t E_P^\pi[f(X_{t+1})]
\le
\gamma^{-1}\int f\,d\mathrm G_{P,X}^{\pi}.
\]
Jensen's inequality gives
\((\mathcal P_P^\pi h)^2\le \mathcal P_P^\pi(h^2)\), so
\(\|\gamma\mathcal P_P^\pi h\|_{P,\gamma}
\le\sqrt{\gamma}\|h\|_{P,\gamma}\). Hence the Neumann series
\(\sum_{j=0}^{\infty}(\gamma\mathcal P_P^\pi)^j\) converges in operator norm
on \(L^2(\mathrm G_{P,X}^{\pi})\) and equals
\((I-\gamma\mathcal P_P^\pi)^{-1}\) on the range of \(\mathcal T_P\), with
norm at most \((1-\sqrt\gamma)^{-1}\). Since the \(t=0\) term gives
\(\|h\|_P\le \|h\|_{P,\gamma}\) and the domination bound gives
\(\|\mathcal T_P h\|_{P,\gamma}\le \sqrt{B_\gamma}\|\mathcal T_P h\|_P\), the
displayed \(L^2(P_X)\) inverse bound follows. The lower bound implies that, if
\(\mathcal T_P h_n\) converges in \(L^2(P_X)\) for a sequence \(h_n\) in a
closed linear subspace, then \(h_n\) is Cauchy in \(L^2(P_X)\). Continuity of
\(\mathcal T_P\) gives closedness of the range and boundedness of the inverse
on that range.
\end{proof}

\begin{lemma}[Reduced discounted-occupancy inverse]
\label{lemma::reducedDiscountedBellmanInverse}
Let \(\phi:\mathcal A\times\mathcal S\to\mathbb R^m\) be measurable, let \(U=\phi(X)\), and let \(P_{0,\phi}\) be the law of \(U\) under \(P_0\). Define the reduced transition operator
\[
(\mathcal P_{0,\phi}^\pi f)(U)
:=
E_{P_0}\!\left[V^\pi(f\circ\phi)(S')\mid \phi(X)=U\right],
\]
and set \(\mathcal T_{0,\phi}:=I-\gamma\mathcal P_{0,\phi}^\pi\). Let \(K_{0,\phi}^{\pi}\) be the Markov kernel corresponding to \(\mathcal P_{0,\phi}^{\pi}\), and define the reduced discounted occupancy measure by
\[
\mathrm G_{0,\phi}^{\pi}(B)
:=
\sum_{t=0}^{\infty}\gamma^t
\int (K_{0,\phi}^{\pi})^t(B\mid z)\,dP_{0,\phi}(z).
\]
If \(\mathrm G_{0,\phi}^{\pi}\ll P_{0,\phi}\) and \(\|d\mathrm G_{0,\phi}^{\pi}/dP_{0,\phi}\|_\infty\le B_\phi\), then \(\mathcal T_{0,\phi}\) is a continuous bijection on \(L^2(P_{0,\phi})\), and
\[
\|\mathcal T_{0,\phi}f\|_{P_{0,\phi}}
\le
\{1+\sqrt{\gamma B_\phi}\}\|f\|_{P_{0,\phi}},
\qquad
\|f\|_{P_{0,\phi}}
\le
\frac{\sqrt{B_\phi}}{1-\sqrt{\gamma}}
\|\mathcal T_{0,\phi} f\|_{P_{0,\phi}}.
\]
\end{lemma}

\begin{proof}
The proof is the same Neumann-series argument as in Lemma~\ref{lemma::discountedBellmanInverse}, applied to the reduced Markov kernel \(\mathcal P_{0,\phi}^\pi\). The \(t=0\) term gives \(\|f\|_{P_{0,\phi}}\le \|f\|_{\mathrm G_{0,\phi}^{\pi}}\). Jensen's inequality gives \((\mathcal P_{0,\phi}^\pi f)^2\le \mathcal P_{0,\phi}^\pi(f^2)\), and hence \(\|\gamma\mathcal P_{0,\phi}^\pi f\|_{\mathrm G_{0,\phi}^{\pi}}\le\sqrt\gamma\|f\|_{\mathrm G_{0,\phi}^{\pi}}\). Thus \(I-\gamma\mathcal P_{0,\phi}^\pi\) has inverse \(\sum_{j=0}^{\infty}(\gamma\mathcal P_{0,\phi}^\pi)^j\) on \(L^2(\mathrm G_{0,\phi}^{\pi})\), with norm at most \((1-\sqrt\gamma)^{-1}\). The domination bound gives \(\|\mathcal T_{0,\phi}f\|_{\mathrm G_{0,\phi}^{\pi}}\le\sqrt{B_\phi}\|\mathcal T_{0,\phi}f\|_{P_{0,\phi}}\), yielding the inverse bound in \(L^2(P_{0,\phi})\). For the forward bound, Jensen's inequality and the one-step term in \(\mathrm G_{0,\phi}^{\pi}\) give
\[
\|\gamma\mathcal P_{0,\phi}^\pi f\|_{P_{0,\phi}}
\le
\sqrt{\gamma}\|f\|_{\mathrm G_{0,\phi}^{\pi}}
\le
\sqrt{\gamma B_\phi}\|f\|_{P_{0,\phi}},
\]
and the triangle inequality gives the displayed bound for
\(\mathcal T_{0,\phi}=I-\gamma\mathcal P_{0,\phi}^{\pi}\). Since
\(L^2(P_{0,\phi})\subseteq L^2(\mathrm G_{0,\phi}^{\pi})\) under the same
domination bound, the inverse maps every element of \(L^2(P_{0,\phi})\) to an
element of \(L^2(P_{0,\phi})\).
\end{proof}

\subsection{Invertibility from compactness of the transition operator}
\label{appendix::fredholmAlternative}

Another route to the restricted inverse condition uses compactness of the
Bellman transition operator. Let \(\lambda\) be a measure dominating the
relevant state-action laws, and define
\[
    \mathcal K_P(h)(x)
    :=
    E_P\!\left[V^\pi(h)(S')\mid X=x\right],
    \qquad
    \mathcal K_P:L^\infty(\lambda)\to L^\infty(\lambda).
\]
Then \(\mathcal T_P=I-\gamma\mathcal K_P\). When \(\mathcal K_P\) is compact,
\(\mathcal T_P\) is the identity minus a compact operator, so invertibility can
be verified by ruling out nonzero solutions of \(h=\gamma\mathcal K_P h\). The
following conditions use this argument to establish existence and, under
invertibility, uniqueness with a bounded inverse.

\begin{enumerate}[label=\textbf{(A\arabic*)}, ref=A\arabic*, series=condsuff]
    \item \textit{Compactness and Fredholm property:} \label{cond::compact}
    The operator \(\mathcal{K}_P\) is continuous on \(H\) with respect to
    \(\|\cdot\|_P\), and its unique continuous extension is compact on the
    closure \((\overline H,\|\cdot\|_P)\).
\end{enumerate}

Condition~\ref{cond::compact} can be verified from regularity of the transition
kernel. Suppose that \(\pi(\cdot\mid s')\ll g_P(\cdot\mid s')\), where
\(g_P\) is the behavior policy under \(P\), and that
\(P_{S'\mid X=x}\ll P_S\). Then \(\mathcal K_P\) has the representation
\[
\mathcal{K}_P(h)(a,s)
=
\int h(a',s')K_P(a',s'\mid a,s)\,dP_X(a',s'),
\]
where
\[
K_P(a', s' \mid a, s)
:=
\frac{d\pi(\cdot\mid s')}{dg_P(\cdot\mid s')}(a')
\frac{dP_{S'\mid X=(a,s)}}{dP_S}(s').
\]
Condition~\ref{cond::compact} holds with \(H=L^\infty(\lambda)\) when the
state and action spaces are compact subsets of \(\mathbb R^d\) and the kernel
\(K_P\) is continuous and bounded. In that case, \(K_P\) is square integrable
with respect to \(P_X\otimes P_X\), so \(\mathcal K_P\) is a
Hilbert--Schmidt operator on \(L^2(P_X)\) and is therefore compact
\citep{conway1994course}. Related conditions for nonparametric well-posedness
of the integral equation for the \(Q\)-function are given in Section~3.2 of
\citet{chen2022well}. Condition~\ref{cond::compact} also holds when \(H\) is
finite-dimensional and \(\mathcal K_P(H)\subseteq H\).

The following theorem shows that Condition~\ref{cond::compact} gives existence
of the Bellman projection. Adding the next invertibility condition gives
uniqueness.

\begin{enumerate}[label=\textbf{(A\arabic*)}, ref=A\arabic*, resume=condsuff]
    \item \textit{Invertibility:} \label{cond::invertible}
    If \(\gamma>0\), then \(\gamma^{-1}\) is not an eigenvalue of
    \(\mathcal{K}_P\) restricted to
    \((\overline H,\|\cdot\|_P)\). When \(\gamma=0\),
    \(\mathcal T_P=I\) and this condition is unnecessary.
\end{enumerate}

\begin{theorem}[Existence and uniqueness of the unweighted projection]
\label{theorem::uniqueprojection}
Assume Condition~\ref{cond::compact} holds at \(P\in\mathcal P\). Then
\(\mathcal{T}_P(\overline H)\) is a closed subspace of \(L^2(P_X)\),
and there exists \(q_{P,H,1}\in\overline H\) satisfying the unweighted
specialization of \eqref{eqn::projection}. If
Condition~\ref{cond::invertible} also holds, then
\(\mathcal{T}_P:(\overline H,\|\cdot\|_P)\to L^2(P_X)\) has a bounded
inverse on its range, and
\(q_{P,H,1}\) is the unique solution.
\end{theorem}

\begin{proof}
\noindent\textbf{Existence.}
By Condition~\ref{cond::compact}, the operator \(\mathcal K_P\) is compact on
\(\overline H\). Hence
\(\mathcal T_P=I-\gamma\mathcal K_P\) is a Fredholm operator of index zero on
the Banach space \((\overline H,\|\cdot\|_P)\) by the Riesz--Schauder
theorem \citep{conway1994course}. In particular, its range
\(\mathcal R:=\mathcal T_P(\overline H)\) is a closed subspace of
\(L^2(P_X)\).

Now write \(Y = r_P(X) + \epsilon\), where \(E_P[\epsilon \mid A, S] = 0\). Then for any \(q \in \overline H\),
\[
E_P\left[(Y - \mathcal{T}_P(q))^2\right] = E_P[\epsilon^2] + \|r_P - \mathcal{T}_P(q)\|_P^2.
\]
Hence the minimizer in \eqref{eqn::projection} corresponds to the Hilbert
projection
\[
    r_{P,H,1}:=\argmin_{r\in\mathcal R}\|r_P-r\|_P^2 .
\]
Because \(\mathcal R\) is closed, this projection exists and belongs to
\(\mathcal R\). Therefore there exists \(q_{P,H,1}\in\overline H\) such that
\(\mathcal T_P(q_{P,H,1})=r_{P,H,1}\), which proves existence for the
unweighted specialization of \eqref{eqn::projection}.

\smallskip
\noindent\textbf{Uniqueness.}
Under Condition~\ref{cond::invertible}, \(\gamma^{-1}\) is not an eigenvalue
of \(\mathcal K_P\), so
\(\mathcal T_P=I-\gamma\mathcal K_P\) has trivial null space on
\(\overline H\). Since \(\mathcal T_P\) is Fredholm of index zero
\citep{conway1994course}, its range is all of \(\overline H\), and the
bounded inverse theorem gives a bounded inverse. Hence
\(q_{P,H,1}=\mathcal T_P^{-1}(r_{P,H,1})\) is the unique solution to the
unweighted specialization of \eqref{eqn::projection}.
\end{proof}

Condition~\ref{cond::compact} makes
\(\mathcal T_P=I-\gamma\mathcal K_P\) a Fredholm operator of index zero on
\(\overline H\), so the unweighted projection is a Fredholm equation of the
second kind \citep{conway1994course}. Theorem~\ref{theorem::uniqueprojection}
gives a closed range under the unweighted norm. The bounded-weight norm
equivalence therefore makes this range closed in \(L^2_w(P_X)\), which gives
existence of \(q_{P,H,w}\) in \eqref{eqn::projection}.
Condition~\ref{cond::invertible} gives uniqueness and holds whenever
\(\gamma<\|\mathcal K_P\|_P^{-1}\). Because a compact operator has countable
nonzero spectrum with possible accumulation only at zero, uniqueness fails
for at most a countable set of discount factors.

The usual existence and uniqueness result for the \(Q\)-function \(q_P\) in
\(L^\infty(\lambda)\) follows from Banach's fixed point theorem when
\(\gamma\in(0,1)\), because \(\gamma\mathcal K_P\) is then a sup-norm
contraction. That argument does not apply directly to the Bellman projection
\(q_{P,H,1}\): the \(L^2\)-projection operator need not be a contraction in
supremum norm. Moreover, for \(q_P\) itself, the sup-norm contraction argument
does not imply uniqueness in \(L^2(P_X)\); it gives injectivity of
\(\mathcal T_P\) on the dense subspace \(L^\infty(\lambda)\).

\section{Verification of Condition~\ref{cond::boundednuis} for policy-value examples}
\label{appendix::policyValueBoundedNuisanceVerification}

This section verifies the bounded-nuisance condition used for pathwise
differentiability and efficiency in the two policy-value examples from
Appendix~\ref{appendix::policyValueConditionVerification}. Throughout this
section, Condition~\ref{cond::boundedWeightsPolicyRewards} holds, the local model
\(\mathcal P\) satisfies the behavior-policy overlap and target-transition
domination conditions in
Appendix~\ref{app:models}, and \(\gamma<1\).

\begin{lemma}[Bounded policy-value nuisances for finite-dimensional \(H\)]
\label{lemma::policyValueFiniteDimBoundedNuis}
Under the finite-dimensional assumptions of
Lemma~\ref{lemma::policyValueFiniteDimConditionC}, after possibly shrinking the
radius defining \(\mathfrak P\), Condition~\ref{cond::boundednuis} holds.
\end{lemma}

\begin{proof}
For \(\mathbb P\in\mathfrak P\), with pooled transition law \(P\), define
\[
    \phi_{\gamma,P}(X)
    :=
    \phi(X)-\gamma E_P\{V^\pi(\phi)(S')\mid X\},
    \qquad
    G^w_{\gamma,P}
    :=
    E_P\{w(X)\phi_{\gamma,P}(X)\phi_{\gamma,P}(X)^\top\}.
\]
Boundedness of \(\phi\), boundedness of the target-policy density, and
\(\gamma<1\) imply \(\sup_{P\in\mathcal P}\|\phi_{\gamma,P}\|_\infty<\infty\).
The unweighted eigenvalue assumption in
Lemma~\ref{lemma::policyValueFiniteDimConditionC} and
Condition~\ref{cond::boundedWeightsPolicyRewards} imply that the weighted
matrix \(G^w_{\gamma,0}\) is positive definite.
Lemma~\ref{lemma::primitiveHellingerPerturbations}, applied coordinatewise,
gives \(G^w_{\gamma,P}\to G^w_{\gamma,0}\) as
\(\mathrm H(P,P_0)\to0\). Shrinking the radius defining \(\mathfrak P\) if
needed, the eigenvalues of \(G^w_{\gamma,P}\) are therefore uniformly bounded
away from zero.

Writing \(q_{P,H,w}(x)=\theta_{q,P}^\top\phi(x)\), the projection coefficient
solves
\[
    G^w_{\gamma,P}\theta_{q,P}
    =
    E_P\{w(X)r_P(X)\phi_{\gamma,P}(X)\}.
\]
The right-hand side is uniformly bounded because rewards and
\(\phi_{\gamma,P}\) are uniformly bounded. Hence
\(\sup_{P\in\mathcal P}\|\theta_{q,P}\|<\infty\), and boundedness of
\(\phi\) gives \(\sup_{P\in\mathcal P}\|q_{P,H,w}\|_\infty<\infty\).

Similarly, writing
\(\alpha_{\mathbb P,H,w}(x)=\theta_{\alpha,\mathbb P}^\top\phi(x)\), the
Riesz coefficient solves
\[
    G^w_{\gamma,P}\theta_{\alpha,\mathbb P}
    =
    E_{\mathbb P}\{V^\pi(\phi)(S_0)\}.
\]
The right-hand side is uniformly bounded by boundedness of \(\phi\) and the
target-policy density. The same uniform eigenvalue bound gives
\(\sup_{\mathbb P\in\mathfrak P}\|\theta_{\alpha,\mathbb P}\|<\infty\), and therefore
\(\sup_{\mathbb P\in\mathfrak P}\|\alpha_{\mathbb P,H,w}\|_\infty<\infty\).
\end{proof}

We next record a uniform occupancy-ratio bound used for the unrestricted
working space. For \(\mathbb P\in\mathfrak P\), with pooled transition law \(P\), let
\(\mu_{\mathbb P,0}^\pi\) be the target
initial state-action law,
\[
    \mu_{\mathbb P,0}^\pi(B)
    :=
    \int \ind\{(s,a)\in B\}\pi(da\mid s)\mathbb P_{S_0}(ds),
\]
and let \(Q_P^\pi(x,B):=\Lambda_P(B\mid x)\) be the one-step target
state-action transition kernel. Define the unnormalized policy-value occupancy
measure
\[
    G_{\mathbb P,X}^{\pi,0}(B)
    :=
    \sum_{t=0}^\infty \gamma^t\mu_{\mathbb P,0}^\pi (Q_P^\pi)^t(B).
\]
When \(G_{\mathbb P,X}^{\pi,0}\ll P_X\), write
\(d_{\mathbb P}^\pi:=dG_{\mathbb P,X}^{\pi,0}/dP_X\).

\begin{lemma}[Bounded policy-value occupancy under the local model]
\label{lemma::policyValueOccupancyBound}
Under the local model conditions stated above,
\[
    \sup_{\mathbb P\in\mathfrak P}\|d_{\mathbb P}^\pi\|_\infty<\infty .
\]
\end{lemma}

\begin{proof}
Because \(\mathbb P_{S_0}\le T P_S\), behavior-policy overlap gives
\(\mu_{\mathbb P,0}^\pi(B)\le TC_g P_X(B)\) for every measurable
\(B\subseteq\mathcal X\). The target-transition domination requirement in the
definition of \(\mathcal P\) and the local likelihood-ratio bounds give, for
\(P_X\)-almost every
\(x\),
\[
    Q_P^\pi(x,B)
    \le
    C_\Lambda P_{0,X}(B)
    \le
    C_Q P_X(B),
    \qquad C_Q:=C_\Lambda/c.
\]
Therefore \(\mu_{\mathbb P,0}^\pi(Q_P^\pi)^t(B)\le C_Q P_X(B)\) for every
\(t\ge1\). Combining this bound with the \(t=0\) bound and summing the
geometric series yields
\[
    G_{\mathbb P,X}^{\pi,0}(B)
    \le
    \left\{TC_g+\frac{\gamma C_Q}{1-\gamma}\right\}P_X(B),
\]
which proves the uniform density bound.
\end{proof}

\begin{lemma}[Bounded policy-value nuisances for \(\overline H=L^2(P_{0,X})\)]
\label{lemma::policyValueL2BoundedNuis}
Under the full-space assumptions of Lemma~\ref{lemma::policyValueL2ConditionC},
Condition~\ref{cond::boundednuis} holds.
\end{lemma}

\begin{proof}
Because \(\overline H=L^2(P_{0,X})\), Lemma~\ref{lemma::localNormEquivalence}
implies \(\overline H=L^2_w(P_X)\) as sets. Thus the Bellman projection is
the true Bellman solution, so \(q_{P,H,w}=q_P\), where
\[
    q_P
    =
    \sum_{j=0}^\infty \gamma^j(\mathcal P_P^\pi)^j r_P,
    \qquad
    \mathcal P_P^\pi h(x):=E_P\{V^\pi(h)(S')\mid X=x\}.
\]
Bounded rewards and \(\gamma<1\) imply
\[
    \sup_{P\in\mathcal P}\|q_{P,H,w}\|_\infty
    \le
    \frac{\sup_{P\in\mathcal P}\|r_P\|_\infty}{1-\gamma}
    <\infty .
\]

By Lemma~\ref{lemma::policyValueOccupancyBound}, the policy-value occupancy
ratio \(d_{\mathbb P}^\pi\) is uniformly bounded. For every bounded \(h\),
the definition of \(G_{\mathbb P,X}^{\pi,0}\) gives
\[
\begin{aligned}
E_P\{d_{\mathbb P}^\pi(X)\mathcal T_Ph(X)\}
&=\sum_{t=0}^\infty\gamma^t
E_{\mathbb P}^\pi\{h(X_t)-\gamma h(X_{t+1})\}\\
&=E_{\mathbb P}\{V^\pi(h)(S_0)\},
\end{aligned}
\]
where the discounted series telescopes. Therefore
\[
    E_{\mathbb P}\{V^\pi(h)(S_0)\}
    =
    E_P\{d_{\mathbb P}^\pi(X)\mathcal T_P h(X)\}
    =
    E_P\{w(X)[d_{\mathbb P}^\pi(X)/w(X)]\mathcal T_P h(X)\}.
\]
Thus the unrestricted Bellman--Riesz image is
\(\tau_{\mathbb P}=d_{\mathbb P}^\pi/w\), which is uniformly bounded because
\(d_{\mathbb P}^\pi\) is
uniformly bounded and \(w\) is bounded away from zero. Define
\[
    \alpha_{\mathbb P}
    :=
    \sum_{j=0}^\infty \gamma^j(\mathcal P_P^\pi)^j\tau_{\mathbb P} .
\]
Since \(\mathcal P_P^\pi\) is a contraction on \(L^\infty\),
\[
    \sup_{\mathbb P\in\mathfrak P}\|\alpha_{\mathbb P}\|_\infty
    \le
    \frac{\sup_{\mathbb P\in\mathfrak P}\|\tau_{\mathbb P}\|_\infty}{1-\gamma}
    <\infty .
\]
Moreover, \(\mathcal T_P\alpha_{\mathbb P}=\tau_{\mathbb P}\), so
\(\alpha_{\mathbb P}\) is the
Bellman--Riesz representer in the full \(L^2_w\) working space. Hence
\(\alpha_{\mathbb P,H,w}=\alpha_{\mathbb P}\), and
Condition~\ref{cond::boundednuis} holds.
\end{proof}

\section{Weighted Bellman projection and optimal-weighting proofs}
\label{appendix::weightedProjectionOptimalProofs}

This section proves the weighted Bellman projection results used in
Section~\ref{section::estmeth}. Appendix~\ref{appendix::restrictedmodelEIF}
gives the restricted-model EIF calculation for the optimal-weighting theorem.
Throughout, \(X=(S,A)\), \(Z=(S,A,Y,S')\), and
\[
    \langle f,g\rangle_{w,P}:=E_P\{w(X)f(X)g(X)\},
    \qquad
    \Delta_{P,q}(Z):=Y+\gamma V^\pi(q)(S')-q(X).
\]
Whenever a weighted result is invoked, we assume the weight bound in
Condition~\ref{cond::boundedWeightsPolicyRewards}.

For probability laws \(P\) and \(Q\) dominated by a common measure \(\mu\),
write \(p=dP/d\mu\) and \(q=dQ/d\mu\), and define
\[
\mathrm H^2(P,Q)
:=
\frac12\int \{\sqrt p-\sqrt q\}^2\,d\mu .
\]
The definition is independent of the choice of dominating measure. We write
\(\mathrm H(P,Q)\) for its square root.

\subsection{Technical lemmas}

\subsubsection{Primitive analytic lemmas}

The following conditional-operator derivative is used in the
pathwise-differentiability proofs. Write \(X := (S,A)\),
\(Z := (S,A,Y,S')\), and, for each \(h \in H\),
\[
g_h(Z) := V^\pi(h)(S'), \qquad
\mu_{P,h}(X) := E_P[g_h(Z)\mid X].
\]
Thus, \(\mathcal{T}_P(h)(X)=h(X)-\gamma \mu_{P,h}(X)\).

\begin{lemma}[Bounded-tilt derivative of the conditional Bellman operator]
\label{lemma::conditionalBellmanDerivative}
Let \(P \in \mathcal{P}\), let \(\ell\in L_0^\infty(P)\), and let
\(dP_\varepsilon/dP=1+\varepsilon\ell\) for sufficiently small
\(|\varepsilon|\). For each fixed bounded \(h \in H\),
\[
\mu_{P_\varepsilon,h}(X)-\mu_{P,h}(X)
=
\varepsilon E_P\!\left[\{g_h(Z)-\mu_{P,h}(X)\}\ell(Z)\mid X\right]
+o_{L^2(P_X)}(\varepsilon).
\]
Consequently,
\[
d\mathcal{T}_P(h)(\ell)(X)
= E_P\!\left[
\{h(X)-\gamma g_h(Z)-\mathcal{T}_P(h)(X)\}\ell(Z)\mid X
\right],
\]
and, for every \(f \in L^2(P_X)\),
\[
\langle f,d\mathcal{T}_P(h)(\ell)\rangle_P
= E_P\!\left[
f(X)\{h(X)-\gamma g_h(Z)-\mathcal{T}_P(h)(X)\}\ell(Z)
\right].
\]
The same derivative formula holds for any dominated regular quadratic-mean
differentiable submodel through \(P\) with score \(\phi\in L^2_0(P)\) whose
conditional likelihood ratios satisfy
\begin{equation}
\left\|
\ell_{\varepsilon\mid X}-1-\varepsilon\{\phi-E_P(\phi\mid X)\}
\right\|_{L^2(P)}
=o(\varepsilon).
\label{eqn::conditionalLikelihoodExpansion}
\end{equation}
\end{lemma}

\begin{proof}
For the bounded linear tilt, the conditional likelihood ratio is
\[
\frac{dP_\varepsilon(\cdot\mid X)}{dP(\cdot\mid X)}(Z)
=
\frac{1+\varepsilon\ell(Z)}
{1+\varepsilon E_P\{\ell(Z)\mid X\}}.
\]
Since \(\ell\) is bounded, the denominator is bounded away from zero for
sufficiently small \(|\varepsilon|\), and the quotient expansion gives
\[
\frac{dP_\varepsilon(\cdot\mid X)}{dP(\cdot\mid X)}(Z)
=
1+\varepsilon\{\ell(Z)-E_P(\ell\mid X)\}+O(\varepsilon^2)
\]
uniformly. Because \(g_h\) is bounded,
\[
\begin{aligned}
\mu_{P_\varepsilon,h}(X)-\mu_{P,h}(X)
&=
E_P\!\left[
g_h(Z)
\left\{
\frac{dP_\varepsilon(\cdot\mid X)}{dP(\cdot\mid X)}(Z)-1
\right\}
\mid X
\right]\\
&=
\varepsilon E_P\!\left[
g_h(Z)\{\ell(Z)-E_P(\ell\mid X)\}\mid X
\right]
+o_{L^2(P_X)}(\varepsilon)\\
&=
\varepsilon E_P\!\left[
\{g_h(Z)-\mu_{P,h}(X)\}\ell(Z)\mid X
\right]
+o_{L^2(P_X)}(\varepsilon).
\end{aligned}
\]
This proves the bounded-tilt derivative of \(\mu_{P,h}\). The derivative of
\(\mathcal T_P(h)\) follows from \(\mathcal T_P(h)=h-\gamma\mu_{P,h}\), and the
inner-product identity follows by iterated expectation.

For a general dominated QMD submodel satisfying
\eqref{eqn::conditionalLikelihoodExpansion}, let
\(\ell_\varepsilon=dP_\varepsilon/dP\) and
\(\ell_{\varepsilon,X}=dP_{\varepsilon,X}/dP_X\). The conditional likelihood
ratio is
\(\ell_{\varepsilon\mid X}(Z):=\ell_\varepsilon(Z)/\ell_{\varepsilon,X}(X)\).
The conditional score in \eqref{eqn::conditionalLikelihoodExpansion} is
\[
\dot{\ell}_{Z\mid X}(Z)
= \phi(Z)-E_P\{\phi(Z)\mid X\}.
\]
Moreover,
\[
E_{P_\varepsilon}[g_h(Z)\mid X]-E_P[g_h(Z)\mid X]
=E_P[g_h(Z)\{\ell_{\varepsilon\mid X}(Z)-1\}\mid X].
\]
Equation~\eqref{eqn::conditionalLikelihoodExpansion}, together with the
boundedness of \(g_h\), makes the remainder \(o(\varepsilon)\) in
\(L^2(P_X)\). Substituting the conditional score and centering \(g_h(Z)\)
yields
\[
E_P[g_h(Z)\{\phi(Z)-E_P(\phi(Z)\mid X)\}\mid X]
=E_P[\{g_h(Z)-\mu_{P,h}(X)\}\phi(Z)\mid X].
\]
This proves the QMD extension of the same derivative formula.
\end{proof}

\begin{lemma}[Dense-score extension]
\label{lemma::denseScoreExtension}
Let \(\nu:\mathcal P\to\mathcal H\) be a parameter taking values in a
Hilbert space \(\mathcal H\). Suppose there is a bounded linear map
\(\dot\nu_0:L^2_0(P_0)\to\mathcal H\) such that, for every
\(\ell\in L^\infty_0(P_0)\), the bounded linear tilt
\(dP_\varepsilon^\ell/dP_0=1+\varepsilon\ell\) lies in \(\mathcal P\) for all
sufficiently small \(|\varepsilon|\) and satisfies
\[
\|\nu(P_\varepsilon^\ell)-\nu(P_0)-\varepsilon\dot\nu_0(\ell)\|_{\mathcal H}
=o(\varepsilon).
\]
If \(\nu\) is locally Hellinger Lipschitz at \(P_0\), that is,
\[
\|\nu(P_1)-\nu(P_2)\|_{\mathcal H}
\le C\,\mathrm H(P_1,P_2)
\]
for \(P_1,P_2\) in a Hellinger neighborhood of \(P_0\), then \(\nu\) is
pathwise differentiable at \(P_0\) along every regular QMD submodel contained
in \(\mathcal P\) with score \(\ell\in L^2_0(P_0)\), with derivative
\(\dot\nu_0(\ell)\).
\end{lemma}
\begin{proof}
This follows by a direct application of Lemma~2 of
\cite{luedtke2024one}; we give the short argument. Let
\((P_\varepsilon:\varepsilon\in(-\eta,\eta))\subset\mathcal P\) be a regular
QMD submodel through \(P_0\) with score \(\ell\). Choose
\(\ell_k\in L^\infty_0(P_0)\) such that
\(\|\ell_k-\ell\|_{P_0}\to0\), and let \(P_\varepsilon^{\ell_k}\) denote the
corresponding linear tilt through \(P_0\). The QMD expansion for
\(P_\varepsilon\) and the square-root expansion for the linear tilt imply
\[
    \limsup_{\varepsilon\to0}
    \frac{\mathrm H(P_\varepsilon,P_\varepsilon^{\ell_k})}{|\varepsilon|}
    \le C\|\ell-\ell_k\|_{P_0}.
\]
By the local Hellinger-Lipschitz bound and the bounded-tilt expansion, the
triangle inequality gives
\[
\begin{aligned}
\limsup_{\varepsilon\to0}
\frac{
\|\nu(P_\varepsilon)-\nu(P_0)-\varepsilon\dot\nu_0(\ell)\|_{\mathcal H}
}{|\varepsilon|}
&\le
C\|\ell-\ell_k\|_{P_0}
+
\|\dot\nu_0(\ell_k-\ell)\|_{\mathcal H}.
\end{aligned}
\]
The right-hand side tends to zero because \(\dot\nu_0\) is bounded and
\(\ell_k\to\ell\) in \(L^2(P_0)\).
\end{proof}

\begin{lemma}[Local norm equivalence and closure stability]
\label{lemma::localNormEquivalence}
Suppose the weight bound in Condition~\ref{cond::boundedWeightsPolicyRewards}
holds. For every \(P\in\mathcal P\), the marginal likelihood ratio satisfies
\[
c\le \frac{dP_X}{dP_{0,X}}\le C
\qquad P_{0,X}\text{-almost surely}.
\]
Consequently, for every measurable \(f\),
\[
c_w^{1/2}\|f\|_{P_0}
\le
\|f\|_{w,P_0}
\le
C_w^{1/2}\|f\|_{P_0},
\qquad
c^{1/2}\|f\|_{w,P_0}
\le
\|f\|_{w,P}
\le
C^{1/2}\|f\|_{w,P_0}.
\]
Thus the closures of \(H\) in \(L^2(P_{0,X})\), \(L^2_w(P_{0,X})\), and
\(L^2_w(P_X)\) agree as sets. In particular, \(\overline H\) is closed in
each of these weighted spaces.
Moreover, the target-transition domination condition in the definition of
\(\mathcal P\) implies
\[
\Lambda_P(B\mid x)
\le
C_\Lambda c_w^{-1}\int_B w(u)\,dP_{0,X}(u)
\]
for every measurable \(B\subseteq\mathcal X\) and \(P_{0,X}\)-almost every
\(x\).
\end{lemma}

\begin{proof}
Let \(L=dP/dP_0\). Since \(c\le L\le C\), the marginal likelihood ratio is
\(dP_X/dP_{0,X}=E_{P_0}(L\mid X)\), which satisfies the same bounds
\(P_{0,X}\)-almost surely. The weighted \(P_0\)-norm bounds follow from
\(c_w\le w\le C_w\). The comparison between the weighted \(P\)- and
\(P_0\)-norms follows from
\[
\|f\|_{w,P}^2
=
E_{P_0}\!\left[
w(X)f(X)^2\frac{dP_X}{dP_{0,X}}(X)
\right],
\]
which implies
\[
c^{1/2}\|f\|_{w,P_0}
\le
\|f\|_{w,P}
\le
C^{1/2}\|f\|_{w,P_0}.
\]
Equivalent norms have the same Cauchy sequences and therefore the same closure
of \(H\). The final domination display follows because
\(P_{0,X}(B)\le c_w^{-1}\int_B w(u)\,dP_{0,X}(u)\).
\end{proof}

\begin{lemma}[Primitive Hellinger perturbation bounds]
\label{lemma::primitiveHellingerPerturbations}
Let \(K_P(\cdot\mid x)\) denote the conditional law of \((Y,S')\) given
\(X=x\). Under Condition~\ref{cond::bounded} and the definition of
\(\mathcal P\), for all
\(P_1,P_2\in\mathcal P\) and all \(h\in \overline{H}\),
\[
\begin{aligned}
\|r_{P_1}-r_{P_2}\|_{w,P_0}
&\le C\,\mathrm H(P_1,P_2),\\
\|(\mathcal T_{P_1}-\mathcal T_{P_2})h\|_{w,P_0}
&\le C\,\mathrm H(P_1,P_2)\|h\|_{w,P_0}.
\end{aligned}
\]
\end{lemma}
\begin{proof}
Let \(\ell_j=dP_{j,X}/dP_{0,X}\), \(j=1,2\). Since
\(P_j(dz)=K_{P_j}(d(y,s')\mid x)P_{j,X}(dx)\), the Hellinger disintegration
identity \citep{le2012asymptotic} gives
\[
\mathrm H^2(P_1,P_2)
=
\mathrm H^2(P_{1,X},P_{2,X})
+
\int
\mathrm H^2\{K_{P_1}(\cdot\mid x),K_{P_2}(\cdot\mid x)\}
\{\ell_1(x)\ell_2(x)\}^{1/2}\,dP_{0,X}(x).
\]
Therefore, if \(\ell_j\ge c>0\), \(j=1,2\), then
\begin{equation}
\int \mathrm H^2\{K_{P_1}(\cdot\mid x),K_{P_2}(\cdot\mid x)\}\,
dP_{0,X}(x)
\le
c^{-1}\mathrm H^2(P_1,P_2).
\label{eqn::conditionalHellingerControl}
\end{equation}
Moreover, for any two probability measures \(Q_1,Q_2\) and any \(g\) with
finite second moments under both measures, the Hellinger inequality
\begin{equation}
\left|\int g\,d(Q_1-Q_2)\right|
\le
C\,\mathrm H(Q_1,Q_2)
\left\{\int g^2\,dQ_1+\int g^2\,dQ_2\right\}^{1/2}
\label{eqn::hellingerExpectationIneq}
\end{equation}
follows by writing \(dQ_j=q_j\,d\nu\) and applying Cauchy--Schwarz to
\[
\int g(q_1-q_2)\,d\nu
=
\int g(\sqrt{q_1}-\sqrt{q_2})(\sqrt{q_1}+\sqrt{q_2})\,d\nu .
\]
Applying \eqref{eqn::hellingerExpectationIneq} conditionally with
\(g(Y,S')=Y\), using the reward bound in the definition of \(\mathcal P\),
and then using \eqref{eqn::conditionalHellingerControl}, gives
\[
\|r_{P_1}-r_{P_2}\|_{w,P_0}\le C\,\mathrm H(P_1,P_2).
\]
Before applying the same argument with \(g(Y,S')=V^\pi(h)(S')\), note that
the target-transition domination requirement in the definition of
\(\mathcal P\) gives
\[
E_P\{V^\pi(h)(S')^2\mid X=x\}
\le
\int h(u)^2\,\Lambda_P(du\mid x)
\le
C\|h\|_{w,P_0}^2,
\]
where the first inequality is Jensen's inequality applied to
\(V^\pi(h)(s')=\int h(s',a')\pi(da'\mid s')\). Therefore
\eqref{eqn::hellingerExpectationIneq}, \eqref{eqn::conditionalHellingerControl},
and the weight bound in Condition~\ref{cond::boundedWeightsPolicyRewards} give
\[
\begin{aligned}
\|(\mathcal T_{P_1}-\mathcal T_{P_2})h\|_{w,P_0}
&=
\gamma\left\|E_{P_1}\{V^\pi(h)(S')\mid X\}
-E_{P_2}\{V^\pi(h)(S')\mid X\}\right\|_{w,P_0}\\
&\le C\,\mathrm H(P_1,P_2)\|h\|_{w,P_0}.
\end{aligned}
\]
\end{proof}

\begin{lemma}[Baseline-functional perturbation]
\label{lemma::baselineFunctionalPerturbation}
Suppose Condition~\ref{cond::targetMapContinuity} holds.
For \(\mathbb P_1,\mathbb P_2\in\mathfrak P\) and \(h\in\overline H\),
\[
\left|E_{\mathbb P_1}\{m(S_0,h)\}-E_{\mathbb P_2}\{m(S_0,h)\}\right|
\le C\,\mathrm H(\mathbb P_1,\mathbb P_2)\|h\|_{w,P_0}.
\]
\end{lemma}

\begin{proof}
Apply \eqref{eqn::hellingerExpectationIneq} with
\(g(O)=m(S_0,h)\), then use
Condition~\ref{cond::targetMapContinuity} and the bounded likelihood ratios
in the definition of \(\mathfrak P\).
\end{proof}

\begin{lemma}[Local inverse persistence and coercivity]
\label{lemma::localInversePersistence}
Suppose Condition~\ref{cond::bounded} holds.
After possibly reducing the Hellinger radius defining \(\mathcal P\), there
are constants \(\kappa<\infty\) and \(c_a>0\) such that, for every
\(P\in\mathcal P\),
\[
\|h\|_{w,P_0}
\le
\kappa\|\mathcal T_P h\|_{w,P_0},
\qquad h\in \overline{H},
\]
the operators are uniformly bounded,
\[
\|\mathcal T_P h\|_{w,P_0}
\le C\|h\|_{w,P_0},
\qquad h\in \overline{H},
\]
the range \(\mathcal T_P(\overline{H})\) is closed in \(L^2_w(P_{0,X})\), and
\[
E_P\{w(X)\mathcal T_P h(X)^2\}
\ge
c_a\|h\|_{w,P_0}^2,
\qquad h\in \overline{H} .
\]
\end{lemma}

\begin{proof}
By Lemma~\ref{lemma::primitiveHellingerPerturbations} with \(P_1=P\) and
\(P_2=P_0\),
\[
\|(\mathcal T_P-\mathcal T_0)h\|_{w,P_0}
\le
C\,\mathrm H(P,P_0)\|h\|_{w,P_0}.
\]
Condition~\ref{cond::restrictedBellmanInverse} and the weight bounds in
Condition~\ref{cond::boundedWeightsPolicyRewards} imply
\[
\|h\|_{w,P_0}
\le
\kappa_w\|\mathcal T_0h\|_{w,P_0},
\qquad
\kappa_w:=\sqrt{C_w/c_w}\,\kappa_0 .
\]
Shrink \(\delta\) so that \(C\delta\le (2\kappa_w)^{-1}\). Then
\[
\begin{aligned}
\|\mathcal T_Ph\|_{w,P_0}
&\ge
\|\mathcal T_0h\|_{w,P_0}
-\|(\mathcal T_P-\mathcal T_0)h\|_{w,P_0} \\
&\ge
(2\kappa_w)^{-1}\|h\|_{w,P_0}.
\end{aligned}
\]
This proves the displayed inverse bound with \(\kappa=2\kappa_w\). The operator
\(\mathcal T_P:\overline{H}\to L^2_w(P_{0,X})\) is bounded because
\(\mathcal T_0\) is bounded under
Conditions~\ref{cond::boundedWeightsPolicyRewards}
and~\ref{cond::targetTransitionDomination}
and \(\mathcal T_P-\mathcal T_0\) is bounded by
Lemma~\ref{lemma::primitiveHellingerPerturbations}. A bounded operator that is
bounded below has closed range. Finally,
Lemma~\ref{lemma::localNormEquivalence} gives
\[
E_P\{w(X)\mathcal T_P h(X)^2\}
=\|\mathcal T_P h\|_{w,P}^2
\ge
c\,\|\mathcal T_P h\|_{w,P_0}^2
\ge
c(2\kappa_w)^{-2}\|h\|_{w,P_0}^2,
\]
which proves coercivity.
\end{proof}

\begin{corollary}[Bounded influence-function envelopes]
\label{corollary::boundedIFEnvelopes}
Under Conditions~\ref{cond::bounded} and~\ref{cond::boundednuis}, for
\(\mathbb P\in\mathfrak P\) with pooled transition law \(P\), let
\[
q_P:=q_{P,H,w},\qquad
\alpha_{\mathbb P}:=\alpha_{\mathbb P,H,w},\qquad
\tau_{\mathbb P}:=\mathcal T_P(\alpha_{\mathbb P}),\qquad
e_P:=\mathcal T_P(q_P)-r_P.
\]
Then the bounds
\[
\sup_{\mathbb P\in\mathfrak P}
\left\{
\|q_P\|_\infty+\|\alpha_{\mathbb P}\|_\infty
+\|\tau_{\mathbb P}\|_\infty+\|e_P\|_\infty
\right\}<\infty
\]
hold. Moreover, with
\[
\Delta_P(q_P):=Y+\gamma V^\pi(q_P)(S')-q_P(X),
\qquad
B_{\mathbb P}(\alpha_{\mathbb P})
:=\alpha_{\mathbb P}(X)-\gamma V^\pi(\alpha_{\mathbb P})(S')
-\tau_{\mathbb P}(X),
\]
\[
\sup_{\mathbb P\in\mathfrak P}
\left\{
\|\Delta_P(q_P)\|_\infty+\|B_{\mathbb P}(\alpha_{\mathbb P})\|_\infty
+\|w\tau_{\mathbb P}\Delta_P(q_P)\|_{P_0}
+\|w e_PB_{\mathbb P}(\alpha_{\mathbb P})\|_{P_0}
\right\}<\infty .
\]
\end{corollary}

\begin{proof}
Set \(U_{P,q}:=\mathcal T_P(q_P)\) and
\(U_{\mathbb P,\alpha}:=\tau_{\mathbb P}
=\mathcal T_P(\alpha_{\mathbb P})\). By
Condition~\ref{cond::boundednuis},
\[
\sup_{\mathbb P\in\mathfrak P}\{
\|q_P\|_\infty+\|\alpha_{\mathbb P}\|_\infty
\}<\infty .
\]
For any bounded \(h\), \(\|V^\pi(h)\|_\infty\le\|h\|_\infty\), and hence
\[
\|\mathcal T_P h\|_\infty
\le
(1+\gamma)\|h\|_\infty .
\]
Applying this bound with \(h=q_P\) and \(h=\alpha_{\mathbb P}\) gives uniform
supremum-norm bounds for \(U_{P,q}\) and
\(U_{\mathbb P,\alpha}=\tau_{\mathbb P}\). The reward
bound in the definition of \(\mathcal P\) gives
\(\sup_P\|r_P\|_\infty<\infty\);
hence
\(\|e_P\|_\infty\le \|U_{P,q}\|_\infty+\|r_P\|_\infty\) is uniformly bounded.
Since \(|V^\pi(h)(s)|\le \|h\|_\infty\),
\[
\|\Delta_P(q_P)\|_\infty
\le
\|Y\|_\infty+\gamma\|q_P\|_\infty+\|q_P\|_\infty
\]
and
\[
\|B_{\mathbb P}(\alpha_{\mathbb P})\|_\infty
\le
\|\alpha_{\mathbb P}\|_\infty+\gamma\|\alpha_{\mathbb P}\|_\infty
+\|\tau_{\mathbb P}\|_\infty
\]
are uniformly bounded. Because weights are uniformly bounded under
Condition~\ref{cond::boundedWeightsPolicyRewards} and \(P_0\) is a probability
law, the two displayed \(L^2(P_0)\) product bounds follow from the
corresponding supremum-norm bounds.
\end{proof}

\subsubsection{Projection identities and stability}

\begin{lemma}[Weighted Bellman projection and Riesz identities]
\label{lemma::weightedProjectionIdentities}
Suppose Condition~\ref{cond::bounded} holds, \(\mathbb P\in\mathfrak P\),
and \(P\) is its pooled transition law. Then
\(q_{P,H,w}\) exists and is unique in \(\overline{H}\). It is
equivalently characterized by the normal equations
\[
\langle r_P-\mathcal T_P(q_{P,H,w}),\mathcal T_P(h)\rangle_{w,P}=0,
\qquad h\in \overline{H}.
\]
Moreover, there is a unique \(\alpha_{\mathbb P,H,w}\in \overline{H}\) such that
\[
E_{\mathbb P}\{m(S_0,h)\}
=
\langle \mathcal T_P(\alpha_{\mathbb P,H,w}),\mathcal T_P(h)\rangle_{w,P},
\qquad h\in \overline{H},
\]
and
\[
\Psi_{H,w}(\mathbb P)
=E_{\mathbb P}\{m(S_0,q_{P,H,w})\}
=E_P\{w(X)\mathcal T_P(\alpha_{\mathbb P,H,w})(X)Y\}.
\]
Equivalently, with
\[
\tau_{\mathbb P,H,w}:=\mathcal T_P(\alpha_{\mathbb P,H,w}),
\qquad
e_{P,H,w}:=\mathcal T_P(q_{P,H,w})-r_P,
\]
the target identity is
\[
\Psi_{H,w}(\mathbb P)=E_P\{w(X)\tau_{\mathbb P,H,w}(X)Y\}.
\]
\end{lemma}

\begin{proof}
By Lemma~\ref{lemma::localInversePersistence}, the range
\(\mathcal R_{P,H,w}:=\mathcal T_P(\overline{H})\) is a closed linear
subspace of \(L^2_w(P_{0,X})\). Lemma~\ref{lemma::localNormEquivalence} makes
the \(L^2_w(P_{0,X})\) and \(L^2_w(P_X)\) norms equivalent on this space, so
\(\mathcal R_{P,H,w}\) is also closed in \(L^2_w(P_X)\). Hence \(r_P\) has a
unique \(L^2_w(P_X)\)-orthogonal projection \(r_{P,H,w}\) onto
\(\mathcal R_{P,H,w}\). The restricted inverse bound in
Lemma~\ref{lemma::localInversePersistence} makes \(\mathcal T_P\) one-to-one on
\(\overline{H}\), so the unique element satisfying
\(\mathcal T_P(q_{P,H,w})=r_{P,H,w}\) is the minimizer in
\eqref{eqn::projection}. Orthogonality of Hilbert projections gives the displayed
normal equations.

Condition~\ref{cond::targetMapContinuity},
Lemma~\ref{lemma::localNormEquivalence}, and
Lemma~\ref{lemma::localInversePersistence} imply that
\(h\mapsto E_{\mathbb P}\{m(S_0,h)\}\) is continuous under the Hilbert norm
\(h\mapsto\|\mathcal T_P(h)\|_{w,P_0}\). Lemma~\ref{lemma::localNormEquivalence}
makes this norm equivalent to \(h\mapsto\|\mathcal T_P(h)\|_{w,P}\). The Riesz
representation theorem on
\(\mathcal R_{P,H,w}\) gives a unique representer
\(\mathcal T_P(\alpha_{\mathbb P,H,w})\in\mathcal R_{P,H,w}\), and the restricted inverse
identifies \(\alpha_{\mathbb P,H,w}\). Taking \(h=q_{P,H,w}\) and using the normal
equation with \(h=\alpha_{\mathbb P,H,w}\) gives
\[
E_{\mathbb P}\{m(S_0,q_{P,H,w})\}
=\langle \mathcal T_P(\alpha_{\mathbb P,H,w}),\mathcal T_P(q_{P,H,w})\rangle_{w,P}
=\langle \mathcal T_P(\alpha_{\mathbb P,H,w}),r_P\rangle_{w,P}.
\]
The last inner product equals
\(E_P\{w(X)\mathcal T_P(\alpha_{\mathbb P,H,w})(X)Y\}\) by iterated expectation. This
proves the identities.
\end{proof}

\begin{lemma}[Local Hellinger stability of weighted Bellman projection nuisances]
\label{lemma::weightedProjectionHellingerStability}
Suppose Conditions~\ref{cond::bounded} and~\ref{cond::boundednuis} hold.
For \(\mathbb P\in\mathfrak P\) with pooled transition law \(P\), write
\[
q_P:=q_{P,H,w},\qquad
\alpha_{\mathbb P}:=\alpha_{\mathbb P,H,w},\qquad
\tau_{\mathbb P}:=\mathcal T_P(\alpha_{\mathbb P}),\qquad
e_P:=\mathcal T_P(q_P)-r_P.
\]
Then, for \(\mathbb P_1,\mathbb P_2\in\mathfrak P\) sufficiently close to
\(\mathbb P_0\), with pooled transition laws \(P_1,P_2\),
\[
\|q_{P_1}-q_{P_2}\|_{w,P_0}
+\|\alpha_{\mathbb P_1}-\alpha_{\mathbb P_2}\|_{w,P_0}
+\|\tau_{\mathbb P_1}-\tau_{\mathbb P_2}\|_{w,P_0}
+\|e_{P_1}-e_{P_2}\|_{w,P_0}
\le C\,\mathrm H(\mathbb P_1,\mathbb P_2).
\]
Consequently,
\[
\|\Phi^*_{\mathbb P_1,H,w}-\Phi^*_{\mathbb P_2,H,w}\|_{\mathbb P_0}
\le C\,\mathrm H(\mathbb P_1,\mathbb P_2),
\]
after possibly shrinking \(\mathfrak P\).
\end{lemma}

\begin{proof}
Set \(\delta_{12}:=\mathrm H(\mathbb P_1,\mathbb P_2)\). Hellinger
contraction under the randomized-time transition map gives
\(\mathrm H(P_1,P_2)\le\delta_{12}\).
Set \(\|h\|_0:=\|h\|_{w,P_0}\) and
\[
a_P(f,h):=E_P\{w(X)\mathcal T_P(f)(X)\mathcal T_P(h)(X)\}.
\]
Write \(U_{P,h}:=\mathcal T_P h\).
By Lemma~\ref{lemma::localInversePersistence}, there is \(c_a>0\) such that
\[
a_P(h,h)\ge c_a\|h\|_0^2,\qquad h\in \overline{H},
\]
uniformly over \(P\in\mathcal P\). Let
\[
b_P^q(h):=E_P\{w(X)r_P(X)\mathcal T_P(h)(X)\},
\qquad
b_{\mathbb P}^\alpha(h):=E_{\mathbb P}\{m(S_0,h)\}.
\]
Lemma~\ref{lemma::weightedProjectionIdentities} gives the variational equations
\[
a_P(q_P,h)=b_P^q(h),
\qquad
a_P(\alpha_{\mathbb P},h)=b_{\mathbb P}^\alpha(h),
\qquad h\in \overline{H}.
\]
Lemma~\ref{lemma::baselineFunctionalPerturbation} gives, for
\(h\in \overline{H}\),
\[
|b_{\mathbb P_1}^\alpha(h)-b_{\mathbb P_2}^\alpha(h)|
\le C\,\delta_{12}\|h\|_0.
\]
For \(b^q\), write
\[
\begin{aligned}
&b_{P_1}^q(h)-b_{P_2}^q(h)\\
&=
E_{P_1}\{w(X)[r_{P_1}(X)-r_{P_2}(X)]\mathcal T_{P_1}h(X)\}\\
&\quad+
E_{P_1}\{w(X)r_{P_2}(X)[\mathcal T_{P_1}h(X)-\mathcal T_{P_2}h(X)]\}\\
&\quad+
(P_1-P_2)\{w(X)r_{P_2}(X)\mathcal T_{P_2}h(X)\}.
\end{aligned}
\]
The first two terms are bounded by the pairwise bounds for \(r_P\) and
\(\mathcal T_P h\), Condition~\ref{cond::boundedWeightsPolicyRewards}, and
Lemma~\ref{lemma::localNormEquivalence}. The last term is bounded by
\eqref{eqn::hellingerExpectationIneq}, because
\(r_{P_2}\) is uniformly bounded and \(\mathcal T_{P_2}h\) is uniformly bounded
in \(L^2(P_0)\). Hence
\[
|b_{P_1}^q(h)-b_{P_2}^q(h)|
\le C\,\mathrm H(P_1,P_2)\|h\|_0.
\]

For each \(\mathbb P\in\mathfrak P\) and its pooled transition law \(P\),
Conditions~\ref{cond::boundedWeightsPolicyRewards}
and~\ref{cond::targetMapContinuity}, together with
Lemmas~\ref{lemma::localNormEquivalence}
and~\ref{lemma::localInversePersistence}, give
\[
|b_P^q(h)|+|b_{\mathbb P}^\alpha(h)|\le C\|h\|_0,\qquad h\in \overline{H} .
\]
Taking \(h=q_P\) and \(h=\alpha_{\mathbb P}\) in the variational equations and using
coercivity of \(a_P\) gives
\[
\sup_{\mathbb P\in\mathfrak P}
\{\|q_P\|_0+\|\alpha_{\mathbb P}\|_0\}<\infty .
\]

We now prove the one-sided bilinear perturbation bound used to compare the
variational equations. Let
\(g\in\{q_{P_2},\alpha_{\mathbb P_2}\}\). The preceding display gives
\(\|g\|_0\le C\), and Condition~\ref{cond::boundednuis} gives
\(\|g\|_\infty\le C\). If \(g=q_{P_2}\), then
\(U_{P_2,g}=\mathcal T_{P_2}q_{P_2}\); if
\(g=\alpha_{\mathbb P_2}\), then
\(U_{P_2,g}=\tau_{\mathbb P_2}\). Since
\(\|\mathcal T_{P_2}g\|_\infty\le(1+\gamma)\|g\|_\infty\), we have
\(\|U_{P_2,g}\|_\infty\le C\).
Decompose
\[
\begin{aligned}
&a_{P_1}(g,h)-a_{P_2}(g,h)\\
&=
E_{P_1}\{w(X)[U_{P_1,g}(X)-U_{P_2,g}(X)]U_{P_1,h}(X)\}\\
&\quad+
E_{P_1}\{w(X)U_{P_2,g}(X)[U_{P_1,h}(X)-U_{P_2,h}(X)]\}\\
&\quad+
(P_1-P_2)\{w(X)U_{P_2,g}(X)U_{P_2,h}(X)\}.
\end{aligned}
\]
The first two terms are bounded by
\(C\,\mathrm H(P_1,P_2)\|h\|_0\) using
Lemma~\ref{lemma::primitiveHellingerPerturbations}, the boundedness of
\(\mathcal T_P:\overline{H}\to L^2_w(P_{0,X})\) from
Lemma~\ref{lemma::localInversePersistence}, and
Lemma~\ref{lemma::localNormEquivalence}. For the
last term, \eqref{eqn::hellingerExpectationIneq}, the boundedness of
\(wU_{P_2,g}\), the boundedness of \(\mathcal T_P\), and
Lemma~\ref{lemma::localNormEquivalence}
give
\[
|(P_1-P_2)\{w(X)U_{P_2,g}(X)U_{P_2,h}(X)\}|
\le C\,\mathrm H(P_1,P_2)\|h\|_0 .
\]
Thus
\begin{equation}
|a_{P_1}(g,h)-a_{P_2}(g,h)|
\le C\,\mathrm H(P_1,P_2)\|h\|_0,
\qquad g\in\{q_{P_2},\alpha_{\mathbb P_2}\}.
\label{eqn::oneSidedBilinearPerturbation}
\end{equation}
Taking \(h=q_{P_1}-q_{P_2}\) in the difference of the two \(q\)-equations yields
\[
c_a\|q_{P_1}-q_{P_2}\|_0^2
\le
|b_{P_1}^q(h)-b_{P_2}^q(h)|
+|a_{P_1}(q_{P_2},h)-a_{P_2}(q_{P_2},h)|
\le
C\,\mathrm H(P_1,P_2)\|h\|_0.
\]
Thus \(\|q_{P_1}-q_{P_2}\|_0\le C\delta_{12}\). The same argument with
\(\alpha_{\mathbb P_1}-\alpha_{\mathbb P_2}\),
\(b_{\mathbb P}^\alpha\), and
\eqref{eqn::oneSidedBilinearPerturbation} gives
\(\|\alpha_{\mathbb P_1}-\alpha_{\mathbb P_2}\|_0\le C\delta_{12}\).

For the Bellman--Riesz image,
\[
\tau_{\mathbb P_1}-\tau_{\mathbb P_2}
=
\mathcal T_{P_1}(\alpha_{\mathbb P_1}-\alpha_{\mathbb P_2})
+(\mathcal T_{P_1}-\mathcal T_{P_2})\alpha_{\mathbb P_2}.
\]
The boundedness of \(\mathcal T_P\) on \(\overline{H}\) from
Lemma~\ref{lemma::localInversePersistence}, the bound on
\(\alpha_{\mathbb P_1}-\alpha_{\mathbb P_2}\), and
Lemma~\ref{lemma::primitiveHellingerPerturbations} give
\(\|\tau_{\mathbb P_1}-\tau_{\mathbb P_2}\|_0\le C\delta_{12}\). Similarly,
\[
e_{P_1}-e_{P_2}
=
\mathcal T_{P_1}(q_{P_1}-q_{P_2})
+(\mathcal T_{P_1}-\mathcal T_{P_2})q_{P_2}
-(r_{P_1}-r_{P_2}),
\]
so \(\|e_{P_1}-e_{P_2}\|_0\le C\delta_{12}\). Condition~\ref{cond::targetMapContinuity}
and Lemma~\ref{lemma::baselineFunctionalPerturbation} give
\[
\|m(S_0,q_{P_1}-q_{P_2})\|_{\mathbb P_0}
+|\Psi_{H,w}(\mathbb P_1)-\Psi_{H,w}(\mathbb P_2)|
\le C\delta_{12}.
\]

It remains to control the two product terms in the influence function. Write
\[
\Delta_j(Z):=Y+\gamma V^\pi(q_{P_j})(S')-q_{P_j}(X),
\qquad
B_j(Z):=\alpha_{\mathbb P_j}(X)
-\gamma V^\pi(\alpha_{\mathbb P_j})(S')-\tau_{\mathbb P_j}(X).
\]
The target-transition domination in the definition of \(\mathcal P\) and
Condition~\ref{cond::boundedWeightsPolicyRewards} give
\[
\|\Delta_1-\Delta_2\|_{P_0}
+\|B_1-B_2\|_{P_0}
\le
C\{\|q_{P_1}-q_{P_2}\|_0
+\|\alpha_{\mathbb P_1}-\alpha_{\mathbb P_2}\|_0
+\|\tau_{\mathbb P_1}-\tau_{\mathbb P_2}\|_0\}.
\]
Corollary~\ref{corollary::boundedIFEnvelopes} gives uniform
supremum-norm bounds for \(\tau_{\mathbb P}\), \(e_P\),
\(\Delta_P(q_P)\), and \(B_{\mathbb P}(\alpha_{\mathbb P})\).
Combining these bounds with the preceding
nuisance-stability bounds gives
\[
\begin{aligned}
&\|w\tau_{\mathbb P_1}\Delta_1-w\tau_{\mathbb P_2}\Delta_2\|_{P_0}\\
&\quad\le
C\|\tau_{\mathbb P_1}-\tau_{\mathbb P_2}\|_0
+C\|\Delta_1-\Delta_2\|_{P_0}
\le C\delta_{12},
\end{aligned}
\]
and, using \(e_P=\mathcal T_P(q_P)-r_P\),
\[
\begin{aligned}
&\|w e_{P_1}B_1-w e_{P_2}B_2\|_{P_0}\\
&\quad\le
C\|e_{P_1}-e_{P_2}\|_0
+C\|B_1-B_2\|_{P_0}
\le C\delta_{12}.
\end{aligned}
\]
Jensen's inequality bounds the \(L^2(\mathbb P_0)\) norm of each time average
by the corresponding pooled \(L^2(P_0)\) norm. Combining the preceding bounds
proves the displayed pairwise \(L^2(\mathbb P_0)\) bound.
\end{proof}

\begin{lemma}[Trajectory lifting]
\label{lemma::trajectoryLifting}
Let \((\mathbb P_\varepsilon)\subset\mathfrak P\) be a regular trajectory
submodel through \(\mathbb P_0\) with score \(\ell\), and let
\(P_\varepsilon\) denote its pooled transition law. For bounded \(b\) and
\(f\),
\[
\left.\frac{d}{d\varepsilon}E_{\mathbb P_\varepsilon}\{b(S_0)\}
\right|_{\varepsilon=0}=E_{\mathbb P_0}\{\ell(O)b(S_0)\},\qquad
\left.\frac{d}{d\varepsilon}P_\varepsilon f\right|_{\varepsilon=0}
=E_{\mathbb P_0}\!\left\{\ell(O)\frac1T\sum_{t=0}^{T-1}f(Z_t)\right\}.
\]
\end{lemma}

\begin{proof}
Both identities follow by differentiating the trajectory expectation; the
second also uses the definition of the pooled transition law.
\end{proof}

\subsection{Fixed-weight projection theory}

\begin{proof}[Proof of Theorem~\ref{theorem::projectionPD}]
Fix \(w\) satisfying Condition~\ref{cond::boundedWeightsPolicyRewards}.
For a trajectory law \(\mathbb P\) with pooled transition law \(P\), write
\(q_P^\circ=q_{P,H,w}\),
\(\alpha_{\mathbb P}^\circ=\alpha_{\mathbb P,H,w}\),
\(\tau_{\mathbb P}^\circ=\mathcal T_P(\alpha_{\mathbb P}^\circ)\), and
\(e_P^\circ=\mathcal T_P(q_P^\circ)-r_P\). We use
Lemma~\ref{lemma::weightedProjectionIdentities} throughout. The proof has
three steps. First, we establish a functional von Mises expansion. We then
establish pathwise differentiability along bounded linear tilt submodels and
determine the EIF. Finally, we extend pathwise differentiability to regular
QMD submodels by a dense-score argument.

To derive the von Mises identity, fix any
\(\overline{\mathbb P}\in\mathfrak P\), with pooled transition law
\(\overline P\). The normal equations and Riesz identity are
\begin{equation}
\langle r_P-\mathcal T_P(q_P^\circ),\mathcal T_P(h)\rangle_{w,P}=0,
\qquad
E_{\mathbb P}\{m(S_0,h)\}
=\langle\mathcal T_P(\alpha_{\mathbb P}^\circ),\mathcal T_P(h)\rangle_{w,P}.
\label{eqn::weightedProjectionNormalRiesz}
\end{equation}
In particular, at \(\mathbb P_0\),
\[
r_0=\mathcal T_0(q_0^\circ)-e_0^\circ,
\qquad
\langle e_0^\circ,\mathcal T_0(\alpha_{\overline{\mathbb P}}^\circ)\rangle_{w,P_0}=0.
\]
The Riesz identity at \(\mathbb P_0\) also gives
\[
E_{\mathbb P_0}\{m(S_0,q_{\overline P}^\circ)\}-\Psi_{H,w}(\mathbb P_0)
=
\left\langle \tau_0^\circ,
\mathcal T_0(q_{\overline P}^\circ-q_0^\circ)
\right\rangle_{w,P_0}.
\]
By iterated expectation under \(\mathbb P_0\),
\[
\begin{aligned}
\mathbb P_0\Phi^*_{\overline{\mathbb P},H,w}
&=
E_{\mathbb P_0}\{m(S_0,q_{\overline P}^\circ)\}
-\Psi_{H,w}(\overline{\mathbb P})
+\left\langle \tau_{\overline{\mathbb P}}^\circ,
r_0-\mathcal T_0(q_{\overline P}^\circ)\right\rangle_{w,P_0}\\
&\quad
-\left\langle e_{\overline P}^\circ,
\mathcal T_0(\alpha_{\overline{\mathbb P}}^\circ)
-\tau_{\overline{\mathbb P}}^\circ\right\rangle_{w,P_0}.
\end{aligned}
\]
Combining the preceding displays yields
\begin{equation}
\begin{aligned}
&\Psi_{H,w}(\overline{\mathbb P})-\Psi_{H,w}(\mathbb P_0)
+\mathbb P_0\Phi^*_{\overline{\mathbb P},H,w}\\
&=
\left\langle
\tau_{\overline{\mathbb P}}^\circ-\tau_0^\circ,
\mathcal T_0(q_0^\circ-q_{\overline P}^\circ)
\right\rangle_{w,P_0}\\
&\quad+
\left\langle
\tau_{\overline{\mathbb P}}^\circ
-\mathcal T_0(\alpha_{\overline{\mathbb P}}^\circ),
e_{\overline P}^\circ-e_0^\circ
\right\rangle_{w,P_0}.
\end{aligned}
\label{eqn::weightedProjectionVonMises}
\end{equation}
This proves the von Mises expansion.

It remains to establish pathwise differentiability. Take
\(\ell\in L^\infty_0(\mathbb P_0)\) and define the bounded linear tilt
\(d\mathbb P_\varepsilon/d\mathbb P_0=1+\varepsilon\ell\) for sufficiently
small \(|\varepsilon|\). Let \(P_\varepsilon\) be its pooled transition law.
The pooled likelihood ratio is bounded above and away from zero, and
\[
\mathrm H(P_\varepsilon,P_0)
\le \mathrm H(\mathbb P_\varepsilon,\mathbb P_0)=O(|\varepsilon|).
\]
Choose the fixed local-model bounds with strict slack at \(\mathbb P_0\),
which is possible by enlarging the upper bounds and reducing the positive
lower bound. Thus, after shrinking \(|\varepsilon|\),
\(\mathbb P_\varepsilon\in\mathfrak P\). By
Lemma~\ref{lemma::weightedProjectionHellingerStability},
\[
\|q_{P_\varepsilon}^\circ-q_0^\circ\|_{w,P_0}
+\|\alpha_{\mathbb P_\varepsilon}^\circ-\alpha_0^\circ\|_{w,P_0}
+\|\tau_{\mathbb P_\varepsilon}^\circ-\tau_0^\circ\|_{w,P_0}
+\|e_{P_\varepsilon}^\circ-e_0^\circ\|_{w,P_0}
=O(|\varepsilon|).
\]
Lemma~\ref{lemma::primitiveHellingerPerturbations} also gives
\[
\|\mathcal T_0(q_0^\circ-q_{P_\varepsilon}^\circ)\|_{w,P_0}
+\|\tau_{\mathbb P_\varepsilon}^\circ
-\mathcal T_0(\alpha_{\mathbb P_\varepsilon}^\circ)\|_{w,P_0}
=O(|\varepsilon|).
\]
Therefore both products in \eqref{eqn::weightedProjectionVonMises} are
\(O(\varepsilon^2)\).

For every \(\mathbb P\in\mathfrak P\),
\(\mathbb P\Phi^*_{\mathbb P,H,w}=0\): the baseline term is centered, the
Bellman-residual term is zero by
\eqref{eqn::weightedProjectionNormalRiesz} with
\(h=\alpha_{\mathbb P}^\circ\), and the final term has pooled conditional mean
zero given \(X\). Hence
\[
-\frac{\mathbb P_0\Phi^*_{\mathbb P_\varepsilon,H,w}}{\varepsilon}
=
\frac{(\mathbb P_\varepsilon-\mathbb P_0)
\Phi^*_{\mathbb P_\varepsilon,H,w}}{\varepsilon}
=E_{\mathbb P_0}\{\ell(O)\Phi^*_{\mathbb P_\varepsilon,H,w}(O)\}.
\]
By Lemma~\ref{lemma::weightedProjectionHellingerStability},
\(\Phi^*_{\mathbb P_\varepsilon,H,w}\to\Phi^*_{0,H,w}\) in
\(L^2(\mathbb P_0)\). Combining this fact with
\eqref{eqn::weightedProjectionVonMises} gives
\[
\frac{\Psi_{H,w}(\mathbb P_\varepsilon)-\Psi_{H,w}(\mathbb P_0)}{\varepsilon}
=E_{\mathbb P_0}\{\ell(O)\Phi^*_{0,H,w}(O)\}+o(1).
\]
This derivative also follows directly from
Lemma~\ref{lemma::trajectoryLifting}: its adjoint maps the baseline derivative
to the first term of \(\Phi^*_{0,H,w}\) and each pooled-transition derivative
to the corresponding time average.

Finally, \(\Phi^*_{0,H,w}\in L^2_0(\mathbb P_0)\), and the same stability
bounds show that \(\Psi_{H,w}\) is locally Hellinger Lipschitz on
\(\mathfrak P\). Lemma~\ref{lemma::denseScoreExtension} therefore extends the
bounded-tilt derivative to every regular QMD submodel. Since bounded linear
tilts generate the tangent space \(L^2_0(\mathbb P_0)\),
\(\Phi^*_{0,H,w}\) is the efficient influence function.
\end{proof}

\begin{proof}[Proof of Theorem~\ref{theorem::projectionApprox}]
Let \(b_0:=\mathcal T_0(q_0)\), and let \(\Pi=\Pi_{0,H,w}\) be the
\(L^2_w(P_{0,X})\)-projection onto \(\mathcal T_0(\overline H)\). By the
definitions of the weighted Bellman projection and weighted Riesz representer,
\[
\mathcal T_0(q_{0,H,w})=\Pi b_0,
\qquad
\mathcal T_0(\alpha_{0,H,w})
=\Pi\mathcal T_0(\alpha^{\rm np}_{0,w}).
\]
Hence
\[
\begin{aligned}
\Psi(\mathbb P_0)-\Psi_{H,w}(\mathbb P_0)
&=
\left\langle \mathcal T_0(\alpha^{\rm np}_{0,w}),b_0\right\rangle_{w,P_0}
-\left\langle
\Pi\mathcal T_0(\alpha^{\rm np}_{0,w}),\Pi b_0
\right\rangle_{w,P_0}\\
&=
\left\langle
\mathcal T_0(\alpha^{\rm np}_{0,w})
-\Pi\mathcal T_0(\alpha^{\rm np}_{0,w}),
b_0-\Pi b_0
\right\rangle_{w,P_0}.
\end{aligned}
\]
The second equality uses the orthogonality of both projection residuals to
\(\mathcal T_0(\overline H)\). Substituting
\(b_0=\mathcal T_0(q_0)\) gives the stated identity.
\end{proof}

\begin{proof}[Proof of Theorem~\ref{theorem::projectionEfficiency}]
Let \(D_{n,H,w}\) denote the uncentered summand defining
\(\psi^*_{n,H,w}=\mathbb P_nD_{n,H,w}\) in the main text. Abbreviate
\(q_n=q_{n,H,w}\), \(\alpha_n=\alpha_{n,H,w}\),
\(\tau_n=\tau_{n,H,w}\), \(e_n=e_{n,H,w}\), \(q_0=q_{0,H,w}\),
\(\alpha_0=\alpha_{0,H,w}\), \(\tau_0=\tau_{0,H,w}\), and
\(e_0=e_{0,H,w}\).

First we derive the von Mises remainder. By iterated expectation under
\(\mathbb P_0\),
\[
\begin{aligned}
\mathbb P_0D_{n,H,w}-\Psi_{H,w}(\mathbb P_0)
&=
E_{\mathbb P_0}\{m(S_0,q_n)\}-\Psi_{H,w}(\mathbb P_0)
+\left\langle \tau_n,r_0-\mathcal T_0(q_n)\right\rangle_{w,P_0}\\
&\quad
-\left\langle e_n,\mathcal T_0(\alpha_n)-\tau_n\right\rangle_{w,P_0}.
\end{aligned}
\]
The Riesz identity at \(P_0\) gives
\[
E_{\mathbb P_0}\{m(S_0,q_n)\}-\Psi_{H,w}(\mathbb P_0)
=
\left\langle \tau_0,\mathcal T_0(q_n-q_0)\right\rangle_{w,P_0}.
\]
Also \(r_0=\mathcal T_0(q_0)-e_0\), and the projection normal equation gives
\[
\left\langle e_0,\mathcal T_0(\alpha_n)\right\rangle_{w,P_0}=0,
\]
because \(\alpha_n\in\overline H\). Therefore
\[
\begin{aligned}
\mathbb P_0D_{n,H,w}-\Psi_{H,w}(\mathbb P_0)
&=
\left\langle \tau_0,\mathcal T_0(q_n-q_0)\right\rangle_{w,P_0}
+
\left\langle \tau_n,\mathcal T_0(q_0-q_n)-e_0\right\rangle_{w,P_0}\\
&\quad
-\left\langle e_n,\mathcal T_0(\alpha_n)-\tau_n\right\rangle_{w,P_0}\\
&=
\left\langle
\tau_n-\tau_0,\mathcal T_0(q_0-q_n)
\right\rangle_{w,P_0}\\
&\quad+
\left\langle
\tau_n-\mathcal T_0(\alpha_n),e_n-e_0
\right\rangle_{w,P_0}.
\end{aligned}
\]
The last equality uses
\(\left\langle e_0,\mathcal T_0(\alpha_n)\right\rangle_{w,P_0}=0\).

Hence
\[
\psi^*_{n,H,w}-\Psi_{H,w}(\mathbb P_0)
=
(\mathbb P_n-\mathbb P_0)D_{n,H,w}
+
R_{n,H,w},
\]
where
\[
R_{n,H,w}
:=
\mathbb P_0D_{n,H,w}-\Psi_{H,w}(\mathbb P_0)
\]
satisfies
\[
\begin{aligned}
|R_{n,H,w}|
&\le
\left\|\tau_{n,H,w}-\tau_{0,H,w}\right\|_{w,P_0}
\left\|\mathcal T_0(q_{n,H,w})-\mathcal T_0(q_{0,H,w})\right\|_{w,P_0}\\
&\quad+
\left\|\mathcal T_0(\alpha_{n,H,w})-\tau_{n,H,w}\right\|_{w,P_0}
\left\|e_{n,H,w}-e_{0,H,w}\right\|_{w,P_0}.
\end{aligned}
\]
Condition~\ref{cond::projectionRates} gives
\(R_{n,H,w}=o_p(n^{-1/2})\).

Now write
\[
\Phi^*_{n,H,w}:=D_{n,H,w}-\Psi_{H,w}(\mathbb P_0),
\qquad
\Phi^*_{0,H,w}:=D_{0,H,w}-\Psi_{H,w}(\mathbb P_0).
\]
Since constants vanish under \(\mathbb P_n-\mathbb P_0\),
\[
(\mathbb P_n-\mathbb P_0)D_{n,H,w}
=
(\mathbb P_n-\mathbb P_0)\Phi^*_{n,H,w}.
\]
Let \(f_n:=\Phi^*_{n,H,w}-\Phi^*_{0,H,w}\). By
Condition~\ref{cond::projectionSampleSplit}, conditional on
\(\mathcal D_n^{\rm tr}\), \(f_n\) is fixed and the analysis trajectories are
i.i.d. from \(\mathbb P_0\). Therefore
\[
E\!\left[
\left\{\sqrt n(\mathbb P_n-\mathbb P_0)f_n\right\}^2
\mid \mathcal D_n^{\rm tr}
\right]
=
\mathbb P_0(f_n-\mathbb P_0f_n)^2
\le
\mathbb P_0f_n^2.
\]
Condition~\ref{cond::projectionIFConsistency} and conditional Chebyshev's
inequality give
\[
(\mathbb P_n-\mathbb P_0)\{\Phi^*_{n,H,w}-\Phi^*_{0,H,w}\}
=
o_p(n^{-1/2}).
\]
Combining the preceding displays yields
\[
\psi^*_{n,H,w}-\Psi_{H,w}(\mathbb P_0)
=
(\mathbb P_n-\mathbb P_0)\Phi^*_{0,H,w}
+
o_p(n^{-1/2}).
\]
Equivalently,
\[
\sqrt n\{\psi^*_{n,H,w}-\Psi_{H,w}(\mathbb P_0)\}
=
\sqrt n(\mathbb P_n-\mathbb P_0)\Phi^*_{0,H,w}+o_p(1).
\]
The central limit theorem follows from
\(\Phi^*_{0,H,w}\in L^2_0(\mathbb P_0)\). Under
Condition~\ref{cond::boundednuis}, efficiency follows from
Theorem~\ref{theorem::projectionPD}, which identifies
\(\Phi^*_{0,H,w}\) as the nonparametric efficient influence function for
\(\Psi_{H,w}\).
\end{proof}

\subsection{Correct specification and optimal weighting}

\begin{lemma}[Correct-specification reduction]
\label{lemma::correctSpecificationWeightedReduction}
Suppose Condition~\ref{cond::bounded} holds. If
\(q_0\in\overline H\), then \(q_{0,H,w}=q_0\),
\(\Psi_{H,w}(\mathbb P_0)=\Psi(\mathbb P_0)\), and
\[
\Phi^*_{0,H,w}(O)
=
m(S_0,q_0)-\Psi(\mathbb P_0)
+\frac1T\sum_{t=0}^{T-1}
w(X_t)\tau_{0,H,w}(X_t)\Delta_0(Z_t),
\]
where \(\Delta_0(Z)=Y+\gamma V^\pi(q_0)(S')-q_0(X)\).
\end{lemma}

\begin{proof}
Correct specification gives \(r_0=\mathcal T_0(q_0)\) with
\(q_0\in\overline H\). Thus the weighted projection of \(r_0\) onto
\(\mathcal T_0(\overline H)\) is \(r_0\) itself, so
\(q_{0,H,w}=q_0\) by the \(P_0\)-restricted inverse in
Condition~\ref{cond::restrictedBellmanInverse}.
The target identity then gives
\(\Psi_{H,w}(\mathbb P_0)=\Psi(\mathbb P_0)\). Substituting
\(e_{0,H,w}=\mathcal T_0(q_{0,H,w})-r_0=0\) into
\(\Phi^*_{0,H,w}\) removes the final projection term and gives the displayed
influence function.
\end{proof}

\begin{proof}[Proof of Theorem~\ref{theorem::optimalWeighting}]
By Lemma~\ref{lemma::correctSpecificationWeightedReduction} with
\(w=w_{\rm opt}\),
\(\Psi_{H,w_{\rm opt}}(\mathbb P_0)=\Psi(\mathbb P_0)\) and the weighted
projection influence function reduces to
\[
m(S_0,q_0)-\Psi(\mathbb P_0)
+\frac1T\sum_{t=0}^{T-1}
w_{\rm opt}(X_t)\tau_{0,H,w_{\rm opt}}(X_t)\Delta_0(Z_t).
\]
The non-degeneracy bound in Condition~\ref{cond::restrictedModelRegularity}
gives two-sided boundedness of \(w_{\rm opt}=\sigma_0^{-2}\). Together,
Conditions~\ref{cond::bounded}, \ref{cond::boundednuis}, and
\ref{cond::restrictedModelRegularity}, with \(w=w_{\rm opt}\), put the local
restricted model in the setting of Theorem~\ref{theorem::EIFrestricted}. The
weighted Riesz equation in \eqref{eqn::weightedRieszRestricted} is the defining
equation for \(\alpha_{0,H,w_{\rm opt}}\), so
Theorem~\ref{theorem::EIFrestricted} identifies the displayed function as the
restricted-model efficient influence function.
\end{proof}

\begin{proof}[Proof of Theorem~\ref{theorem::optWeightedAL}]
Condition on the auxiliary training sample used to construct
\(\widehat w_n\), \(q_{n,H,\widehat w_n}\), and
\(\tau_{n,H,\widehat w_n}\). On the analysis sample these quantities are fixed.
Let \(A_n\) be the event that
Condition~\ref{cond::boundedWeightsPolicyRewards} holds with
\(w=\widehat w_n\). By assumption, \(\Pr(A_n)\to1\), so it suffices to work on
\(A_n\). Lemma~\ref{lemma::correctSpecificationWeightedReduction} then applies
with \(w=\widehat w_n\), and the projection residual term in
\(\Phi^*_{n,H,\widehat w_n}\) is absent. The fixed-weight von Mises
expansion gives
\[
\psi^{\rm opt}_{n,H,\widehat w_n}-\Psi(\mathbb P_0)
=
(\mathbb P_n-\mathbb P_0)\Phi^{\widehat w_n}_{n,H}
+R_{n,H,\widehat w_n},
\]
where, under correct specification,
\[
|R_{n,H,\widehat w_n}|
\le
\left\|
\tau_{n,H,\widehat w_n}
-\tau_{0,H,\widehat w_n}
\right\|_{\widehat w_n,P_0}
\left\|
\mathcal T_0(q_{n,H,\widehat w_n})-\mathcal T_0(q_0)
\right\|_{\widehat w_n,P_0}.
\]
Indeed, on \(A_n\),
Lemma~\ref{lemma::correctSpecificationWeightedReduction} gives
\(q_{0,H,\widehat w_n}=q_0\). Hence
Condition~\ref{cond::optweightRates} implies that
\(R_{n,H,\widehat w_n}=o_p(n^{-1/2})\).

Adding and subtracting the fixed limit
\(\Phi^{w_\infty}_{0,H}\) yields
\[
\psi^{\rm opt}_{n,H,\widehat w_n}-\Psi(\mathbb P_0)
=
(\mathbb P_n-\mathbb P_0)\Phi^{w_\infty}_{0,H}
+
(\mathbb P_n-\mathbb P_0)\{\Phi^{\widehat w_n}_{n,H}
-\Phi^{w_\infty}_{0,H}\}
+o_p(n^{-1/2}).
\]
Let \(g_n:=\Phi^{\widehat w_n}_{n,H}-\Phi^{w_\infty}_{0,H}\). By
Condition~\ref{cond::optweightSampleSplit}, conditional on
\(\mathcal D_n^{\rm tr}\), \(g_n\) is fixed and the analysis trajectories are
i.i.d. from \(\mathbb P_0\). Therefore
\[
E\!\left[
\left\{\sqrt n(\mathbb P_n-\mathbb P_0)g_n\right\}^2
\mid \mathcal D_n^{\rm tr}
\right]
=
\mathbb P_0(g_n-\mathbb P_0g_n)^2
\le
\mathbb P_0g_n^2
=
\|\Phi^{\widehat w_n}_{n,H}
-\Phi^{w_\infty}_{0,H}\|_{\mathbb P_0}^2 .
\]
Condition~\ref{cond::optweightIFConsistency} and conditional Chebyshev's
inequality give
\[
(\mathbb P_n-\mathbb P_0)g_n=o_p(n^{-1/2}).
\]
Thus
\[
\psi^{\rm opt}_{n,H,\widehat w_n}-\Psi(\mathbb P_0)
=
(\mathbb P_n-\mathbb P_0)\Phi^{w_\infty}_{0,H}
+o_p(n^{-1/2}).
\]
The central limit theorem follows from
\(\Phi^{w_\infty}_{0,H}\in L^2_0(\mathbb P_0)\) and
\(E_{\mathbb P_0}\{(\Phi^{w_\infty}_{0,H})^2\}\in(0,\infty)\). If the local
restricted-model conditions in Theorem~\ref{theorem::optimalWeighting}
hold and \(w_\infty=cw_{\rm opt}\) for some \(c>0\), then multiplying the
weight by \(c\) leaves the projection unchanged and divides its Riesz image by
\(c\). Consequently,
\[
(cw_{\rm opt})\tau_{0,H,cw_{\rm opt}}
=w_{\rm opt}\tau_{0,H,w_{\rm opt}},
\]
so \(\Phi^{w_\infty}_{0,H}=\Phi^{w_{\rm opt}}_{0,H}\). Theorem~\ref{theorem::optimalWeighting}
therefore identifies the limiting variance as the restricted-model efficiency
bound.
\end{proof}

\section{Restricted-model EIF}
\label{appendix::restrictedmodelEIF}

\subsection{Local model and tangent space}

Appendix~\ref{appendix::weightedProjectionOptimalProofs} identifies influence
functions for the weighted Bellman projection parameters. This section proves the
restricted-model efficiency claim in Theorem~\ref{theorem::optimalWeighting}
in the time-homogeneous restricted model
\(\mathfrak P_H^{\rm M}\) from Definition~\ref{def::restrictedLocalModel}.
Define
\[
\Delta_0(Z):=Y+\gamma V^\pi(q_0)(S')-q_0(X),
\qquad
\sigma_0^2(X):=E_{P_0}\{\Delta_0(Z)^2\mid X\}.
\]
Condition~\ref{cond::restrictedModelRegularity} gives
\(q_0\in\overline H\) and \(0<c\le\sigma_0^2(X)\le C<\infty\). Since
\(q_0\) solves the Bellman equation, \(E_{P_0}\{\Delta_0(Z)\mid X\}=0\).

Under \(\mathbb P_0\), let \(\mu_0\) denote the law of \(S_0\), let
\(g_0(da\mid s)\) denote the behavior policy, and let
\(K_0(dy,ds'\mid x)\) denote the common outcome--transition kernel. Define
the restricted trajectory score class
\[
\begin{aligned}
\mathcal S_H^{\rm M}:=\Big\{\,u_0(S_0)
+\sum_{t=0}^{T-1}\{b(A_t,S_t)+a(X_t)\Delta_0(Z_t)\\[-1mm]
&\hspace{35mm}+\eta(Z_t)\}:\\[-1mm]
&\ E_{\mathbb P_0}\{u_0(S_0)\}=0,\\
&\ E_{P_0}\{b(A,S)\mid S\}=0,\\
&\ E_{P_0}\{\eta(Z)\mid X\}=0,\quad
E_{P_0}\{\Delta_0(Z)\eta(Z)\mid X\}=0,\\
&\ a(X)\sigma_0^2(X)\in\mathcal T_0(\overline H)\,\Big\},
\end{aligned}
\]
with all displayed summands square integrable. Let
\((\mathcal S_H^{\rm M})^\infty\) denote the elements for which
\(u_0\), \(b\), \(a\Delta_0+\eta\), and
\[
h_a:=\mathcal T_0^{-1}\{a(X)\sigma_0^2(X)\}\in \overline H
\]
are bounded.

\begin{lemma}[Density of bounded restricted scores]
\label{lemma::restrictedBoundedScoreDensity}
Suppose \(H\subset L^\infty(P_{0,X})\) is linear,
\(\overline H\) is its \(L^2(P_{0,X})\)-closure,
Condition~\ref{cond::bounded} holds for the weight \(w\) used below, and
Condition~\ref{cond::restrictedModelRegularity} holds. Then
\[
\overline{(\mathcal S_H^{\rm M})^\infty}^{L^2(\mathbb P_0)}
=
\overline{\mathcal S_H^{\rm M}}^{L^2(\mathbb P_0)} .
\]
\end{lemma}

\begin{proof}
We approximate the four components in the definition of
\(\mathcal S_H^{\rm M}\). If \(u_0(S_0)\in L^2(\mathbb P_0)\) and
\(E_{\mathbb P_0}\{u_0(S_0)\}=0\), then the
centered truncations
\[
u_{0,M}(S_0):=\{-M\vee u_0(S_0)\wedge M\}
-E_{\mathbb P_0}\{-M\vee u_0(S_0)\wedge M\}
\]
are bounded, mean zero, and converge to \(u_0\) in \(L^2(\mathbb P_0)\).
Likewise, if \(E_{P_0}\{b(A,S)\mid S\}=0\), conditionally centering bounded
truncations of \(b\) gives bounded behavior-policy scores converging to \(b\)
in pooled \(L^2(P_0)\).

For the residual-score component \(a(X)\Delta_0(Z)\), suppose
\(a\sigma_0^2=\mathcal T_0h_a\) for some \(h_a\in\overline H\). Since \(H\) is
dense in \(\overline H\), choose \(h_n\in H\) with
\(\|h_n-h_a\|_{w,P_0}\to0\), and set
\[
a_n(X):=\frac{\mathcal T_0h_n(X)}{\sigma_0^2(X)} .
\]
Each \(h_n\) is bounded because \(H\subset L^\infty(P_{0,X})\). Hence
\(a_n(X)\Delta_0(Z)\) is bounded: \(\mathcal T_0h_n\) is bounded by
boundedness of \(h_n\) and \(\gamma<1\), \(\sigma_0^2\) is bounded away from
zero by Condition~\ref{cond::restrictedModelRegularity}, and
the baseline reward bound in Condition~\ref{cond::boundedWeightsPolicyRewards}
and the \(\gamma<1\) Bellman contraction give
\(\|q_0\|_\infty\le \|Y\|_\infty/(1-\gamma)\), hence
\(\|\Delta_0\|_\infty<\infty\). Moreover,
\[
\begin{aligned}
E_{P_0}\{(a_n-a)^2\Delta_0^2\}
&=
E_{P_0}\{(a_n-a)^2\sigma_0^2(X)\}\\
&\le
c^{-1}E_{P_0}[\{\mathcal T_0(h_n-h_a)(X)\}^2]
\to0,
\end{aligned}
\]
because \(\mathcal T_0\) is bounded on \(\overline H\) by
Lemma~\ref{lemma::localInversePersistence} and Condition~\ref{cond::boundedWeightsPolicyRewards}.

For the component orthogonal to both \(1\) and \(\Delta_0\) conditionally on
\(X\), let \(\eta\in L^2(P_0)\) satisfy
\(E_{P_0}[\eta\mid X]=0\) and \(E_{P_0}[\Delta_0\eta\mid X]=0\). Let
\(\eta_M:=-M\vee\eta\wedge M\), and define
\[
\eta_M^\perp
:=
\eta_M
-E_{P_0}(\eta_M\mid X)
-
\frac{E_{P_0}(\Delta_0\eta_M\mid X)}{\sigma_0^2(X)}\Delta_0 .
\]
Then \(E_{P_0}[\eta_M^\perp\mid X]=0\) and
\[
E_{P_0}[\Delta_0\eta_M^\perp\mid X]
=
E_{P_0}[\Delta_0\eta_M\mid X]
-
\frac{E_{P_0}[\Delta_0\eta_M\mid X]}{\sigma_0^2(X)}
E_{P_0}[\Delta_0^2\mid X]
=0.
\]
The functions \(\eta_M^\perp\) are bounded because \(\eta_M\), \(\Delta_0\),
and \(\sigma_0^{-2}\) are bounded. Since \(\eta_M\to\eta\) in \(L^2(P_0)\),
conditional expectation is an \(L^2\)-contraction, and \(\Delta_0\) is bounded,
both correction terms converge to zero in \(L^2(P_0)\). Thus
\(\eta_M^\perp\to\eta\) in \(L^2(P_0)\).

Combining the approximations for \(u_0\), \(b\), \(a\Delta_0\), and \(\eta\)
gives a sequence in \((\mathcal S_H^{\rm M})^\infty\). Since \(T\) is fixed,
pooled componentwise convergence implies convergence of the trajectory scores
in \(L^2(\mathbb P_0)\).
\end{proof}

\begin{lemma}[Bounded product generators for the restricted tangent space]
\label{lemma::restrictedBoundedGenerators}
Under the conditions of Lemma~\ref{lemma::restrictedBoundedScoreDensity}, let
\(\ell\in(\mathcal S_H^{\rm M})^\infty\), set
\(v(Z):=a(X)\Delta_0(Z)+\eta(Z)\), and let
\[
h_a:=\mathcal T_0^{-1}\{a(X)\sigma_0^2(X)\}\in \overline H.
\]
Write \(D_h(Z):=\gamma V^\pi(h_a)(S')-h_a(X)\) and define
\[
c_\varepsilon(X)
:=-\frac{E_{P_0}\{v(Z)D_h(Z)\mid X\}}
{\sigma_0^2(X)+\varepsilon E_{P_0}\{\Delta_0(Z)D_h(Z)\mid X\}}.
\]
For sufficiently small \(|\varepsilon|\), define a time-homogeneous Markov law
by
\[
\begin{aligned}
\frac{d\mu_\varepsilon}{d\mu_0}(S_0)=1+\varepsilon u_0(S_0),\qquad
&\frac{dg_\varepsilon}{dg_0}(A\mid S)=1+\varepsilon b(A,S),\\
\frac{dK_\varepsilon}{dK_0}(Y,S'\mid X)
&=1+\varepsilon v(Z)+\varepsilon^2c_\varepsilon(X)\Delta_0(Z).
\end{aligned}
\]
Then \((\mathbb P_\varepsilon^\ell:\varepsilon)\) is a regular QMD submodel
through \(\mathbb P_0\) with score \(\ell\). Let
\(P_\varepsilon^\ell\) denote its induced pooled transition law. Moreover,
\[
\mathbb P_\varepsilon^\ell\in\mathfrak P_H^{\rm M},
\qquad
q_{P_\varepsilon^\ell}=q_0+\varepsilon h_a .
\]
\end{lemma}

\begin{proof}
Since \(v\), \(h_a\), and \(\Delta_0\) are bounded and
\(\sigma_0^2\ge c>0\), the denominator defining \(c_\varepsilon\) is bounded
away from zero for sufficiently small \(|\varepsilon|\). Thus
\(\sup_{|\varepsilon|\le\varepsilon_0}\|c_\varepsilon\|_\infty<\infty\)
for some \(\varepsilon_0>0\). The conditional centering restrictions,
including \(E_{P_0}(\Delta_0\mid X)=0\), normalize the displayed kernels.
Their likelihood ratios are positive and preserve the local-model bounds
for sufficiently small \(|\varepsilon|\), with those bounds chosen with
strict slack at \(\mathbb P_0\). Expanding the product trajectory likelihood
shows that the path is
regular QMD with score
\(u_0(S_0)+\sum_{t<T}\{b(A_t,S_t)+v(Z_t)\}=\ell(O)\).
By linearity of \(V^\pi\),
\[
Y+\gamma V^\pi(q_0+\varepsilon h_a)(S')
-q_0(X)-\varepsilon h_a(X)
=
\Delta_0(Z)+\varepsilon D_h(Z).
\]
Therefore
\[
\begin{aligned}
&E_{P_\varepsilon^\ell}\!\left[
Y+\gamma V^\pi(q_0+\varepsilon h_a)(S')
-q_0(X)-\varepsilon h_a(X)
\mid X
\right]\\
&\quad =
E_{P_0}\!\left[
\{1+\varepsilon v(Z)+\varepsilon^2c_\varepsilon(X)\Delta_0(Z)\}
\{\Delta_0(Z)+\varepsilon D_h(Z)\}
\mid X
\right]
\\
&\quad =
\varepsilon^2\Big[
E_{P_0}\{vD_h\mid X\}
+c_\varepsilon(X)
\{\sigma_0^2(X)+\varepsilon E_{P_0}(\Delta_0D_h\mid X)\}
\Big]
=0,
\end{aligned}
\]
because \(E_{P_0}(\Delta_0\mid X)=0\),
\(E_{P_0}(D_h\mid X)=-\mathcal T_0h_a\), and
\(E_{P_0}(\Delta_0v\mid X)=a\sigma_0^2=\mathcal T_0h_a\).
The definition of \(c_\varepsilon\) cancels the remaining terms.
This verifies the Bellman equation under
\(P_\varepsilon^\ell\), so
\(q_0+\varepsilon h_a\in \overline H\) is bounded and solves it. If \(q_1\)
and \(q_2\) are two bounded Bellman solutions under \(P_\varepsilon^\ell\),
then
\(d:=q_1-q_2\) satisfies
\[
d(X)=\gamma E_{P_\varepsilon^\ell}\{V^\pi(d)(S')\mid X\}.
\]
Because \(\pi(\cdot\mid s)\) is a probability law,
\(\|V^\pi(d)\|_\infty\le \|d\|_\infty\). Thus
\(\|d\|_\infty\le\gamma\|d\|_\infty\), and \(\gamma<1\) gives \(d=0\).
Hence \(q_{P_\varepsilon^\ell}=q_0+\varepsilon h_a\in \overline H\), so
\(\mathbb P_\varepsilon^\ell\in\mathfrak P_H^{\rm M}\).
\end{proof}

\subsection{Projection agreement}

\begin{lemma}[Projection agreement on the restricted model]
\label{lemma::restrictedProjectionAgreement}
Suppose Condition~\ref{cond::bounded} holds for the fixed weight \(w\).
For every \(\mathbb P\in\mathfrak P_H^{\rm M}\), with pooled transition law
\(P\),
\[
q_{P,H,w}=q_P,\qquad
e_{P,H,w}=0,\qquad
\Psi_{H,w}(\mathbb P)=\Psi(\mathbb P).
\]
\end{lemma}

\begin{proof}
For \(\mathbb P\in\mathfrak P_H^{\rm M}\),
Definition~\ref{def::restrictedLocalModel} gives
\(q_P\in \overline{H}\). By the definition of \(q_P\), the Bellman
equation under \(P\) gives
\[
r_P(X)=\mathcal T_P(q_P)(X).
\]
Thus the objective in \eqref{eqn::projection} has value zero at \(q_P\). The
restricted inverse in Lemma~\ref{lemma::localInversePersistence} gives
uniqueness of the minimizer, so \(q_{P,H,w}=q_P\). Hence
\[
e_{P,H,w}=\mathcal T_P(q_{P,H,w})-r_P=0.
\]
The target identity follows from the definition of \(\Psi_{H,w}\):
\[
\Psi_{H,w}(\mathbb P)=E_{\mathbb P}\{m(S_0,q_{P,H,w})\}
=E_{\mathbb P}\{m(S_0,q_P)\}
=\Psi(\mathbb P).
\]
\end{proof}

\par\medskip

For a fixed weight \(w\), define
\[
\Phi^w_{0,H}(O)
=
m(S_0,q_0)-\psi_0
+\frac1T\sum_{t=0}^{T-1}
w(X_t)\tau_{0,H,w}(X_t)\Delta_0(Z_t).
\]

\begin{theorem}[Weighted restricted-model influence functions]
\label{theorem::weightedIFrestricted}
Suppose Conditions~\ref{cond::bounded} and~\ref{cond::boundednuis}
hold for the fixed weight \(w\), Condition~\ref{cond::restrictedModelRegularity}
holds,
and let \(\mathfrak P_H^{\rm M}\) be the local restricted model in
Definition~\ref{def::restrictedLocalModel}. Then \(\Phi^w_{0,H}\)
is an influence function for \(\Psi:\mathfrak P_H^{\rm M}\to\mathbb R\) at
\(\mathbb P_0\).
\end{theorem}

\begin{proof}
Let \((\mathbb P_\varepsilon:\varepsilon)\subset\mathfrak P_H^{\rm M}\) be
any regular QMD submodel through \(\mathbb P_0\) with score \(\ell\). By
Lemma~\ref{lemma::restrictedProjectionAgreement},
\(\Psi(\mathbb P_\varepsilon)=\Psi_{H,w}(\mathbb P_\varepsilon)\) for all sufficiently small
\(\varepsilon\). Applying Theorem~\ref{theorem::projectionPD} to this submodel
therefore gives
\[
\left.\frac{d}{d\varepsilon}\Psi(\mathbb P_\varepsilon)\right|_{\varepsilon=0}
=
E_{\mathbb P_0}\{\Phi^*_{0,H,w}(O)\ell(O)\}.
\]
At \(P_0\), Condition~\ref{cond::restrictedModelRegularity} and
Lemma~\ref{lemma::restrictedProjectionAgreement} give \(q_{0,H,w}=q_0\) and
\(e_{0,H,w}=0\). Substituting these identities into
\(\Phi^*_{0,H,w}\) gives the displayed \(\Phi^w_{0,H}\).
This proves that \(\Phi^w_{0,H}\) is an influence function for \(\Psi\) in
\(\mathfrak P_H^{\rm M}\).
\end{proof}

\begin{lemma}[Trajectory variance under correct specification]
\label{lemma::correctSpecificationTrajectoryVariance}
Under the conditions of Theorem~\ref{theorem::weightedIFrestricted}, for every
fixed \(w\),
\[
\operatorname{Var}_{\mathbb P_0}\{\Phi^w_{0,H}(O)\}
=\operatorname{Var}_{\mathbb P_0}\{m(S_0,q_0)\}
+\frac1T E_{P_0}[\{w(X)\tau_{0,H,w}(X)\}^2\sigma_0^2(X)].
\]
\end{lemma}

\begin{proof}
Let \(\xi_t=w(X_t)\tau_{0,H,w}(X_t)\Delta_0(Z_t)\). The Markov property and
the Bellman equation imply
\(E_{\mathbb P_0}(\xi_t\mid S_0,A_0,Y_0,\ldots,S_t,A_t)=0\). Hence the cross-time
covariances and the covariances with the baseline term vanish. Finally,
\[
\operatorname{Var}_{\mathbb P_0}\!\left(\frac1T\sum_{t=0}^{T-1}\xi_t\right)
=\frac1{T^2}\sum_{t=0}^{T-1}E_{\mathbb P_0}(\xi_t^2)
=\frac1T E_{P_0}[\{w\tau_{0,H,w}\}^2\sigma_0^2],
\]
which proves the result.
\end{proof}

\subsection{Canonical gradient}

For \(w_{\rm opt}(X)=\sigma_0^{-2}(X)\), write
\(\alpha^{\rm eff}_{0,H}:=\alpha_{0,H,w_{\rm opt}}\). Its weighted Riesz
equation is
\begin{equation}
E_{\mathbb P_0}\{m(S_0,h)\}
=
E_{P_0}\!\left[
\frac{
\mathcal T_0(\alpha^{\rm eff}_{0,H})(X)\,
\mathcal T_0(h)(X)
}{\sigma_0^2(X)}
\right],
\qquad h\in \overline H.
\label{eqn::weightedRieszRestricted}
\end{equation}

\begin{theorem}[Canonical gradient under the restricted model]
\label{theorem::EIFrestricted}
Let \(\mathfrak P_H^{\rm M}\) be the local restricted model in
Definition~\ref{def::restrictedLocalModel}. Suppose
Conditions~\ref{cond::bounded} and~\ref{cond::boundednuis} hold with
\(w=w_{\rm opt}\), and
Condition~\ref{cond::restrictedModelRegularity} holds. Then
\(\Psi:\mathfrak P_H^{\rm M}\to\mathbb R\) is pathwise differentiable at
\(\mathbb P_0\), and
its canonical gradient is
\[
\Phi^{\rm eff}_{0,H}(O)
=
m(S_0,q_0)-\psi_0
+\frac1T\sum_{t=0}^{T-1}
\frac{\mathcal T_0(\alpha^{\rm eff}_{0,H})(X_t)}
{\sigma_0^2(X_t)}
\Delta_0(Z_t).
\]
\end{theorem}

\begin{proof}
Theorem~\ref{theorem::weightedIFrestricted} with \(w=w_{\rm opt}\) shows that
\(\Phi^{\rm eff}_{0,H}\) is an influence function. It remains to verify that
this influence function belongs to the tangent space
\(\dot{\mathfrak P}^{\rm M}_{H,0}\).
Lemma~\ref{lemma::restrictedBoundedGenerators} shows that every element of
\((\mathcal S_H^{\rm M})^\infty\) is the score of a regular QMD submodel
contained in \(\mathfrak P_H^{\rm M}\). Therefore
\(\overline{(\mathcal S_H^{\rm M})^\infty}^{L^2(\mathbb P_0)}
\subseteq\dot{\mathfrak P}^{\rm M}_{H,0}\).
By Lemma~\ref{lemma::restrictedBoundedScoreDensity},
\(\overline{\mathcal S_H^{\rm M}}^{L^2(\mathbb P_0)}
\subseteq\dot{\mathfrak P}^{\rm M}_{H,0}\).

The first component, \(m(S_0,q_0)-\psi_0\), is square integrable by
Condition~\ref{cond::targetMapContinuity} and
Condition~\ref{cond::restrictedModelRegularity}; it has mean zero by the
definition of \(\psi_0\). Hence it is the baseline-score component in
\(\mathcal S_H^{\rm M}\).
For the residual component, set
\[
a_{\rm eff}(X):=
\frac{\mathcal T_0(\alpha^{\rm eff}_{0,H})(X)}{T\sigma_0^2(X)}.
\]
By the Bellman equation, \(E_{P_0}[\Delta_0(Z)\mid X]=0\), and therefore
\[
E_{P_0}\{a_{\rm eff}(X)\Delta_0(Z)\mid X\}=0.
\]
The definition of \(a_{\rm eff}\) and \eqref{eqn::weightedRieszRestricted} give
\[
E_{P_0}[\Delta_0(Z)\{a_{\rm eff}(X)\Delta_0(Z)\}\mid X]
=
a_{\rm eff}(X)\sigma_0^2(X)
=
T^{-1}\mathcal T_0(\alpha^{\rm eff}_{0,H})(X)
\in\mathcal T_0(\overline H).
\]
Thus the time-summed residual component is a valid direction in
\(\mathcal S_H^{\rm M}\), with \(b=\eta=0\). Hence
\(\Phi^{\rm eff}_{0,H}\in\dot{\mathfrak P}^{\rm M}_{H,0}\).
Since an influence function that lies in the tangent space is the canonical
gradient \citep[Chapter~3]{bickel1993efficient}, the displayed function is the
efficient influence function.
\end{proof}

The formula simplifies in the finite-dimensional linear case. Suppose
\(H=\{\theta^\top\varphi:\theta\in\mathbb R^k\}\), and define the
Bellman-shifted feature
\[
\varphi_\gamma(a,s)
:=
\varphi(a,s)
-\gamma E_{P_0}\!\left[
\int \varphi(a',S')\pi(a'\mid S')\nu(da')
\mid A=a,S=s
\right],
\]
with expectations of vectors taken component-wise. If
\(m(s,q)=V^\pi(q)(s)\) is the policy-value functional and the nonparametric
occupancy ratio \(d_0\) in \eqref{eqn::stateratio} exists, then for some
\(\beta_\omega\),
\[
\mathcal T_0(\alpha_{0,H,w})(a,s)=\beta_\omega^\top\varphi_\gamma(a,s),
\]
where
\[
E_{P_0}\!\left[
w(X)\varphi_\gamma(X)\varphi_\gamma(X)^\top
\right]\beta_\omega
=
E_{P_0}[\varphi_\gamma(X)d_0(X)].
\]
The corresponding restricted-model influence function is
\[
\Phi^w_{0,H}(O)
=m(S_0,q_0)-\psi_0
+\frac1T\sum_{t=0}^{T-1}
w(X_t)\beta_\omega^\top\varphi_\gamma(X_t)\Delta_0(Z_t).
\]
For \(w=w_{\rm opt}\), \(\beta_{\rm eff}\) solves
\[
E_{P_0}\!\left[
\frac{\varphi_\gamma(X)\varphi_\gamma(X)^\top}{\sigma_0^2(X)}
\right]\beta_{\rm eff}
=
E_{P_0}[\varphi_\gamma(X)d_0(X)].
\]
Thus the efficient residual multiplier is
\[
\frac{\beta_{\rm eff}^\top\varphi_\gamma(X)}{\sigma_0^2(X)}.
\]

\section{Variational representation of Bellman projections}

\label{appendix::variational}
\subsection{Main result}
Let \(Z=(S,A,Y,S')\) and \(X=(S,A)\). Let
\[
\mathcal S_w
:=
L^2_w(P_{0,X})
=
\left\{s:\ E_{P_0}\{w(X)s(X)^2\}<\infty\right\}.
\]
For an observable offset \(u(Z)\), write
\[
\bar u_0(X):=E_{P_0}\{u(Z)\mid X\}.
\]
For \(h\in\overline H\) and \(s\in\mathcal S_w\), define
\[
\mathcal L_{u,m}(h,s)
:=
E_{P_0}\!\left[
w(X)s(X)\{h(X)-\gamma V^\pi(h)(S')-u(Z)\}
-\frac{1}{2}w(X)s(X)^2
\right]-E_{\mathbb P_0}\{m(S_0,h)\},
\]
and
\[
\mathcal D_{u,m}(s,h)
:=
E_{P_0}\!\left[
\frac{1}{2}w(X)s(X)^2
-w(X)s(X)\{h(X)-\gamma V^\pi(h)(S')-u(Z)\}
\right]+E_{\mathbb P_0}\{m(S_0,h)\}.
\]

\begin{theorem}[Primal--dual characterization of weighted Bellman projections]
\label{lemma::genericBellmanMinmax}
Suppose \(\overline H\) is linear, Condition~\ref{cond::bounded} holds at \(P_0\),
\(E_{P_0}\{w(X)u(Z)^2\}<\infty\), and
\(h\mapsto E_{\mathbb P_0}\{m(S_0,h)\}\) is linear and continuous with respect to
\(\|\mathcal T_0 h\|_{w,P_0}\). There exists a unique
\(h_{0,u,m}\in\overline H\) satisfying
\[
E_{P_0}\!\left[
w(X)\{\mathcal T_0(h_{0,u,m})(X)-\bar u_0(X)\}
\mathcal T_0(g)(X)
\right]
=
E_{\mathbb P_0}\{m(S_0,g)\},
\qquad g\in \overline H.
\]
Define
\[
s_{0,u,m}(X)
:=
\mathcal T_0(h_{0,u,m})(X)-\bar u_0(X).
\]
The following identities hold:
\[
h_{0,u,m}
=
\argmin_{h\in\overline H}\sup_{s\in\mathcal S_w}
\mathcal L_{u,m}(h,s),
\qquad
s_{0,u,m}
=
\argmax_{s\in\mathcal S_w}
\mathcal L_{u,m}(h_{0,u,m},s),
\]
and
\[
s_{0,u,m}
=
\argmin_{s\in\mathcal S_w}\sup_{h\in\overline H}
\mathcal D_{u,m}(s,h),
\qquad
h_{0,u,m}
\in
\argmax_{h\in\overline H}
\mathcal D_{u,m}(s_{0,u,m},h).
\]
\end{theorem}

\paragraph{Weighted Bellman--Riesz representer.}
Taking \(u(Z)\equiv 0\) gives
\[
h_{0,u,m}=\alpha_{0,H,w},
\qquad
s_{0,u,m}=\mathcal T_0(\alpha_{0,H,w}).
\]
Thus Theorem~\ref{lemma::genericBellmanMinmax} gives the primal--dual
characterization of the weighted Bellman--Riesz pair.

\paragraph{Weighted projected \(Q\)-function.}
Taking \(u(Z)=Y\) and \(m(S_0,h)\equiv0\) gives
\[
\bar u_0(X)=E_{P_0}(Y\mid X)=r_0(X),
\qquad
h_{0,u,m}=q_{0,H,w},
\qquad
s_{0,u,m}
=
\mathcal T_0(q_{0,H,w})-r_0.
\]
Thus Theorem~\ref{lemma::genericBellmanMinmax} gives the primal--dual
characterization of the weighted Bellman projection.

\subsection{Proof of Theorem \ref{lemma::genericBellmanMinmax}}

\begin{proof}[Proof of Theorem~\ref{lemma::genericBellmanMinmax}]
Write
\[
L_m(h):=E_{\mathbb P_0}\{m(S_0,h)\},
\qquad
\|g\|_{w,P_0}^2:=E_{P_0}\{w(X)g(X)^2\},
\]
and let
\[
\mathcal M_0:=\mathcal T_0(\overline H)\subset L^2_w(P_{0,X}).
\]
By Condition~\ref{cond::bounded}, \(\mathcal M_0\) is closed in
\(L^2_w(P_{0,X})\), and \(\mathcal T_0\) is injective on \(\overline H\). The
map
\[
\ell(\mathcal T_0 h)
:=
E_{P_0}\{w(X)\bar u_0(X)\mathcal T_0(h)(X)\}+L_m(h)
\]
is therefore a well-defined continuous linear functional on \(\mathcal M_0\).
By the Riesz representation theorem, there exists a unique
\(g_{0,u,m}\in\mathcal M_0\) such that
\[
E_{P_0}\{w(X)g_{0,u,m}(X)\mathcal T_0(h)(X)\}
=
E_{P_0}\{w(X)\bar u_0(X)\mathcal T_0(h)(X)\}+L_m(h),
\qquad h\in \overline H.
\]
Since \(g_{0,u,m}\in\mathcal M_0\), there exists
\(h_{0,u,m}\in\overline H\) such that
\[
\mathcal T_0(h_{0,u,m})=g_{0,u,m}.
\]
Injectivity of \(\mathcal T_0\) on \(\overline H\) gives uniqueness of
\(h_{0,u,m}\).
Equivalently,
\[
E_{P_0}\!\left[
w(X)\{\mathcal T_0(h_{0,u,m})(X)-\bar u_0(X)\}
\mathcal T_0(h)(X)
\right]
=
L_m(h),
\qquad h\in \overline H.
\]

For \(h\in\overline H\), define
\[
B_{u,h}:=h(X)-\gamma V^\pi(h)(S')-u(Z).
\]
For any \(s\in\mathcal S_w\), iterated expectation gives
\[
E_{P_0}\{w(X)s(X)B_{u,h}\}
=
E_{P_0}\!\left[
w(X)s(X)\{\mathcal T_0(h)(X)-\bar u_0(X)\}
\right].
\]
Completing the square yields
\[
\begin{aligned}
& E_{P_0}\!\left[
w(X)s(X)B_{u,h}-\frac{1}{2}w(X)s(X)^2
\right]
\\
&\qquad =
\frac{1}{2}
E_{P_0}\!\left[
w(X)\{\mathcal T_0(h)(X)-\bar u_0(X)\}^2
\right]
-
\frac{1}{2}
E_{P_0}\!\left[
w(X)
\{s(X)-(\mathcal T_0(h)(X)-\bar u_0(X))\}^2
\right].
\end{aligned}
\]
Therefore
\[
\sup_{s\in\mathcal S_w}
\mathcal L_{u,m}(h,s)
=
\frac{1}{2}
E_{P_0}\!\left[
w(X)\{\mathcal T_0(h)(X)-\bar u_0(X)\}^2
\right]
-
L_m(h),
\]
with unique maximizer
\[
s_h(X)=\mathcal T_0(h)(X)-\bar u_0(X).
\]
Thus the primal objective is
\[
J(h)
=
\frac{1}{2}
E_{P_0}\!\left[
w(X)\{\mathcal T_0(h)(X)-\bar u_0(X)\}^2
\right]
-
L_m(h).
\]
For \(\delta=h-h_{0,u,m}\), the defining equation for \(h_{0,u,m}\) gives
\[
L_m(\delta)
=
E_{P_0}\!\left[
w(X)\{\mathcal T_0(h_{0,u,m})(X)-\bar u_0(X)\}
\mathcal T_0(\delta)(X)
\right].
\]
Hence
\[
J(h)-J(h_{0,u,m})
=
\frac{1}{2}
E_{P_0}\!\left[
w(X)\{\mathcal T_0(\delta)(X)\}^2
\right]
\geq 0.
\]
Equality implies \(\mathcal T_0(\delta)=0\), and therefore \(\delta=0\) by the
restricted inverse condition. This proves
\[
h_{0,u,m}
=
\argmin_{h\in\overline H}\sup_{s\in\mathcal S_w}
\mathcal L_{u,m}(h,s),
\]
and the corresponding maximizer is
\[
s_{0,u,m}(X)
=
\mathcal T_0(h_{0,u,m})(X)-\bar u_0(X).
\]

For the dual characterization, for any \(s\in\mathcal S_w\),
\[
\begin{aligned}
\sup_{h\in\overline H}\mathcal D_{u,m}(s,h)
&=
E_{P_0}\!\left[
\frac{1}{2}w(X)s(X)^2+w(X)s(X)\bar u_0(X)
\right]
\\
&\qquad
+
\sup_{h\in\overline H}
\left[
L_m(h)-E_{P_0}\{w(X)s(X)\mathcal T_0(h)(X)\}
\right].
\end{aligned}
\]
The bracketed term is linear in \(h\). Because \(\overline H\) is linear, the
supremum is finite exactly when
\[
E_{P_0}\{w(X)s(X)\mathcal T_0(h)(X)\}=L_m(h),
\qquad h\in\overline H.
\]
If this condition fails for some \(h\), replacing \(h\) by \(t h\) and choosing
the sign of \(t\) makes the supremum infinite. Hence the dual problem is
equivalent to minimizing
\[
E_{P_0}\!\left[
\frac{1}{2}w(X)s(X)^2+w(X)s(X)\bar u_0(X)
\right]
\]
over all \(s\in\mathcal S_w\) satisfying the above moment equations. The element
\(s_{0,u,m}\) satisfies these equations by the defining equation for
\(h_{0,u,m}\).

Let \(s\) be any other feasible element and set \(d=s-s_{0,u,m}\). Then
\[
E_{P_0}\{w(X)d(X)\mathcal T_0(h)(X)\}=0,
\qquad h\in\overline H.
\]
In particular, taking \(h=h_{0,u,m}\) gives
\[
E_{P_0}\{w(X)d(X)\mathcal T_0(h_{0,u,m})(X)\}=0.
\]
Since
\[
s_{0,u,m}(X)
=
\mathcal T_0(h_{0,u,m})(X)-\bar u_0(X),
\]
we have
\[
\begin{aligned}
& E_{P_0}\!\left[
\frac{1}{2}w(X)s(X)^2+w(X)s(X)\bar u_0(X)
\right]
-
E_{P_0}\!\left[
\frac{1}{2}w(X)s_{0,u,m}(X)^2
+w(X)s_{0,u,m}(X)\bar u_0(X)
\right]
\\
&\qquad =
\frac{1}{2}
E_{P_0}\!\left[
w(X)d(X)^2
\right]
+
E_{P_0}\!\left[
w(X)d(X)\left\{s_{0,u,m}(X)+\bar u_0(X)\right\}
\right]
\\
&\qquad =
\frac{1}{2}
E_{P_0}\!\left[
w(X)d(X)^2
\right]
+
E_{P_0}\{w(X)d(X)\mathcal T_0(h_{0,u,m})(X)\}
\\
&\qquad =
\frac{1}{2}
E_{P_0}\!\left[
w(X)\{s(X)-s_{0,u,m}(X)\}^2
\right]
\geq 0.
\end{aligned}
\]
Thus \(s_{0,u,m}\) is the unique dual minimizer. Finally, feasibility of
\(s_{0,u,m}\) implies
\[
L_m(h)-E_{P_0}\{w(X)s_{0,u,m}(X)\mathcal T_0(h)(X)\}=0,
\qquad h\in\overline H.
\]
Therefore \(h_{0,u,m}\) attains the inner supremum in the dual display.
\end{proof}

\subsection{Profiled finite-dimensional and RKHS criteria}
\label{appendix::profiledminimax}

The minimax criteria in Section~\ref{subsection::nuisance} place the
Bellman-image nuisance in the outer minimization and the \(H\)-valued function
in the inner maximization. The profiled criteria below keep this same
orientation: \(s\in L^2_w(P_{0,X})\) remains the outer variable, and the
maximization over \(h\in H\) is the part being removed. This is useful
computationally because the inner problem is linear in \(h\). Profiling it out
gives either a set of Bellman moment constraints or, after quadratic
regularization, a single penalized loss for the outer nuisance.

For \(h\in\overline H\), write
\[
B_h(Z):=h(X)-\gamma V^\pi(h)(S').
\]
At the population level, \(H\) may be replaced by its closure \(\overline H\);
in the finite-dimensional and RKHS examples below, \(H\) denotes the closed
regularized working space.
For an outer function \(s\in L^2_w(P_{0,X})\), define the linear functional
\[
\ell_s(h)
:=
E_{\mathbb P_0}\{m(S_0,h)\}
-
E_{P_0}\{w(X)s(X)B_h(Z)\},
\qquad h\in H.
\]
The dual criterion in Theorem~\ref{lemma::genericBellmanMinmax} can be written
as
\[
E_{P_0}\left\{\frac12 w(X)s(X)^2+w(X)s(X)u(Z)\right\}
+
\sup_{h\in H}\ell_s(h).
\]
If \(H\) is a linear space and no penalty is imposed on \(h\), the supremum is
zero when \(\ell_s(h)=0\) for all \(h\in H\), and is infinite otherwise.
Thus the unpenalized profiled criterion is the quadratic outer loss subject to
the Bellman moment equations
\[
E_{P_0}\{w(X)s(X)B_h(Z)\}=E_{\mathbb P_0}\{m(S_0,h)\},
\qquad h\in H.
\]
For the projected \(Q\)-function, \(u(Z)=Y\), \(m\equiv0\), and
\(s=e\); the constraint is the weighted Bellman normal equation. For the
Bellman--Riesz problem, \(u\equiv0\), \(m\) is the target-map integrand, and
\(s=\tau\); the constraint is the weighted Riesz equation.

A regularized version replaces the hard moment equations by a dual-norm
penalty. Let \(\|\cdot\|_H\) be the norm used to regularize the inner
\(H\)-valued function, and let
\[
\|\ell_s\|_{H^*}
:=
\sup_{\|h\|_H\le1}|\ell_s(h)|
\]
be the corresponding dual norm. For \(\lambda>0\),
\[
\sup_{h\in H}
\left\{
\ell_s(h)-\frac{\lambda}{2}\|h\|_H^2
\right\}
=
\frac{1}{2\lambda}\|\ell_s\|_{H^*}^2.
\]
The regularized profiled population criterion is therefore
\[
\mathcal Q_{u,m,\lambda}(s)
=
E_{P_0}\left\{\frac12 w(X)s(X)^2+w(X)s(X)u(Z)\right\}
+
\frac{1}{2\lambda}\|\ell_s\|_{H^*}^2.
\]
This criterion has the same formal structure for \(e\) and \(\tau\); only the
choice of \(u\) and \(m\) changes.

When \(H\) is finite-dimensional, write
\[
H=\{x\mapsto \theta^\top\phi(x):\theta\in\mathbb R^K\},
\qquad
\phi(x)=(\phi_1(x),\ldots,\phi_K(x))^\top ,
\]
and define the observed Bellman feature
\[
\phi_B(Z)
:=
\phi(X)-\gamma V^\pi(\phi)(S'),
\]
where \(V^\pi(\phi)(S')\) is evaluated componentwise. Linearity of \(m\) gives
a vector \(m_\phi(S_0)\) such that
\(m(S_0,\theta^\top\phi)=\theta^\top m_\phi(S_0)\). Then
\[
\ell_s(\theta^\top\phi)
=
\theta^\top
\left[
E_{\mathbb P_0}\{m_\phi(S_0)\}
-
E_{P_0}\{w(X)s(X)\phi_B(Z)\}
\right].
\]
If \(\|h\|_H^2=\theta^\top R\theta\) for a positive definite matrix \(R\),
the profiled penalty is
\[
\frac{1}{2\lambda}
\left[
E_{\mathbb P_0}\{m_\phi(S_0)\}
-
E_{P_0}\{w(X)s(X)\phi_B(Z)\}
\right]^\top
R^{-1}
\left[
E_{\mathbb P_0}\{m_\phi(S_0)\}
-
E_{P_0}\{w(X)s(X)\phi_B(Z)\}
\right].
\]
The unpenalized criterion is recovered by replacing this penalty with the
constraint that the bracketed vector is zero.

The empirical version averages Bellman terms over the \(nT\) transitions and
the target term over the \(n\) baseline states. Let \(\Phi_B\) be the
\(nT\times K\) matrix with rows \(\phi_B(Z_{i,t})^\top\), and define the
corresponding diagonal matrix \(W\) and vectors \(s\) and \(u\), indexed by
\((i,t)\). Define
\[
\widehat a_s
:=
\frac1n\sum_{i=1}^n m_\phi(S_{i,0})
-
\frac{1}{nT}\Phi_B^\top W s .
\]
The regularized single-level empirical criterion is
\[
\widehat{\mathcal Q}_{u,m,\lambda}(s)
=
\frac{1}{2nT}s^\top W s
+
\frac{1}{nT}s^\top W u
+
\frac{1}{2\lambda}\widehat a_s^\top R^{-1} \widehat a_s .
\]
When \(\Phi_B^\top W\Phi_B/(nT)\) is full rank, choosing
\(R=\Phi_B^\top W\Phi_B/(nT)\) gives the weighted least-squares geometry for the
Bellman-feature moments. In numerical work one may instead use
\(R=\Phi_B^\top W\Phi_B/(nT)+\rho I_K\) with \(\rho>0\).

For an RKHS \(H=\mathcal H_k\) with a measurable, bounded kernel \(k\), the
same construction uses the RKHS norm; assume the kernel-mean integrals below
exist as elements of \(\mathcal H_k\). Define the empirical linear functional
\[
\widehat\ell_s(h)
=
\frac1n\sum_{i=1}^n m(S_{i,0},h)
-P_n\{w(X)s(X)B_h(Z)\}.
\]
Assume that \(h\mapsto m(S_{i,0},h)\) has an RKHS representer \(\xi_i\), and let
\[
\eta_{i,t}(\cdot)
:=
k(X_{i,t},\cdot)
-
\gamma\int k((a,S_{i,t+1}),\cdot)\pi(da\mid S_{i,t+1})
\]
be the representer of \(h\mapsto B_h(Z_{i,t})\). Then
\[
\widehat\ell_s(h)
=
\left\langle h,\,
\frac{1}{n}\sum_{i=1}^n\xi_i
-\frac{1}{nT}\sum_{i=1}^n\sum_{t=0}^{T-1}
w(X_{i,t})s(X_{i,t})\eta_{i,t}
\right\rangle_{\mathcal H_k}.
\]
The regularized profiled empirical criterion is therefore
\[
\frac{1}{2nT}s^\top W s
+
\frac{1}{nT}s^\top W u
+
\frac{1}{2\lambda}\|\widehat\ell_s\|_{\mathcal H_k^*}^2 .
\]
To compute the dual norm, define Gram matrices
\[
K_{\eta\eta,(i,t),(j,r)}:=\langle\eta_{i,t},\eta_{j,r}\rangle_{\mathcal H_k},\qquad
K_{\xi\eta,i,(j,r)}:=\langle\xi_i,\eta_{j,r}\rangle_{\mathcal H_k},\qquad
K_{\xi\xi,ij}:=\langle\xi_i,\xi_j\rangle_{\mathcal H_k}.
\]
With \(\mathbf 1_n\) denoting the \(n\)-vector of ones,
\[
\|\widehat\ell_s\|_{\mathcal H_k^*}^2
=
\frac1{n^2}\mathbf 1_n^\top K_{\xi\xi}\mathbf 1_n
-\frac{2}{n^2T}\mathbf 1_n^\top K_{\xi\eta}Ws
+\frac1{n^2T^2}s^\top WK_{\eta\eta}Ws.
\]
This is the kernel analogue of the finite-dimensional dual-norm penalty. If
the outer class for \(s\) is also linear or finite-dimensional, minimizing the
displayed criterion reduces to a ridge-type linear system in the coefficients
of \(s\). The optimized outer variable remains \(s=e\) for the \(Q\)-criterion
and \(s=\tau\) for the Bellman--Riesz criterion.

\section{Sieve estimation and model approximation error}
\label{appendix:sieveapprox}
The sieve, adaptive-model-selection, and data-fusion extensions below retain
the one-transition state--action functional and are stated for \(T=1\).
In this special case, \(O=Z\), so the trajectory and one-transition laws
coincide; we write \(P\) and \(P_n\) throughout. The fixed-weight projection
results have direct state--action analogues obtained by replacing
\(E_{\mathbb P}\{m(S_0,h)\}\) with \(E_P\{m(X,h)\}\) in the target, Riesz
equation, and continuity condition. The proofs are unchanged, with the
baseline score term in Lemma~\ref{lemma::trajectoryLifting} replaced by
\(E_{P_0}\{\ell(Z)m(X,h)\}\).

Throughout this section and Appendix~\ref{sec::ADMLE}, \(w\equiv1\). We
abbreviate \(q_{P,H}:=q_{P,H,1}\),
\(\alpha_{P,H}:=\alpha_{P,H,1}\), and
\(\Psi_H:=\Psi_{H,1}\).
Sieve methods combine the finite-sample stability of parametric models with
the flexibility of nonparametric function approximation
\citep{shen1997methods,newey1997convergence,spnpsieve}. Rather than fixing a
single semiparametric restriction \(H\), one may select the \(Q\)-function model from a
sequence of linear spaces
\[
H_1\subset H_2\subset\cdots\subset H_k\subset\cdots\subset H_\infty:=H
\]
of increasing complexity. For example, \(H_k\) could be a finite-dimensional
series space spanned by trigonometric polynomials \citep{JacksonOnAB}, splines
\citep{Gordon1974BSPLINECA}, or wavelet bases \citep{Antoniadis_2007}. It
could also be infinite-dimensional, such as a space of additively decomposable
functions in \(k\) variables \citep{stone1985additive, hastie1987generalized,
zhang2022regression}.

Given \(H_n:=H_{k(n)}\) for a possibly data-dependent sequence \(k(n)\), a
sieve DRL estimator of \(\psi_0\) is constructed as in
Section~\ref{section::estmeth::riesz}, yielding \(\psi_{n,H_n}\). The next
theorem shows that the corresponding misspecification bias
\(\Psi_{H_n}(P_0)-\Psi(P_0)\) is second order in the sieve approximation error.
Appendix~\ref{sec::ADMLE} gives the asymptotic theory for sieve DRL estimators.
Let \(H_0\) be a linear oracle subspace with \(q_0\in\overline H_0\). For any
linear \(H_n\), let \(H_{n,0}\) be a linear supermodel containing \(H_n+H_0\),
where one may take the algebraic sum
\(H_n+H_0:=\{h_n+h_0:h_n\in H_n,\ h_0\in H_0\}\).

\begin{theorem}[Model approximation error]
\label{theorem::oraclebiassieve}
Suppose Condition~\ref{cond::bounded} holds for \(H_n\) and
\(H_{n,0}\), conditionally on the training sample when \(H_n\) is random.
Then
\[
\Psi_{H_n}(P_0) - \Psi(P_0)
=
-\left\langle
    \mathcal{T}_0(\alpha_{0,H_n}) - \mathcal{T}_0(\alpha_{0,H_{n,0}}),
    \mathcal{T}_0(q_{0,H_n}) - \mathcal{T}_0(q_{0})
\right\rangle_{P_0}.
\]
\end{theorem}
The identity expresses the approximation bias as the product of the
\(Q\)-function and Riesz approximation errors. In particular, it applies to a
fixed, possibly misspecified class \(H_n=H\) whenever
Condition~\ref{cond::bounded} also holds for a supermodel containing
\(H+H_0\).

\section{Adaptive model selection and sieve DRL}
\label{sec::ADMLEmain}
\label{sec::ADMLE}

\subsection{General approach}

Selecting an appropriate semiparametric restriction involves a bias--variance tradeoff: an
overly restrictive model can introduce approximation bias, while a fully
nonparametric model can have a large efficiency bound. Adaptive DML (ADML)
\citep{van2023adaptive} addresses this tradeoff by combining debiased
estimation with data-dependent model selection. Here, we use ADML as a template
for Markov decision processes. The resulting adaptive DRL (ADRL) estimator
selects a semiparametric restriction for the \(Q\)-function \(q_0\) and then applies
semiparametric DRL. When the selected model stabilizes to an oracle submodel
and the approximation error is sufficiently small, the estimator has an oracle
influence-function expansion.

Let \(H_n\subseteq H\) be a data-dependent semiparametric restriction for the \(Q\)-function
\(q_P\). ADRL assumes that \(H_n\) converges, in the Bellman projection and
Riesz-representer senses stated below, to a fixed but unknown oracle submodel
\(H_0\subseteq H\) determined by \(P_0\). Suppose we have estimators
\(q_{n,H_n}\in H_n\) and \(\alpha_{n,H_n}\in H_n\) for \(q_{0,H_n}\) and
\(\alpha_{0,H_n}\), respectively. The ADRL estimator is
\[
\begin{aligned}
\psi_{n,H_n}
&=\frac{1}{n}\sum_{i=1}^n m(X_i,q_{n,H_n})\\
&\quad+\frac{1}{n}\sum_{i=1}^n
\widehat{\mathcal T}_n(\alpha_{n,H_n})(A_i,S_i)
\{Y_i+\gamma V^\pi(q_{n,H_n})(S_i')-q_{n,H_n}(A_i,S_i)\},
\end{aligned}
\]
which targets the data-adaptive working parameter \(\Psi_{H_n}(P_0)\). Unlike
the model-robust estimator \(\psi_{n,H}^*\) in
Section~\ref{section::estmeth::riesz}, this estimator omits the projection
residual correction because the theory below requires the model approximation
error to be \(o_p(n^{-1/2})\). Under the stated stability and approximation
conditions, \(\psi_{n,H_n}\) is asymptotically linear with the influence
function of the oracle parameter \(\Psi_{H_0}\). When \(q_0\in\overline H_0\), this
oracle parameter equals \(\Psi(P_0)\), but its efficiency bound may be smaller
than the nonparametric bound for \(\Psi\).

For example, \(H_n\) could be selected by cross-validated FQI over a sieve of
increasingly complex classes
\(H_1\subset H_2\subset H_3\subset\cdots\subset H_\infty:=H\), where \(H\)
contains \(q_0\). In that setting, \(H_0\) is the limiting correctly specified
submodel selected by the procedure, rather than a model known in advance.
Alternatively, \(H_n\) could arise from variable selection or from a learned
state-action feature transformation, with \(H_0\) corresponding to the limiting
variables or features. Data-adaptive methods for learning state-action feature
representations have been proposed by \citet{pritz2021jointly} and
\citet{pavse2024state}. Such transformations can also be derived from a fitted
FQI model \(q_{n,H_n}^{\pi}\), for example by one-hot encoding the leaf nodes
of a random forest or gradient-boosted tree model as in Section~3.1 of
\citet{he2014practical}.

\subsection{Asymptotic theory}

Theorem~\ref{theorem::oraclebiassieve} records the second-order approximation
identity that controls the difference between the learned working parameter and
the oracle target. It is the key technical reason that data-adaptive model
selection can be used without adding a first-order bias term, provided the
selected model is stable in the senses stated below. For the approximation
error \(\Psi_{H_n}(P_0) - \Psi(P_0)\) to be negligible, the learned model must
approximate the Bellman image of \(q_0\) and the Riesz representer relevant to
the union model \(H_{n,0}\). In a nested sieve, this means that the selected
class must grow fast enough to approximate \(q_0\), while the additional
directions in \(H_{n,0}\) not captured by \(H_n\) must have negligible
contribution to the union-model representer. For variable-selection and
feature-learning procedures, the same requirement says that omitted limiting
directions, if any, cannot carry first-order Riesz mass.

To make this requirement concrete, suppose \(H_n:=H_{\phi_n}\) and
\(H_0:=H_{\phi_0}\) are induced by feature transformations. For a
transformation \(\phi:\mathcal A\times\mathcal S\to\mathbb R^m\), define
\(H_\phi:=\{f\circ\phi:f:\mathbb R^m\to\mathbb R\}\). The combined model
\(H_{n,0}\) is then \(H_{(\phi_n,\phi_0)}\), where \((\phi_n,\phi_0)\) stacks
the learned and oracle feature maps. Theorem~\ref{theorem::oraclebiassieve}
shows that the approximation bias is negligible when the relevant nuisance
functions based on \(\phi_n\) and \((\phi_n,\phi_0)\) converge to their oracle
counterparts based on \(\phi_0\).
Lemma~\ref{lemma::lipschitzdependent} gives one sufficient condition: under an
almost-sure Lipschitz condition on the corresponding conditional mean, the
\(L^2(P_0)\) error of estimating \(E_{P_0}[Y\mid\phi_0(X)]\) using either
\(E_{P_0}[Y\mid\phi_n(X),\mathcal D_n]\) or
\(E_{P_0}[Y\mid\phi_n(X),\phi_0(X),\mathcal D_n]\) is controlled by
\[
\left\{\int \|\phi_n(x)-\phi_0(x)\|_{\mathbb R^m}^2\,P_{0,X}(dx)\right\}^{1/2}.
\]

We now present the oracle expansion for \(\psi_{n,H_n}\). In the following
conditions, define for each \(P\in\mathcal P\) and model \(H\):
\begin{align*}
\varphi_{P,H}(s,a,y,s')
&:=\mathcal T_P(\alpha_{P,H})(a,s)
\{y+\gamma V^\pi(q_{P,H})(s')-q_{P,H}(a,s)\}\\
&\quad+m((s,a),q_{P,H})-\Psi_H(P).
\end{align*}
Define the estimated score
\begin{align*}
\ell_{n,H_n}(s,a,y,s')
&:=\widehat{\mathcal T}_n(\alpha_{n,H_n})(a,s)
\{y+\gamma V^\pi(q_{n,H_n})(s')-q_{n,H_n}(a,s)\}
+m((s,a),q_{n,H_n}),
\end{align*}
and let \(\varphi_{n,H_n}:=\ell_{n,H_n}-P_0\ell_{n,H_n}\) denote its
\(P_0\)-centered version. This centering is used only in the expansion below;
it does not enter the estimator.

\begin{enumerate}[label=\textbf{(C\arabic*)}, ref=C\arabic*, resume=cond]
    \item \label{cond::nuiscons}
    \textit{Consistency:}
    \(\sqrt n(P_n-P_0)\{\varphi_{n,H_n}-\varphi_{0,H_n}\}=o_p(1)\).

    \item \label{cond::nuisrate}
    \textit{Nuisance and approximation rate:}
    \(\|\widehat{\mathcal T}_n(\alpha_{n,H_n})
      -\mathcal T_0(\alpha_{0,H_n})\|_{P_0}
    \|\mathcal T_0(q_{n,H_n})-\mathcal T_0(q_0)\|_{P_0}
    =o_p(n^{-1/2})\).

    \item \label{cond::stable}
    \textit{Stabilization of the selected model:}
    \(\sqrt n(P_n-P_0)\{\varphi_{0,H_n}-\varphi_{0,H_0}\}=o_p(1)\).

    \item \label{cond::modelbias}
    \textit{Model approximation error:}
    \(\|\mathcal T_0(\alpha_{0,H_n})
      -\mathcal T_0(\alpha_{0,H_{n,0}})\|_{P_0}
    \|\mathcal T_0(q_{0,H_n})-\mathcal T_0(q_0)\|_{P_0}
    =o_p(n^{-1/2})\).
\end{enumerate}

\begin{theorem}
\label{theorem::ALADML}
Assume that, with probability tending to one over the model-selection sample,
Condition~\ref{cond::bounded} holds for \(H=H_n\), \(H=H_0\), and
\(H=H_{n,0}\). Suppose \(H_n\) converges to an oracle submodel
\(H_0\) with \(q_0\in\overline H_0\) in
the sense that Conditions~\ref{cond::nuiscons}--\ref{cond::modelbias} hold.
Then
\[
\psi_{n,H_n}-\Psi(P_0)
=(P_n-P_0)\varphi_{0,H_0}+o_p(n^{-1/2}).
\]
If, in addition, \(\varphi_{0,H_0}\) is efficient for \(\Psi_{H_0}\), then
the ADRL estimator has asymptotic variance equal to the oracle efficiency
bound at \(P_0\).
\end{theorem}

Together, Conditions~\ref{cond::nuiscons} and~\ref{cond::nuisrate} imply
\[
\psi_{n,H_n}-\Psi_{H_n}(P_0)
=(P_n-P_0)\varphi_{0,H_n}+o_p(n^{-1/2}),
\]
so \(\psi_{n,H_n}\) is debiased for the working parameter
\(\Psi_{H_n}(P_0)\). Conditions~\ref{cond::stable} and~\ref{cond::modelbias}
are the model-selection conditions. The former requires the learned-model
centered score to stabilize around its oracle counterpart.
Analogous stability conditions appear in prior work on ADML
\citep{van2023adaptive, van2024adaptive}. The latter, together with
Theorem~\ref{theorem::oraclebiassieve}, ensures that
\(\Psi_{H_n}(P_0)-\Psi(P_0)=o_p(n^{-1/2})\).

\section{Proofs for ADML in Section \ref{sec::ADMLEmain}}

\subsection{Lemma bounding approximation error of estimated features}
\label{appendix::lemmalipschitz}
In the following lemma, condition on the training sample \(\mathcal D_n\). Let
\(X\) be a covariate, and let \(f_n(X)\in L^2(P_0)\) be
\(\mathcal D_n\)-measurable. For feature transformations
\(\varphi_n,\varphi_0\), define
\[
\begin{aligned}
f_{n,(\varphi_n,\varphi_0)}(x)
&:=E_{P_0}[f_n(X)\mid
\varphi_n(X)=\varphi_n(x),\varphi_0(X)=\varphi_0(x),\mathcal D_n],\\
f_{n,\varphi_n}(x)
&:=E_{P_0}[f_n(X)\mid \varphi_n(X)=\varphi_n(x),\mathcal D_n],\\
f_{n,\varphi_0}(x)
&:=E_{P_0}[f_n(X)\mid \varphi_0(X)=\varphi_0(x),\mathcal D_n].
\end{aligned}
\]
These are versions of regular conditional expectations evaluated at the
observed feature value. The lemma generalizes Lemma~1 of
\citet{vanderlaan2024automatic}.

\begin{lemma}
\label{lemma::lipschitzdependent}
Suppose
\[
g_n(t_1,t_2)
:=
E_{P_0}[f_n(X)\mid \varphi_n(X)=t_1,\varphi_0(X)=t_2,\mathcal D_n]
\]
is almost surely \(L_n\)-Lipschitz continuous, where \(L_n\) is
\(\mathcal D_n\)-measurable and \(L_n=O_p(1)\). Then, for a universal
constant \(C<\infty\),
\[
\|f_{n,\varphi_n}-f_{n,(\varphi_n,\varphi_0)}\|_{P_0}
\le CL_n
\|\|\varphi_n-\varphi_0\|_{\mathbb R^d}\|_{P_0}
\]
and
\[
\|f_{n,\varphi_n}-f_{n,\varphi_0}\|_{P_0}
\le CL_n
\|\|\varphi_n-\varphi_0\|_{\mathbb R^d}\|_{P_0}.
\]
\end{lemma}

\begin{proof}
Condition on \(\mathcal D_n\) throughout and treat \(f_n\), \(\varphi_n\),
and \(\varphi_0\) as fixed. For a real-valued function \(f\) and a
vector-valued function \(v\), let \(\Pi_vf\) be a version of
\(E_{P_0}\{f(X)\mid v(X)\}\), evaluated at \(v(x)\).

Let \(g:\mathbb R^d\times\mathbb R^d\to\mathbb R\) be Lipschitz with
constant \(L_n\). By conditional Jensen's inequality and Lipschitz continuity,
\begin{align*}
&\left|g\{\varphi_n(x),\varphi_0(x)\}
-E_{P_0}[g\{\varphi_n(X),\varphi_0(X)\}\mid
\varphi_n(X)=\varphi_n(x),\mathcal D_n]\right|\\
&\quad\le L_n E_{P_0}[\|\varphi_0(x)-\varphi_0(X)\|_{\mathbb R^d}
\mid\varphi_n(X)=\varphi_n(x),\mathcal D_n].
\end{align*}
Under the regular conditional law given \(\varphi_n(X)=\varphi_n(x)\),
\begin{align*}
\|\varphi_0(x)-\varphi_0(X)\|_{\mathbb R^d}
&\le \|\varphi_0(x)-\varphi_n(x)\|_{\mathbb R^d}
+\|\varphi_n(x)-\varphi_n(X)\|_{\mathbb R^d}\\
&\quad+\|\varphi_n(X)-\varphi_0(X)\|_{\mathbb R^d}\\
&=\|\varphi_0(x)-\varphi_n(x)\|_{\mathbb R^d}
+\|\varphi_n(X)-\varphi_0(X)\|_{\mathbb R^d}.
\end{align*}

Now apply this bound with \(g=g_n\). By the law of iterated expectation,
\(\Pi_{\varphi_n}\Pi_{\varphi_n,\varphi_0}f_n=\Pi_{\varphi_n}f_n\). Therefore
\[
\left| \Pi_{\varphi_n,\varphi_0} f_n-\Pi_{\varphi_n} f_n \right|
\le L_n
\left\|\varphi_n-\varphi_0 \right\|_{\mathbb R^d}
+L_n\Pi_{\varphi_n}(\|\varphi_n-\varphi_0\|_{\mathbb R^d}).
\]
Since conditional expectation is a contraction in \(L^2(P_0)\),
\[
\|\Pi_{\varphi_n}(\|\varphi_n-\varphi_0\|_{\mathbb R^d})\|_{L^2(P_0)}
\le
\| \|\varphi_n-\varphi_0\|_{\mathbb R^d}\|_{L^2(P_0)}.
\]
The pointwise bound therefore gives
\[
\| \Pi_{\varphi_n,\varphi_0} f_n-\Pi_{\varphi_n} f_n\|_{L^2(P_0)}
\le 2L_n
\| \|\varphi_n-\varphi_0\|_{\mathbb R^d} \|_{L^2(P_0)}.
\]

With \(f_n\) as in the lemma,
\(\Pi_{\varphi_n,\varphi_0}f_n=f_{n,(\varphi_n,\varphi_0)}\) and
\(\Pi_{\varphi_n}f_n=f_{n,\varphi_n}\), so the first claimed bound follows.
Swapping \(\varphi_n\) and \(\varphi_0\) gives the analogous bound for
\(f_{n,(\varphi_n,\varphi_0)}-f_{n,\varphi_0}\). The second claimed bound
then follows by the triangle inequality.

\end{proof}

\begin{proof}[Proof of Theorem~\ref{theorem::oraclebiassieve}]
Condition on the training sample when \(H_n\) is random. Because
\(q_{0,H_n}\in\overline H_n\subseteq\overline H_{n,0}\) and
\(q_0\in\overline H_0\subseteq\overline H_{n,0}\), the Riesz identity on
\(H_{n,0}\) gives
\[
\begin{aligned}
\Psi_{H_n}(P_0)-\Psi(P_0)
&=\left\langle
\mathcal T_0(\alpha_{0,H_{n,0}}),
\mathcal T_0(q_{0,H_n})-\mathcal T_0(q_0)
\right\rangle_{P_0}.
\end{aligned}
\]
Moreover, \(\mathcal T_0(q_{0,H_n})\) is the \(L^2(P_0)\)-orthogonal
projection of \(\mathcal T_0(q_0)\) onto
\(\mathcal T_0(\overline H_n)\). Since
\(\alpha_{0,H_n}\in\overline H_n\), projection orthogonality yields
\[
\begin{aligned}
\Psi_{H_n}(P_0)-\Psi(P_0)
&=\left\langle
\mathcal T_0(\alpha_{0,H_{n,0}})-\mathcal T_0(\alpha_{0,H_n}),
\mathcal T_0(q_{0,H_n})-\mathcal T_0(q_0)
\right\rangle_{P_0},
\end{aligned}
\]
which is the stated identity.
\end{proof}

\subsection{Proof of Theorem~\ref{theorem::ALADML}}

\begin{proof}[Proof of Theorem~\ref{theorem::ALADML}]
Condition on the model-selection sample. Decompose
\[
\psi_{n,H_n}-\Psi_{H_0}(P_0)
=\psi_{n,H_n}-\Psi_{H_n}(P_0)
+\Psi_{H_n}(P_0)-\Psi_{H_0}(P_0).
\]
By definition, \(\psi_{n,H_n}=P_n\ell_{n,H_n}\). Hence
\begin{align*}
\psi_{n,H_n}-\Psi_{H_n}(P_0)
&=(P_n-P_0)\varphi_{0,H_n}
+(P_n-P_0)\{\varphi_{n,H_n}-\varphi_{0,H_n}\}
+R_{n,H_n}(P_0),
\end{align*}
where
\[
R_{n,H_n}(P_0)
:=
P_0\ell_{n,H_n}-\Psi_{H_n}(P_0).
\]
Condition~\ref{cond::nuiscons} gives
\((P_n-P_0)\{\varphi_{n,H_n}-\varphi_{0,H_n}\}=o_p(n^{-1/2})\). Also, by
Condition~\ref{cond::stable},
\begin{align*}
P_n\varphi_{0,H_n}
&=(P_n-P_0)\varphi_{0,H_n}\\
&=(P_n-P_0)\varphi_{0,H_0}
+(P_n-P_0)\{\varphi_{0,H_n}-\varphi_{0,H_0}\}\\
&=P_n\varphi_{0,H_0}+o_p(n^{-1/2}),
\end{align*}
where \(P_0\varphi_{0,H_n}=P_0\varphi_{0,H_0}=0\). Thus
\[
\psi_{n,H_n}-\Psi_{H_n}(P_0)
=P_n\varphi_{0,H_0}+R_{n,H_n}(P_0)+o_p(n^{-1/2}).
\]
By the Riesz identity on \(H_n\),
\begin{align*}
R_{n,H_n}(P_0)
&=E_{P_0}\left[
\mathcal T_0(\alpha_{0,H_n})
\{\mathcal T_0(q_{n,H_n})-\mathcal T_0(q_{0,H_n})\}
\right]\\
&\quad+E_{P_0}\left[
\widehat{\mathcal T}_n(\alpha_{n,H_n})
\{\mathcal T_0(q_0)-\mathcal T_0(q_{n,H_n})\}
\right].
\end{align*}
Since \(\mathcal T_0(q_{0,H_n})\) is the \(L^2(P_0)\)-projection of
\(\mathcal T_0(q_0)\) onto \(\mathcal T_0(\overline H_n)\),
\[
E_{P_0}\left[
\mathcal{T}_{0}(\alpha)
\{\mathcal{T}_0(q_0)-\mathcal{T}_0(q_{0,H_n})\}
\right]
=0,
\qquad \alpha\in\overline H_n.
\]
Taking \(\alpha=\alpha_{0,H_n}\) in this identity yields
\begin{align*}
   R_{n,H_n}(P_0)
&=
E_{P_0}\left[
\{\widehat{\mathcal{T}}_n(\alpha_{n,H_n})-\mathcal{T}_0(\alpha_{0,H_n})\}
\{\mathcal{T}_0(q_0)-\mathcal{T}_0(q_{n,H_n})\}
\right]\\
&=
O_p\left(
\|\widehat{\mathcal{T}}_n(\alpha_{n,H_n})-\mathcal{T}_0(\alpha_{0,H_n})\|_{P_0}
\,
\|\mathcal{T}_0(q_{n,H_n})-\mathcal{T}_0(q_0)\|_{P_0}
\right)\\
&=o_p(n^{-1/2}),
\end{align*}
where the final two equalities follow from the Cauchy--Schwarz inequality and
Condition~\ref{cond::nuisrate}. Thus
\[
\psi_{n,H_n}-\Psi_{H_n}(P_0)
=P_n\varphi_{0,H_0}+o_p(n^{-1/2}).
\]

It remains to control \(\Psi_{H_n}(P_0)-\Psi_{H_0}(P_0)\). By
Theorem~\ref{theorem::oraclebiassieve} and Condition~\ref{cond::modelbias},
\begin{align*}
\Psi_{H_n}(P_0)-\Psi_{H_0}(P_0)
&=-\left\langle
\mathcal T_0(\alpha_{0,H_n})-\mathcal T_0(\alpha_{0,H_{n,0}}),
\mathcal T_0(q_{0,H_n})-\mathcal T_0(q_0)
\right\rangle_{P_0}\\
&=O_p\!\left(
\|\mathcal T_0(\alpha_{0,H_n})-\mathcal T_0(\alpha_{0,H_{n,0}})\|_{P_0}
\|\mathcal T_0(q_{0,H_n})-\mathcal T_0(q_0)\|_{P_0}
\right)\\
&=o_p(n^{-1/2}),
\end{align*}
where the final two equalities follow from the Cauchy--Schwarz inequality and
Condition~\ref{cond::modelbias}.

Combining the preceding displays gives
\[
\psi_{n,H_n}-\Psi_{H_0}(P_0)
=P_n\varphi_{0,H_0}+o_p(n^{-1/2}).
\]
Because \(q_0\in\overline H_0\), \(\Psi_{H_0}(P_0)=\Psi(P_0)\), proving the stated
asymptotic expansion. Under the theorem's additional efficiency assumption,
\(\varphi_{0,H_0}\) is the efficient influence function for the oracle
parameter \(\Psi_{H_0}\), so the asymptotic variance of \(\psi_{n,H_n}\)
equals the oracle efficiency bound at \(P_0\).

\end{proof}

\section{Data-fusion extension}
\label{sec::datafusion}
\label{appendix::datafusion}
\subsection{Background on confounding-robust data fusion}

Example~\ref{example::contrast} gives a semiparametric way to use historical
or observational data in long-term A/B-test analysis. We consider a setting in
which randomized experimental data \((D=1)\) are augmented with historical or
observational data \((D=0)\). Here \(D\) indexes the data source or
environment. Following Example~\ref{example::abtest}, write the state as
\(S_t=(\bar S_t,D)\). The experimental source identifies the target effect
under randomization, but it may be much smaller than the observational source.
Pooling the sources can improve precision, but only if the analysis accounts
for possible unmeasured confounding in the observational data.

Confounding-robust data-fusion methods address this tradeoff by imposing
structure on the between-source bias rather than assuming that the
observational source is unconfounded
\citep{kallus2018removing, wu2022integrative, shyr2023multi, van2024adaptive}.
In cross-sectional data fusion, fully nonparametric regular estimators
generally do not gain asymptotic efficiency from biased observational data,
because the variance saved by the larger sample can be offset by the cost of
estimating an unrestricted bias function \citep{dang2022cross}. The same
principle motivates restrictions on the dynamic \(Q\)-function: potential
efficiency gains come from structure linking the experimental and
observational sources.

\subsection{Partially linear \texorpdfstring{\(Q\)}{Q}-function restriction}

Let \(q_P^\pi(a,d,\bar s)\) denote
\(q_P^\pi((\bar s,d),a)\), the policy \(Q\)-function evaluated at source
\(D=d\), and use the same shorthand for generic functions \(h\). The
between-source contrast
\[
b_P^\pi(a,\bar s)
:=
q_P^\pi(a,1,\bar s)-q_P^\pi(a,0,\bar s)
\]
plays the role of a dynamic confounding-bias function. For the experimental
population, let \(p_1:=P_0(D=1)>0\) be the source proportion, treated as fixed
by design. We restrict the statistical model to laws \(P\) satisfying
\(P(D=1)=p_1\), and define
\[
m_1(x,q)
:=
\frac{\ind\{d=1\}}{p_1}
\int q(a',1,\bar s)\pi(a'\mid \bar s,1)\,d\nu(a'),
\]
where \(x=(s,a)\) and \(s=(\bar s,d)\). The policy-value target is
\[
\Psi(P)=E_P\{m_1(X,q_P)\},
\]
which at \(P_0\) equals the policy value in the experimental population.
The A/B contrast target in Example~\ref{example::abtest} is obtained by using
the corresponding contrast in \(m_1\).

For a linear space \(\mathcal B\) of functions of \((a,\bar s)\), define
\[
H_{\mathcal B}
:=
\left\{
h\in L^\infty(P_{0,X}):
(a,\bar s)\mapsto h(a,1,\bar s)-h(a,0,\bar s)\in\mathcal B
\right\}.
\]
The contrast is defined on the common support of \((\bar S,A)\) across the
two sources, as in Example~\ref{example::contrast}.
Equivalently, in a finite-dimensional partially linear model one may take
\[
h(a,1,\bar s)-h(a,0,\bar s)=\beta^\top\phi(a,\bar s),
\qquad \beta\in\mathbb R^p .
\]
This is the data-fusion version of Example~\ref{example::contrast}: the
historical \(Q\)-function \(h(a,0,\bar s)\) remains unrestricted, while the
between-source contrast is restricted. The restriction does not require the
observational source to be unconfounded; it only specifies how the
experimental and observational \(Q\)-functions may differ.

The weighted Bellman projection construction applies with
\(H=H_{\mathcal B}\). Let \(\overline H_{\mathcal B}\) denote the
\(L^2(P_{0,X})\)-closure of this working space. The corresponding weighted
Bellman projection parameter is
\[
\Psi_{H_{\mathcal B},w}(P)
=
E_P\{m_1(X,q_{P,H_{\mathcal B},w})\}.
\]
If \(q_0\in\overline H_{\mathcal B}\), then this parameter equals the
experimental target at \(P_0\). If the restriction is misspecified,
\(\Psi_{H_{\mathcal B},w}(P_0)\) remains the weighted Bellman projection
target associated with the chosen semiparametric restriction.

The restriction space can also be chosen data-adaptively. For example, ADRL
may select among feature maps \(\phi\), sparsity levels, or richer spaces
\(\mathcal B\) for the between-source contrast \(b_P^\pi\). Under
\(w\equiv1\), correct specification of the oracle model, and the conditions of
Theorem~\ref{theorem::ALADML}, the adaptive estimator has the same expansion
as if the limiting restriction had been fixed in advance. Observational data
may therefore improve precision through a learned semiparametric restriction,
while the experimental source remains the inferential anchor.

\end{document}